\documentclass[oneside,11pt]{article}

\usepackage{url,mathrsfs,algorithm}
\usepackage{amsmath,wrapfig,color}
\usepackage{amsfonts}
\usepackage{graphicx}
\usepackage{subfigure}

\newtheorem{lemma}{Lemma}

\begin{document}
\title{\Huge A Comparison Study of Nonlinear Kernels\vspace{0.in}}

\author{ \bf{Ping Li} \\
         Department of Statistics and Biostatistics\\
         Department of Computer Science\\
       Rutgers University\\
          Piscataway, NJ 08854, USA\\
       \texttt{pingli@stat.rutgers.edu}\\
}

\date{}

\maketitle

\begin{abstract}
\noindent Compared to  linear kernel, nonlinear kernels can often substantially improve the accuracies of many machine learning algorithms.  In this paper, we compare 5 different nonlinear kernels: min-max, RBF, fRBF (folded RBF), acos, and acos-$\chi^2$, on a wide range of publicly available datasets. The proposed fRBF kernel performs  very similarly to the RBF  kernel. Both RBF and fRBF kernels require an important tuning parameter ($\gamma$). Interestingly, for a significant portion of the datasets, the min-max kernel outperforms the best-tuned RBF/fRBF kernels. The acos kernel and acos-$\chi^2$ kernel also perform well in general  and in some datasets achieve the best accuracies.\\

\noindent One crucial issue with the use of nonlinear kernels is the  excessive computational and memory cost. These days, one increasingly popular strategy is to linearize the kernels through various randomization algorithms. In our study, the randomization method for the min-max kernel demonstrates excellent performance compared to the randomization methods for other types of nonlinear kernels, measured in terms of the number of nonzero terms in the transformed dataset.\\

\noindent Our study provides  evidence for supporting the use of the min-max kernel and the corresponding randomized linearization method (i.e., the so-called ``0-bit CWS''). Furthermore, the results motivate at least two  directions for future research: (i) To develop new (and linearizable) nonlinear kernels for better accuracies; and (ii) To develop better linearization algorithms for improving the current  linearization methods for the RBF kernel, the acos kernel, and the acos-$\chi^2$ kernel. One attempt  is to combine the min-max kernel with the acos kernel or the acos-$\chi^2$ kernel. The advantages of these two new and tuning-free nonlinear kernels are demonstrated vias our extensive experiments. A variety of  new nonlinear kernels can be constructed in a similar fashion.\\

\noindent Like other tools such as (ensembles of) trees and deep nets, nonlinear kernels have been providing effective solutions to many  machine learning tasks. We hope our (mostly empirical) comparison study  will help advance the development of  the theory and the practice of nonlinear kernels.

\end{abstract}

\section{Introduction}

It is known in statistical machine learning and data mining that nonlinear algorithms can often achieve substantially better accuracies than linear methods, although typically nonlinear algorithms are considerably more expensive in terms of the computation and/or  storage cost. The purpose of this paper is to compare the performance of 5 important nonlinear kernels and their corresponding linearization methods, to provide guidelines for practitioners and motivate  new research directions.\\

We start the introduction with the basic linear kernel. Consider two data vectors $u,v\in\mathbb{R}^D$. It is common to use the  normalized linear kernel (i.e., the correlation):
\begin{align}\label{eqn_rho}
\rho = \rho(u,v) = \frac{\sum_{i=1}^D u_i v_i }{\sqrt{\sum_{i=1}^D u_i^2}\sqrt{\sum_{i=1}^Dv_i^2}}
\end{align}
This normalization step is in general a recommended practice. For example, when using LIBLINEAR or LIBSVM packages~\cite{Article:Fan_JMLR08}, it is often suggested  to  first normalize the input   data vectors to  unit $l_2$ norm. The use of linear kernel is extremely popular in practice. In addition to packages such as LIBLINEAR which implement batch linear algorithms, methods based on stochastic gradient descent (SGD)  become increasing important especially for very large-scale applications~\cite{URL:Bottou_SGD}.\\

Next, we will briefly introduce five different types of nonlinear kernels and the corresponding randomization  algorithms for linearizing these kernels. Without resorting to linearization, it is rather difficult to scale nonlinear kernels for large datasets~\cite{Book:Bottou_07}. In a sense, it is not very practically meaningful to discuss nonlinear kernels without knowing how to compute them efficiently. \\

Note that in this paper, we restrict our attention to nonnegative data, which are common in practice. Several nonlinear kernels to be studied  are only applicable to nonnegative data.

\subsection{The acos Kernel}

Consider two data vectors $u, v\in\mathbb{R}^D$. The acos kernel is defined as a monotonic function of the correlation (\ref{eqn_rho}):
\begin{align}
acos(u,v) = 1-\frac{1}{\pi}\cos^{-1}(\rho(u,v)) = 1-\frac{1}{\pi}\cos^{-1}(\rho)
\end{align}
There is a known randomization algorithm~\cite{Article:Goemans_JACM95,Proc:Charikar} for linearizing the acos kernel. That is, if we sample i.i.d. $r_{ij}$ from the standard normal distribution and compute the inner products:
\begin{align}\notag
x_j = \sum_{i=1}^D u_i r_{ij},\hspace{0.3in}  y_j = \sum_{i=1}^D v_i r_{ij}, \hspace{0.3in} r_{ij}\sim N(0,1)
\end{align}
then the following probability relation holds:
\begin{align}
\mathbf{Pr}\left(sign(x_j) = sign(y_j) \right)= acos(u,v)
\end{align}
If we generate independently $k$ such pairs of ($x_j,\ y_j$), we will be able to estimate the probability which  approximates the  acos kernel. Obviously, this is just a ``pseudo linearization'' and the accuracy of approximation improves with increasing sample size $k$.  In the transformed dataset, the number of nonzero entries in each data vector is  exactly $k$.

Specifically, we can encode (expand) $x_j$ (or $y_j$) as a 2-dim vector $[0\ 1]$ if $x_j\geq 0$ and $[1\ 0]$ if $x_j<0$. Then we concatenate $k$ such 2-dim vectors to form a binary vector of length $2k$. The inner product (divided by $k$) between the two new vectors  approximates the probability $\mathbf{Pr}\left(sign(x_j) = sign(y_j) \right)$.
\subsection{The acos-$\chi^2$ Kernel}

The $\chi^2$ kernel is commonly used for histograms~\cite{Proc:Schiele_ECCV96,Article:Chapelle_99}
\begin{align}
&\rho_{\chi^2}(u,v) = \sum_{i=1}^D \frac{2u_iv_i}{u_i+v_i},\hspace{0.3in} \sum_{i=1}^D u_i=\sum_{i=1}^D v_i=1, \hspace{0.2in} u_i\geq 0, \ v_i\geq 0
\end{align}

For the convenience of linearization via randomization, we consider the following acos-$\chi^2$ kernel:
\begin{align}
acos-\chi^2(u,v) = 1-\frac{1}{\pi}\cos^{-1}(\rho_{\chi^2}(u,v))
\end{align}
As shown in~\cite{Proc:Li_NIPS13}, if we sample i.i.d. $r_{ij}$ from the standard cauchy  distribution $C(0,1)$ and again compute the inner product
\begin{align}\notag
x_j = \sum_{i=1}^D u_i r_{ij},\hspace{0.3in}  y_j = \sum_{i=1}^D v_i r_{ij}, \hspace{0.3in} r_{ij}\sim C(0,1)
\end{align}
then we obtain a good approximation (as extensively validated in~\cite{Proc:Li_NIPS13}):
\begin{align}
\mathbf{Pr}\left(sign(x_j) = sign(y_j) \right)\approx \text{acos-}\chi^2(u,v)
\end{align}
Again, we can encode/expand $x_j$ (or $y_j$) as a 2-dim vector $[0\ 1]$ if $x_j\geq 0$ and $[1\ 0]$ if $x_j<0$. In the transformed dataset, the number of nonzeros per data vector is also exactly $k$.

\subsection{Min-Max Kernel}

The min-max (MM) kernel is also defined on nonnegative data:
\begin{align}
&MM(u,v) = \frac{\sum_{i=1}^D \min(u_i,v_i)}{\sum_{i=1}^D \max(u_i,v_i)}, \hspace{0.3in} u_i\geq 0, \ v_i\geq 0
\end{align}
Given $u$ and $v$, the so-called ``consistent weighted sampling'' (CWS)~\cite{Report:Manasse_CWS10,Proc:Ioffe_ICDM10} generates  random tuples:
\begin{align}
\left(i^*_{u,j}, t^*_{u,j}\right)\ \text{ and }\  \left(i^*_{v,j}, t^*_{v,j}\right),\  \  j = 1, 2, ..., k
\end{align}
where $i^*\in[1,\ D]$  and $t^*$ is unbounded. See Appendix~\ref{app_CWS} for details. The basic theoretical result of CWS says
\begin{align}\label{eqn_CWS_Prob}
\mathbf{Pr}\left\{\left(i^*_{u,j}, t^*_{u,j}\right) =  \left(i^*_{v,j}, t^*_{v,j}\right)\right\} = MM(u,v)
\end{align}
The recent  work on ``0-bit CWS''~\cite{Proc:Li_KDD15} showed that, by discarding $t^*$, $\mathbf{Pr}\left\{ i^*_{u,j} =i^*_{v,j}\right\} \approx MM(u,v)$  is a  good approximation, which also leads to a convenient implementation. Basically, we can keep the lowest $b$ bits (e.g., $b=4$ or 8) of $i^*$ and view $i^*$ as a binary vector of length $2^b$  with exactly one 1. This way, the number of nonzeros per data vector in the transformed dataset is also exactly $k$.

\subsection{RBF Kernel and Folded RBF (fRBF) Kernel}

The  RBF (radial basis function) kernel is commonly used. For convenience (e.g.,  parameter tuning), we recommend this version:
\begin{align}
RBF(u,v;\gamma) = e^{-\gamma(1-\rho)}
\end{align}
where $\rho=\rho(u,v)$ is the correlation defined in (\ref{eqn_rho}) and $\gamma>0$ is a crucial tuning parameter.

Based on Bochner’s Theorem~\cite{Book:Rudin_90}, it is known~\cite{Proc:Rahimi_NIPS07} that, if we sample $w\sim uniform(0,2\pi)$, $r_{i}\sim N(0,1)$ i.i.d., and let  $x = \sum_{i=1}^D u_i r_{ij}$, $y = \sum_{i=1}^D v_i r_{ij}$, where $\|u\|_2=\|v\|_2=1$, then we have
\begin{align}\label{eqn_RFF}
E\left(\cos(\sqrt{\gamma} x+w)\cos(\sqrt{\gamma} y+w)\right)
= e^{-\gamma(1-\rho)}
\end{align}
This provides a mechanism for linearizing the RBF kernel.

It turns out that, one can simplify (\ref{eqn_RFF}) by removing the need of $w$. In this paper, we define the ``folded RBF'' (fRBF) kernel as follows:
\begin{align}
fRBF(u,v;\gamma) = \frac{1}{2}e^{-\gamma(1-\rho)} + \frac{1}{2}e^{-\gamma(1+\rho)}
\end{align}
which is monotonic in $\rho\geq0$.

\begin{lemma}\label{lem_fRBF}
Suppose $x\sim N(0,1)$ and $y\sim N(0,1)$ and $E(xy) = \rho$. Then the following identity holds:
\begin{align}
E\left(\cos(\sqrt{\gamma} x)\cos(\sqrt{\gamma} y)\right)
= \frac{1}{2}e^{-\gamma(1-\rho)}+\frac{1}{2}e^{-\gamma(1+\rho)}
\end{align}
\textbf{Proof:}\ \ See Appendix~\ref{app_fRBF}.$\hfill\Box$
\end{lemma}

\subsection{Summary of Contributions}

\begin{enumerate}
\item We propose  the ``folded RBF'' (fRBF) kernel to simplify the linearization step of the traditional RBF kernel. Via our extensive kernel SVM experiments (i.e., Table~\ref{tab_KernelSVM}), we show that the RBF kernel and the fRBF kernel perform  similarly. Then through the experiments on linearizing RBF and fRBF kernels, both linearization schemes also perform similarly.

\item Our classification experiments on kernel SVM illustrate that even the best-tuned RBF/fRBF kernels in many datasets do not perform as well as those tuning-free kernels, i.e., the min-max kernel, the acos kernel, and the acos-$\chi^2$ kernel.

\item It is known that nonlinear kernel machines are in general  expensive in computation and/or storage~\cite{Book:Bottou_07}. For example, for  a small dataset with merely $60,000$ data points, the $60,000 \times 60,000$ kernel matrix already has $3.6\times 10^9$ entries.  Thus, being able to linearize the kernels becomes crucial in practice. Our extensive experiments show that in general, the consistent weighted sampling (CWS) for linearizing the min-max kernel performs  well, compared to randomization methods for linearizing the RBF/fRBF kernel, the acos kernel, or the acos-$\chi^2$ kernel. In particular, CWS usually requires only a relatively small number of samples to reach a good accuracy while other methods typically need a large number of samples.
\item We propose two new nonlinear kernels by combining the min-max kernel with the acos kernel or the acos-$\chi^2$ kernel. This idea can  be  generalized to create other types of nonlinear kernels.
\end{enumerate}

The work in this paper suggests  at least two  interesting directions for future research: (i) To develop  improved kernel functions. For example, the (tuning-free) min-max kernel in some datasets does not perform as well as the best-tuned RBF/fRBF kernels. Thus there is room for improvement. (ii) To develop better randomization algorithms for linearizing the RBF/fRBF kernels, the  acos kernel, and the acos-$\chi^2$ kernel. Existing methods require  too many samples, which means the transformed dataset will have  many nonzeros per data vector (which will cause significant burden on computation/storage).  Towards the end of the paper, we report our proposal of combining the min-max kernel with the acos kernel or the acos-$\chi^2$ kernel. The initial results appear promising.

\clearpage\newpage

\section{An Experimental Study on Kernel SVMs}

\begin{table}[h!]
\caption{\textbf{35 Datasets}. We use the same 35 datasets as in the recent paper~\cite{Proc:Li_KDD15} on 0-bit CWS.  The data are  public (and mostly well-known), from various sources including the UCI repository, the LIBSVM web site, the  web site for the book~\cite{Book:Hastie_Tib_Friedman}, and the   papers~\cite{Proc:Larochelle_ICML07,Proc:ABC_ICML09,Proc:ABC_UAI10}. Whenever possible, we use the conventional partitions of training and testing sets.  The last column reports the best linear SVM classification results (at the best $C$ value) using LIBLINEAR package and $l_2$-regularization (with a tuning parameter $C$). See Figures~\ref{fig_KernelSVM1} to~\ref{fig_KernelSVM3} for  detailed linear SVM results  for all $C$ values.
}
\begin{center}{
{\begin{tabular}{l r r c c }
\hline \hline
Dataset     &\# train  &\# test  &\# dim &linear (\%) \\
\hline
Covertype10k   &10,000 &50,000 &54 &70.9  \\
Covertype20k   &20,000 &50,000 &54 & 71.1\\
IJCNN5k &5,000 &91,701 &22 &91.6   \\
IJCNN10k &10,000 &91,701 &22 & 91.6  \\
Isolet &6,238 &1,559  &617 & 95.5  \\
Letter &16,000 &4,000 &16 &62.4   \\
Letter4k &4,000 &16,000 &16 & 61.2   \\
M-Basic   &12,000 &50,000 &784 & 90.0   \\
M-Image &12,000 &50,000 &784 & 70.7  \\
MNIST10k &10,000 &60,000&784  &90.0   \\
M-Noise1 &10,000 &4,000 &784 &60.3  \\
M-Noise2 &10,000 &4,000 &784 & 62.1  \\
M-Noise3 &10,000 &4,000 &784 &65.2  \\
M-Noise4 &10,000 &4,000 &784 & 68.4  \\
M-Noise5 &10,000 &4,000 &784 & 72.3  \\
M-Noise6 &10,000 &4,000 &784 & 78.7  \\
M-Rand &12,000 &50,000 &784 &78.9    \\
M-Rotate &12,000 &50,000   &784 &48.0  \\
M-RotImg &12,000 &50,000 &784 & 31.4  \\
Optdigits &3,823 &1,797 &64 & 95.3   \\
Pendigits &7,494 &3,498 &16 & 87.6   \\
Phoneme &3,340 &1,169 &256 & 91.4   \\
Protein &17,766 &6,621 &357 & 69.1   \\
RCV1 &20,242 &60,000 &47,236&96.3  \\
Satimage &4,435 &2,000 &36 &78.5  \\
Segment &1,155 &1,155 &19 & 92.6  \\
SensIT20k &20,000 &19,705 &100 &80.5   \\
Shuttle1k &1,000 &14,500 &9 &90.9  \\
Spam  &3,065 &1,536 &54 &92.6   \\
Splice &1,000 &2,175 &60 & 85.1   \\
USPS   &7,291 &2,007 &256    & 91.7  \\
Vowel &528 &462 &10 &40.9     \\
WebspamN1-20k  &20,000 &60,000 &254 & 93.0   \\
YoutubeVision &11,736 &10,000 &512 &62.3  \\
WebspamN1  &175,000 &175,000 &254 & 93.3   \\
\hline\hline
\end{tabular}}
}
\end{center}\label{tab_data}

\end{table}
\clearpage\newpage

Table~\ref{tab_data} lists the 35 datasets  for our experimental study in this paper. These are the same datasets used in a recent paper~\cite{Proc:Li_KDD15} on the min-max kernel and consistent weighted sampling (0-bit CWS). The last column of Table~\ref{tab_data} also presents the best classification results using linear SVM.\\

\begin{figure}[h!]
\begin{center}

\hspace{-0in}\mbox{
\includegraphics[width=2.2in]{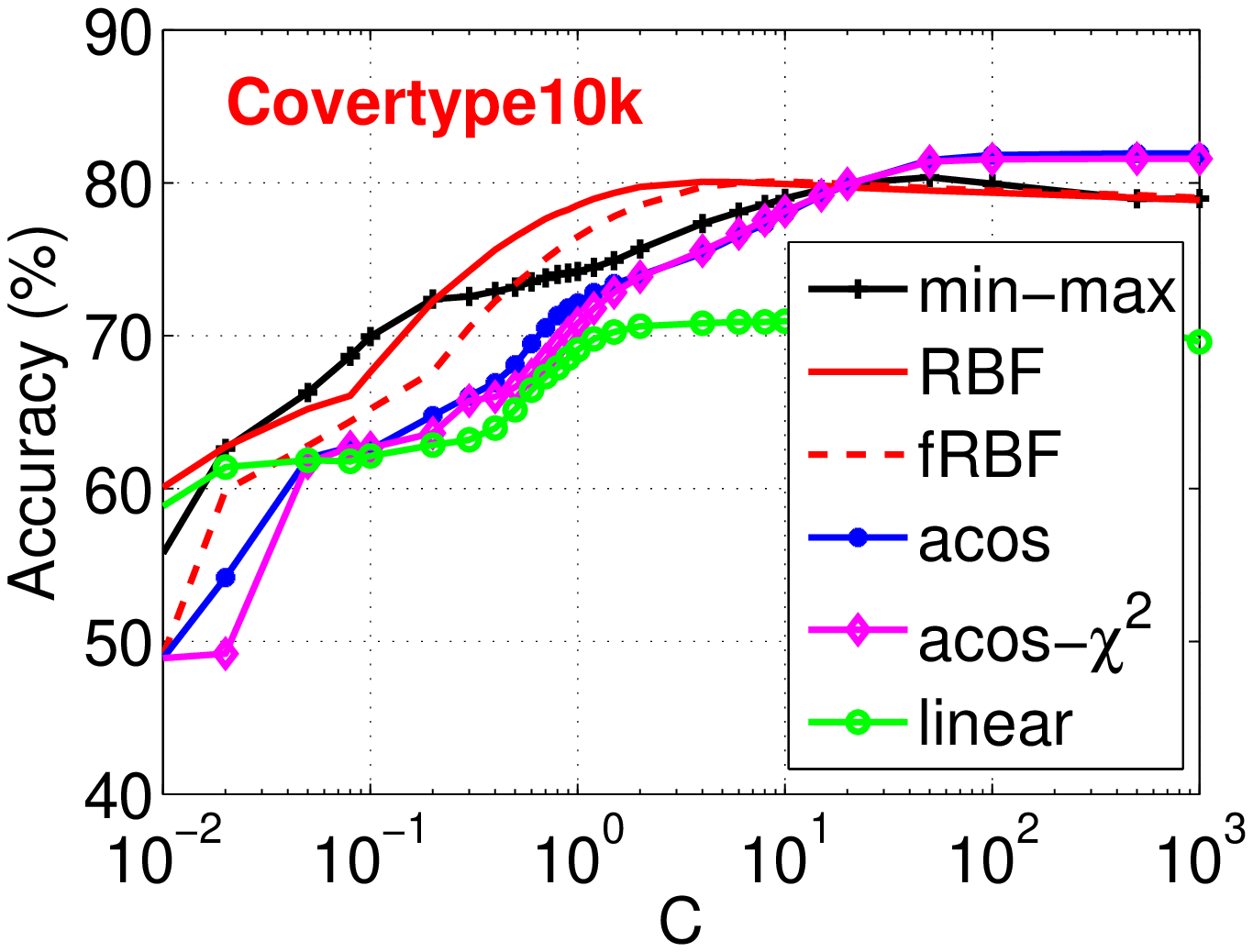}\hspace{0.3in}
\includegraphics[width=2.2in]{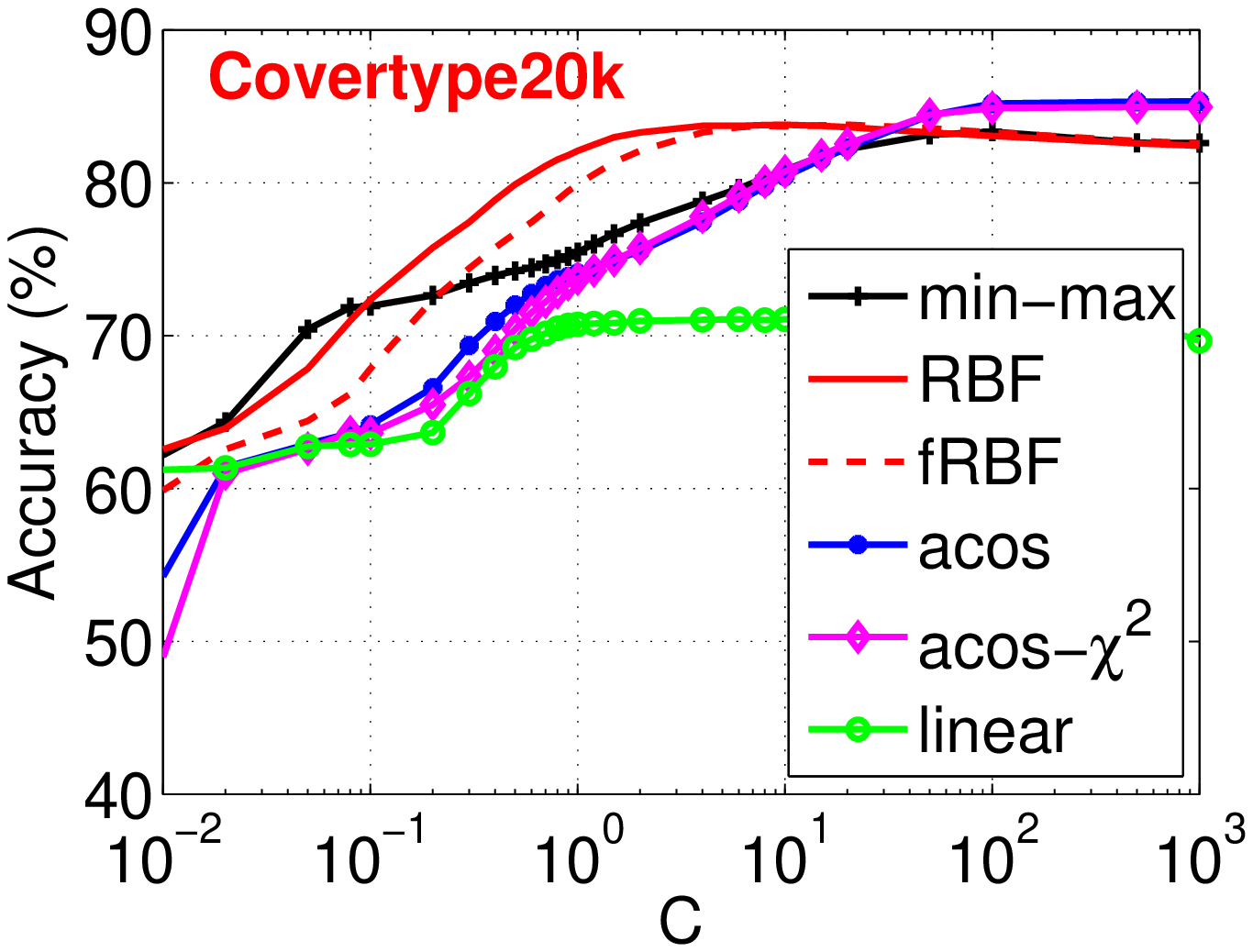}
}

\vspace{-0.038in}

\hspace{-0in}\mbox{
\includegraphics[width=2.2in]{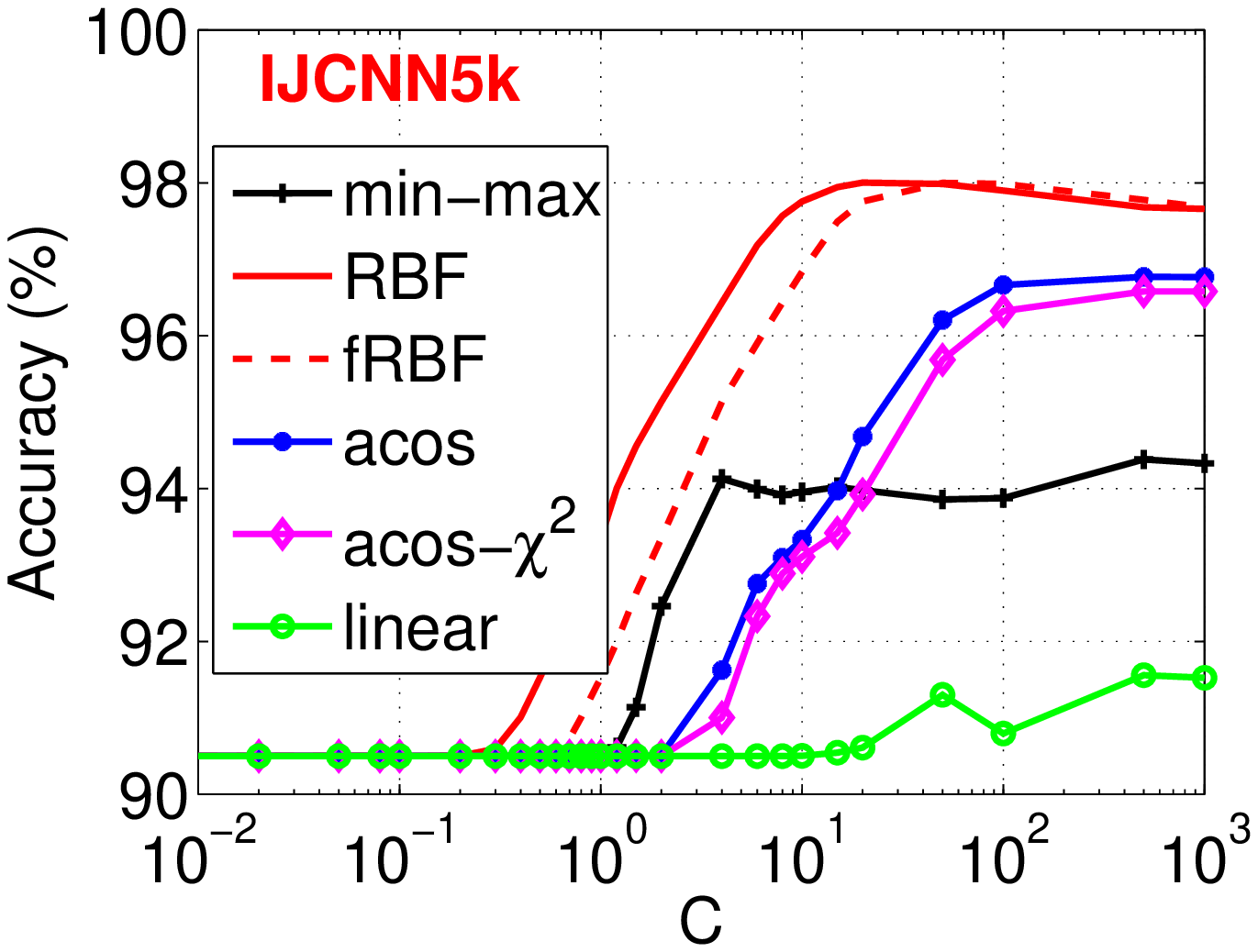}\hspace{0.3in}
\includegraphics[width=2.2in]{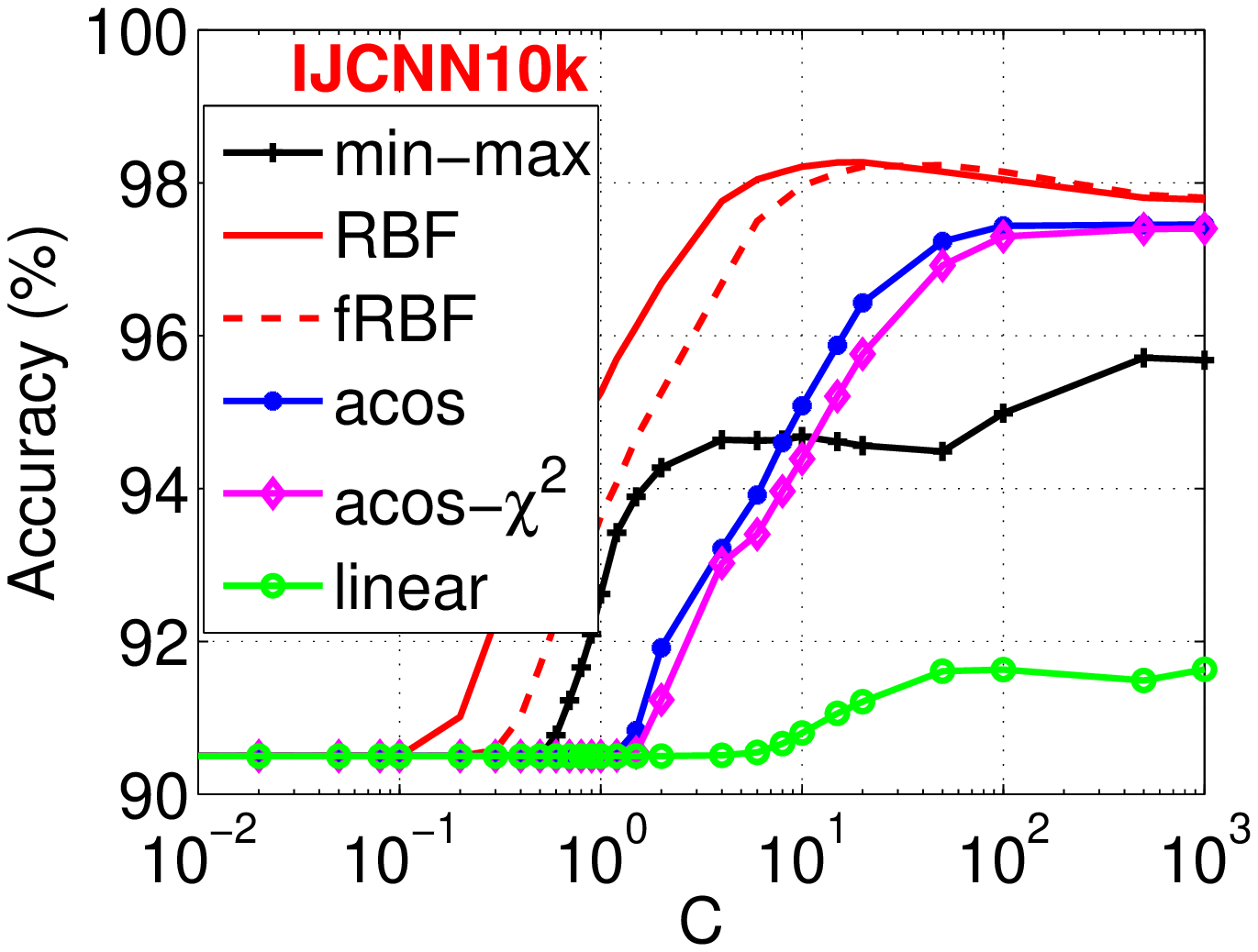}
}

\vspace{-0.038in}

\hspace{-0in}\mbox{
\includegraphics[width=2.2in]{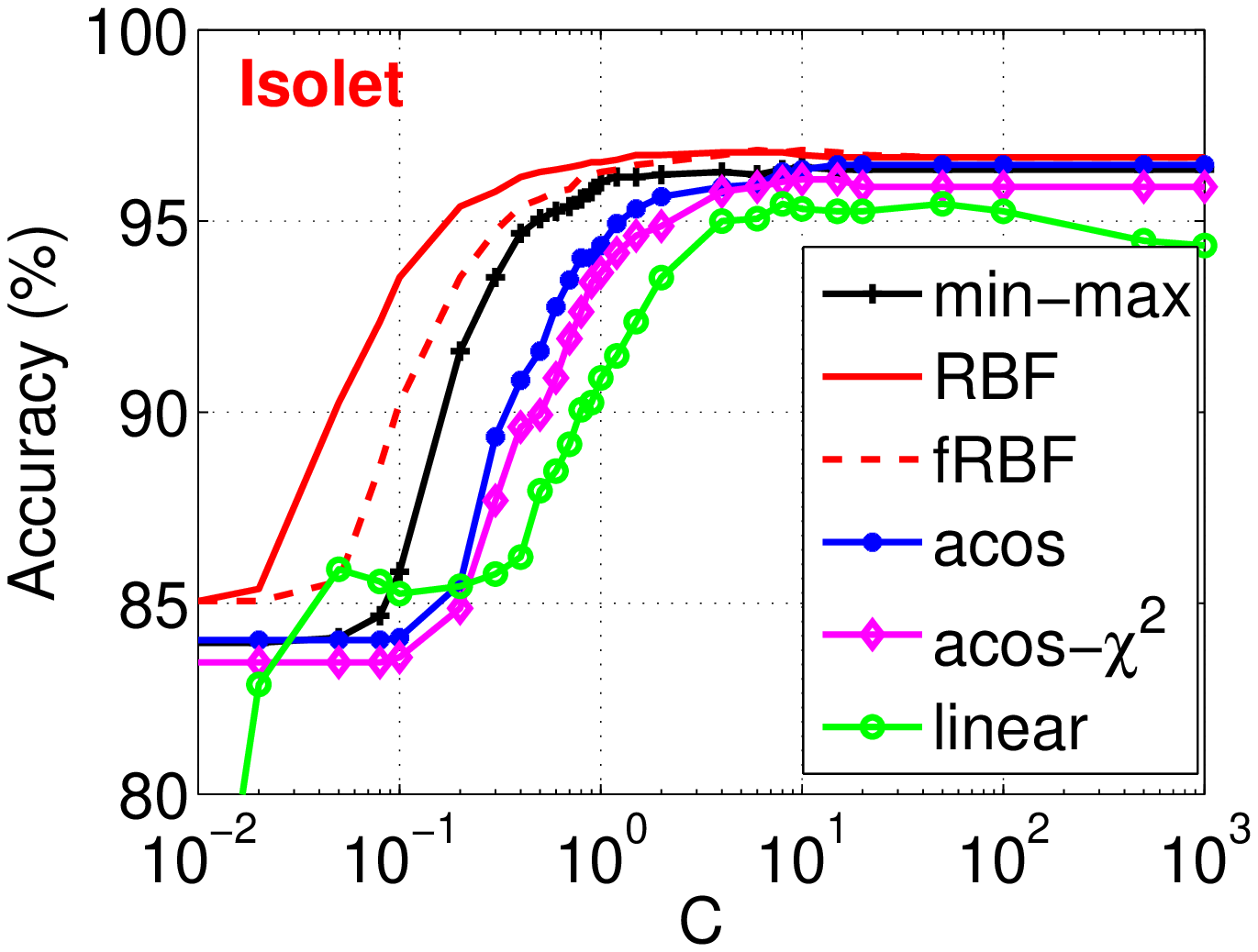}\hspace{0.3in}
\includegraphics[width=2.2in]{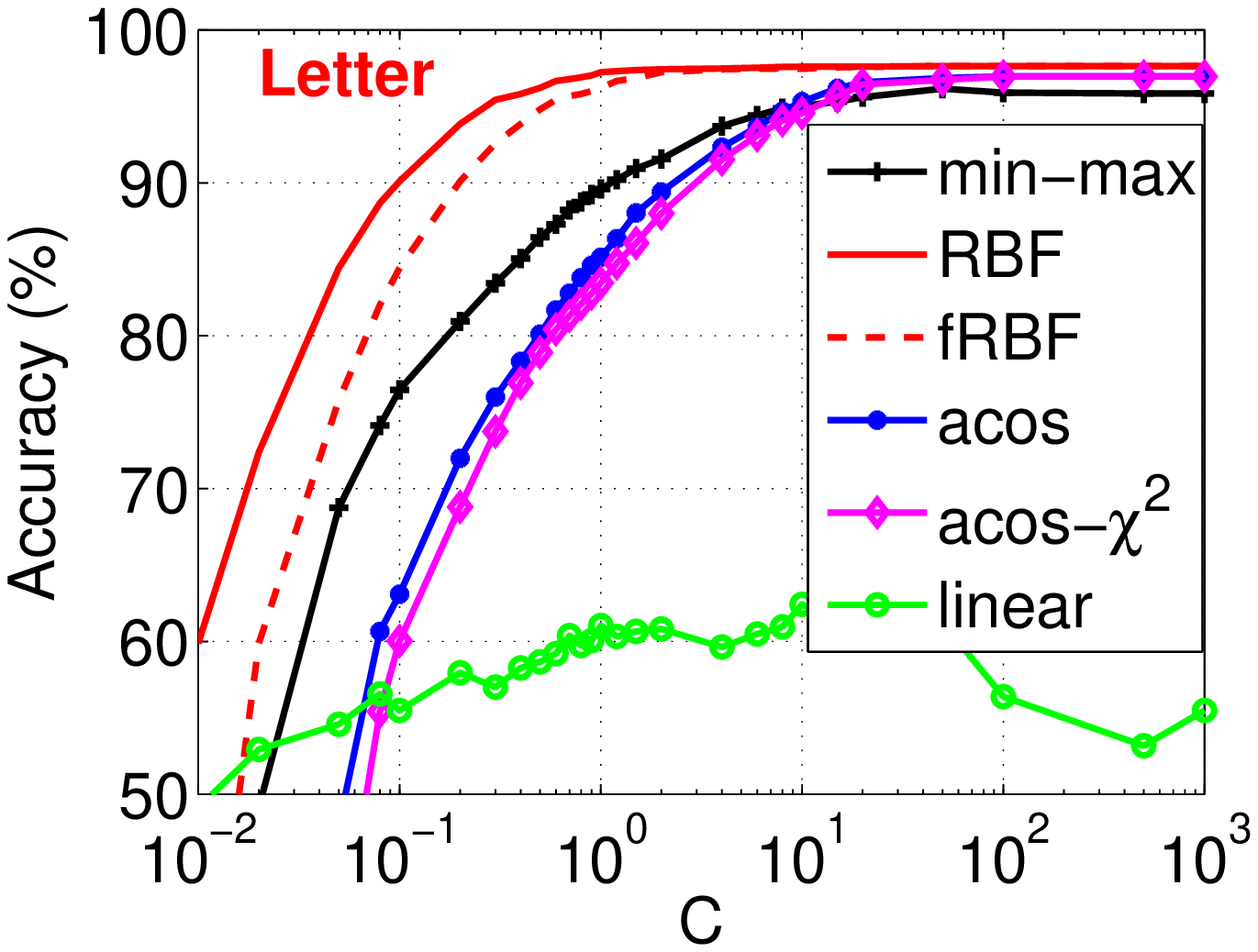}
}

\vspace{-0.038in}

\hspace{-0in}\mbox{
\includegraphics[width=2.2in]{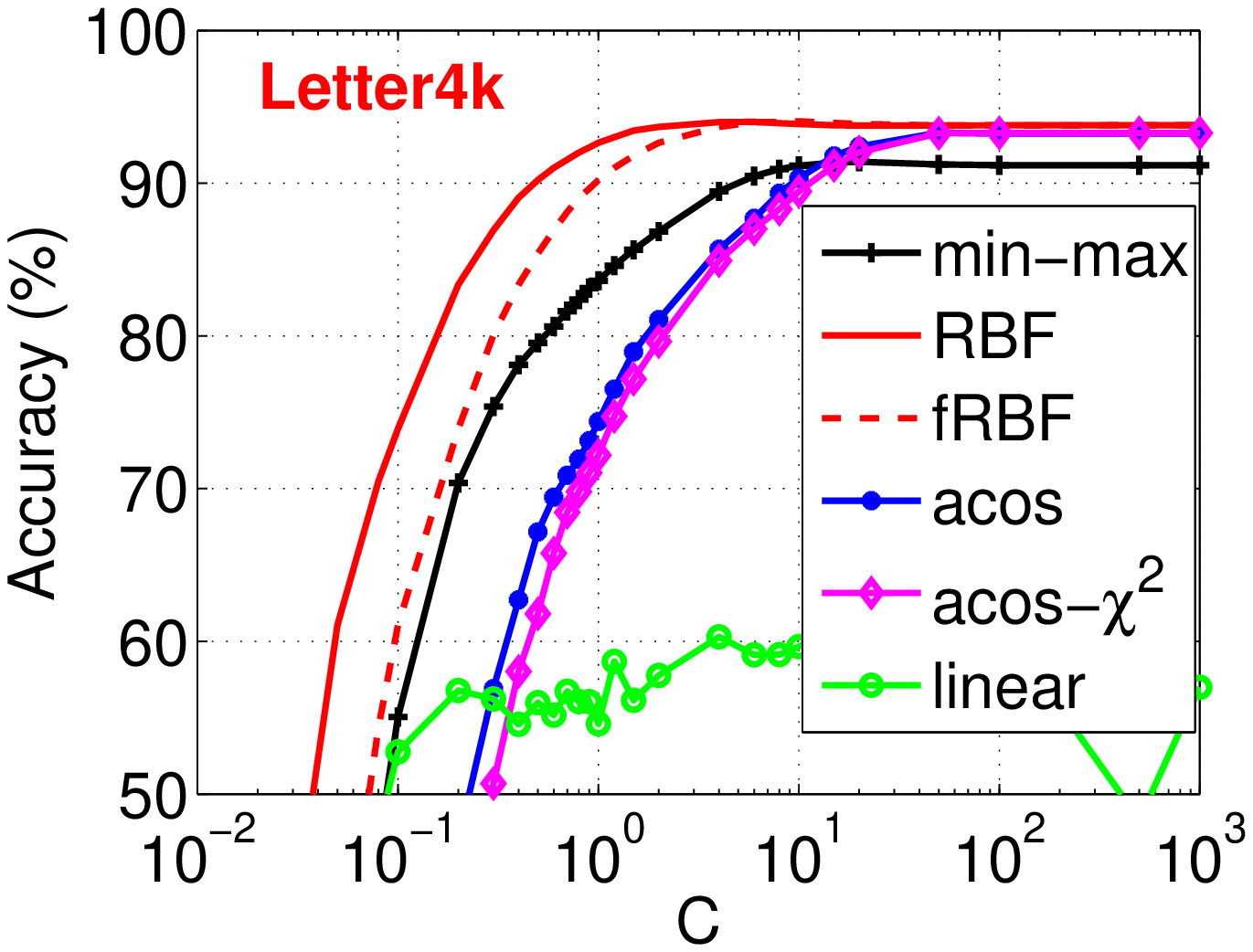}\hspace{0.3in}
\includegraphics[width=2.2in]{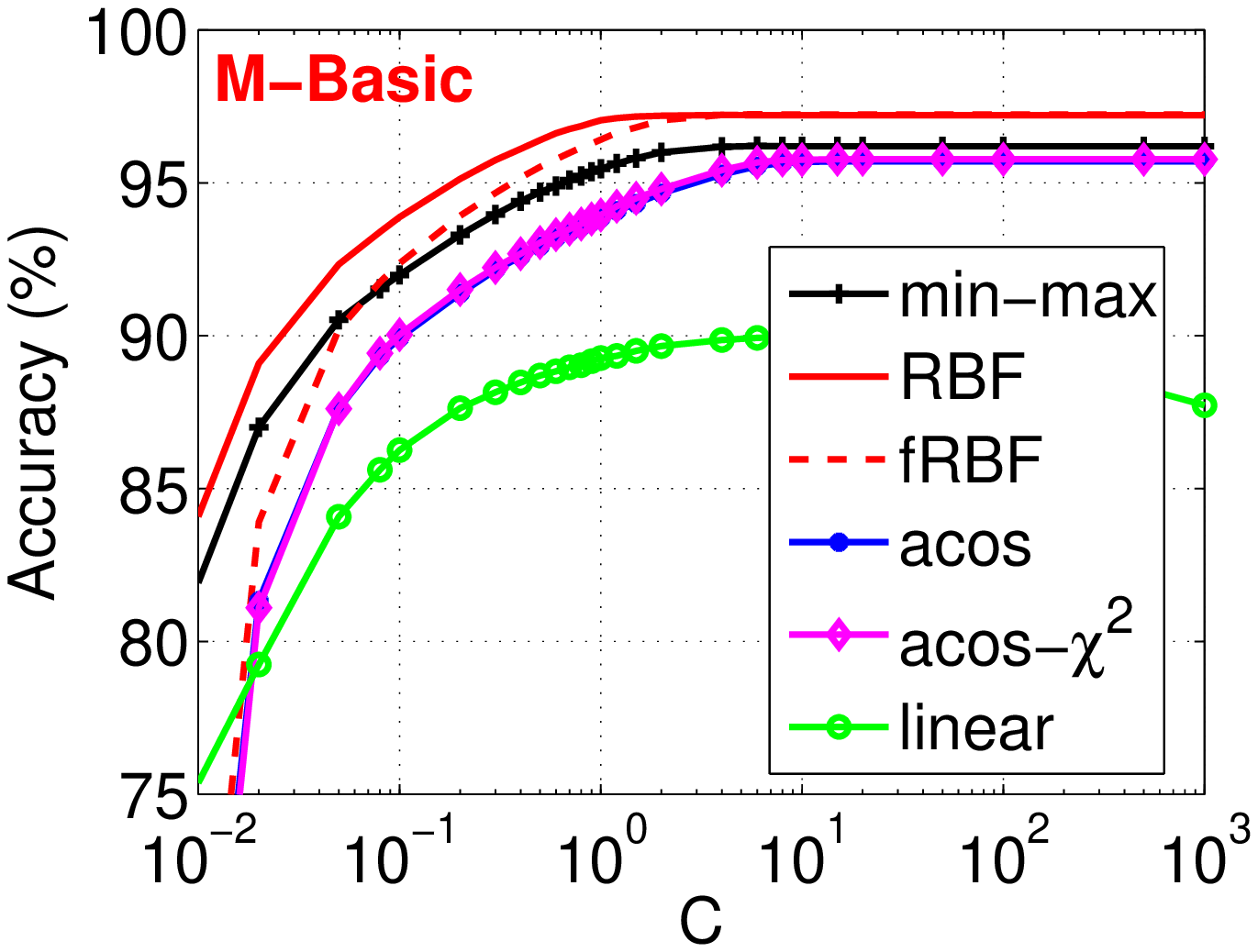}
}

\vspace{-0.038in}

\hspace{-0in}\mbox{
\includegraphics[width=2.2in]{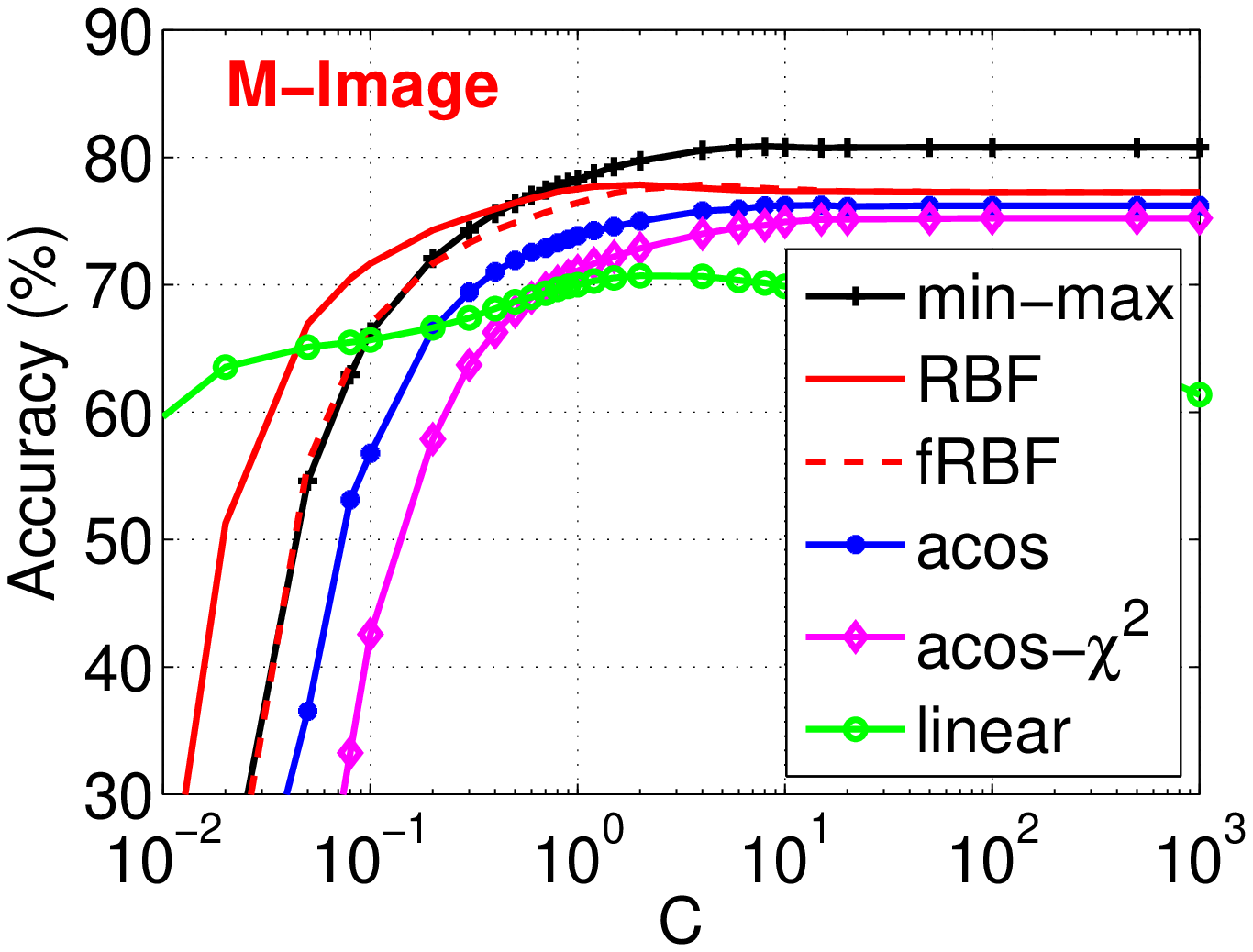}\hspace{0.3in}
\includegraphics[width=2.2in]{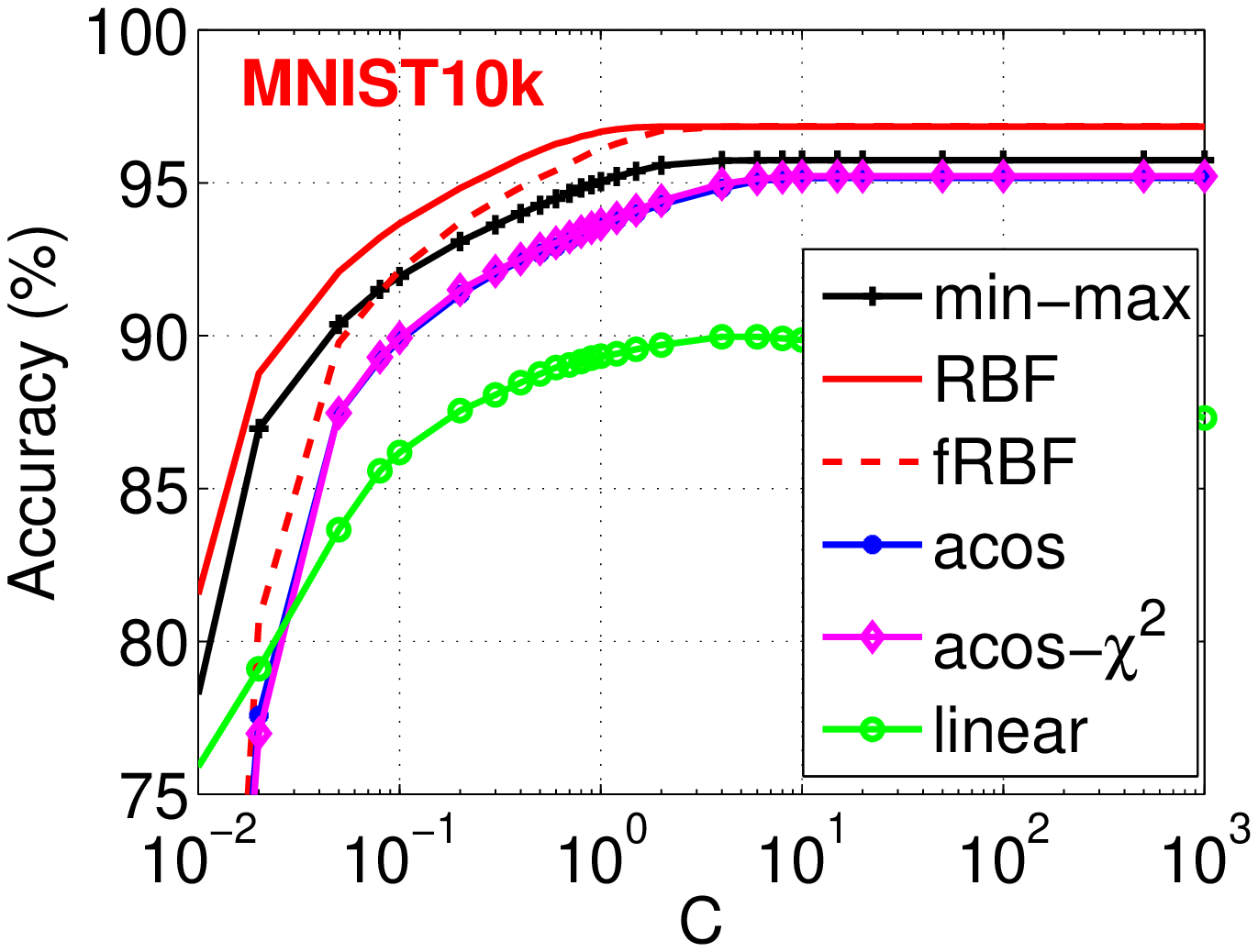}
}

\end{center}
\vspace{-0.3in}
\caption{Test classification accuracies  for 5  nonlinear kernels using $l_2$-regularized  SVM (with a tuning parameter $C$, i.e.,  the x-axis). Each panel presents the results for one dataset (see  data information in Table~\ref{tab_data}). For RBF/fRBF kernels (with a tuning parameter $\gamma$), at each $C$, we report the best accuracy from the results among all $\gamma$ values.  See Figures~\ref{fig_KernelSVM2} and~\ref{fig_KernelSVM3} for  results on more datasets. For comparison, we  include the linear SVM results (green if color is available).   }\label{fig_KernelSVM1}
\end{figure}

\begin{figure}[h!]
\begin{center}

\hspace{-0in}\mbox{
\includegraphics[width=2.2in]{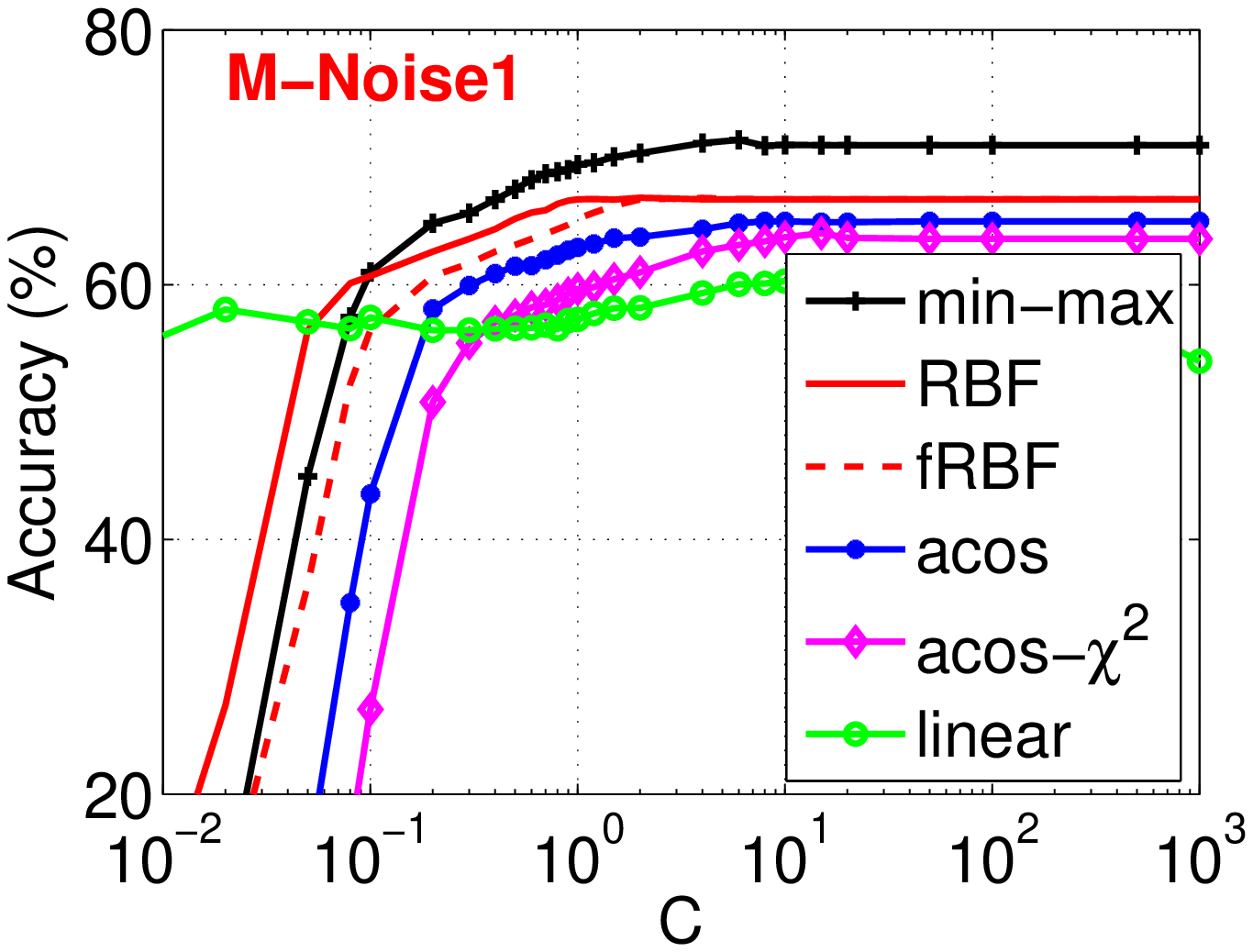}\hspace{0.3in}
\includegraphics[width=2.2in]{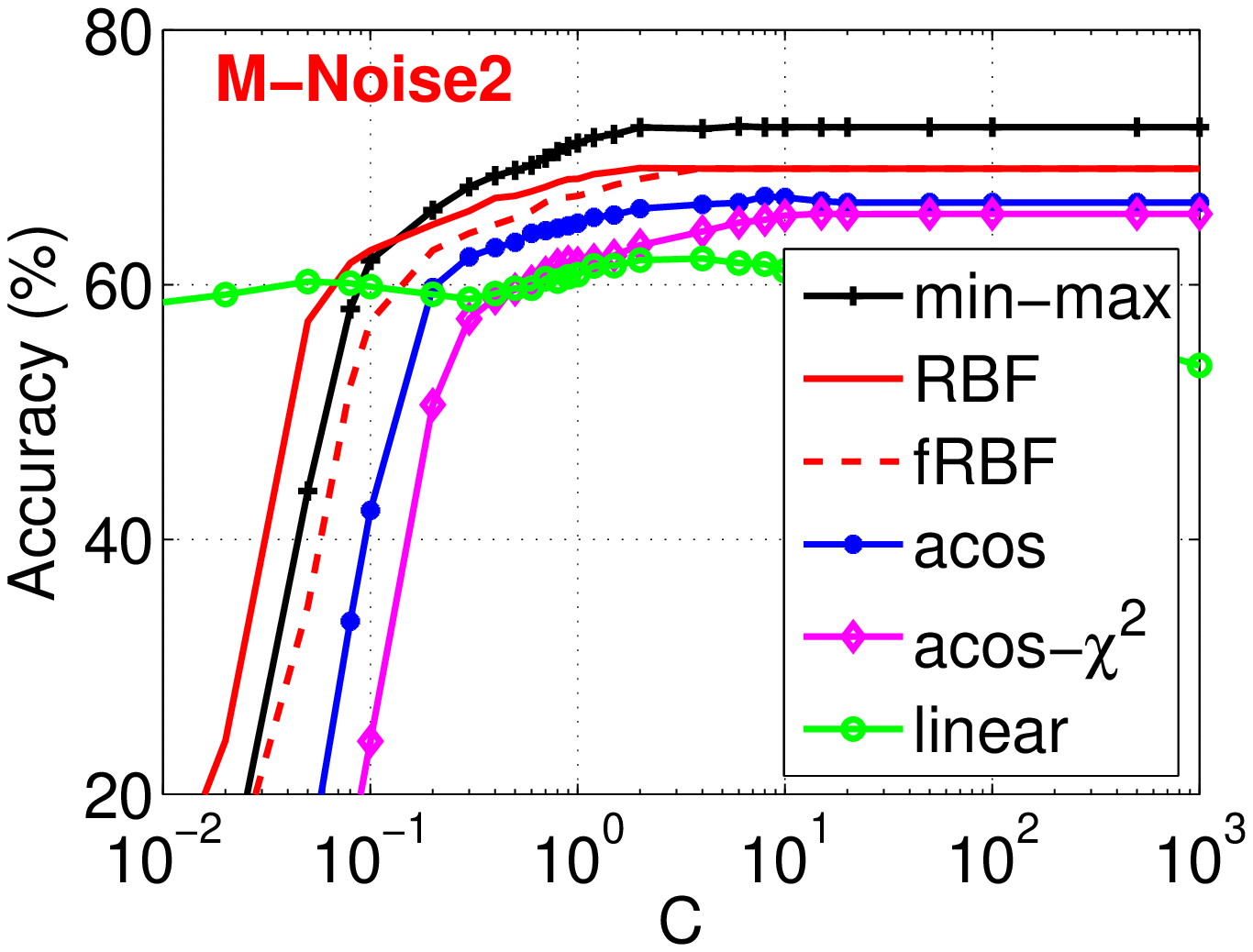}
}

\vspace{-0.041in}

\hspace{-0in}\mbox{
\includegraphics[width=2.2in]{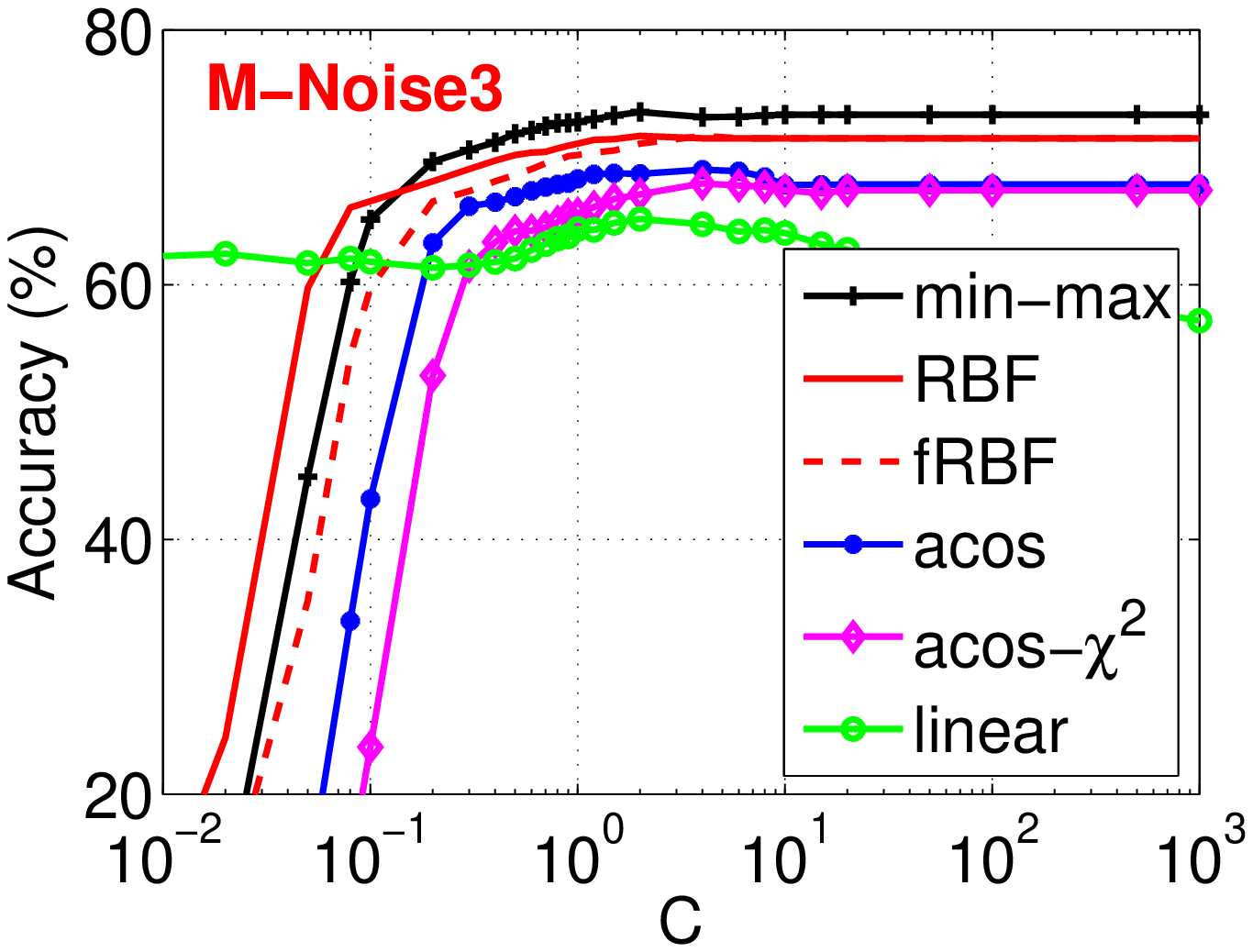}\hspace{0.3in}
\includegraphics[width=2.2in]{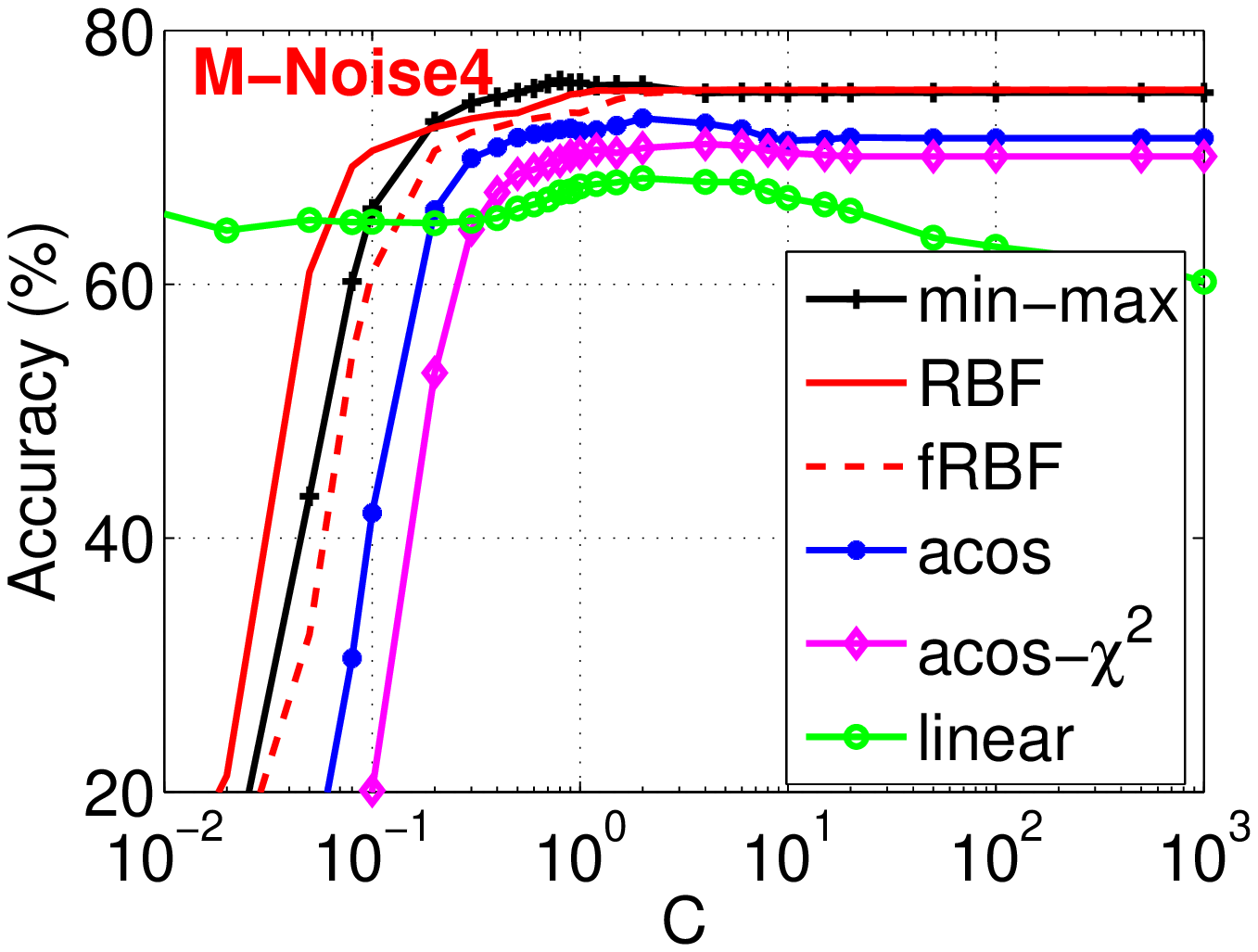}
}

\vspace{-0.041in}

\hspace{-0in}\mbox{
\includegraphics[width=2.2in]{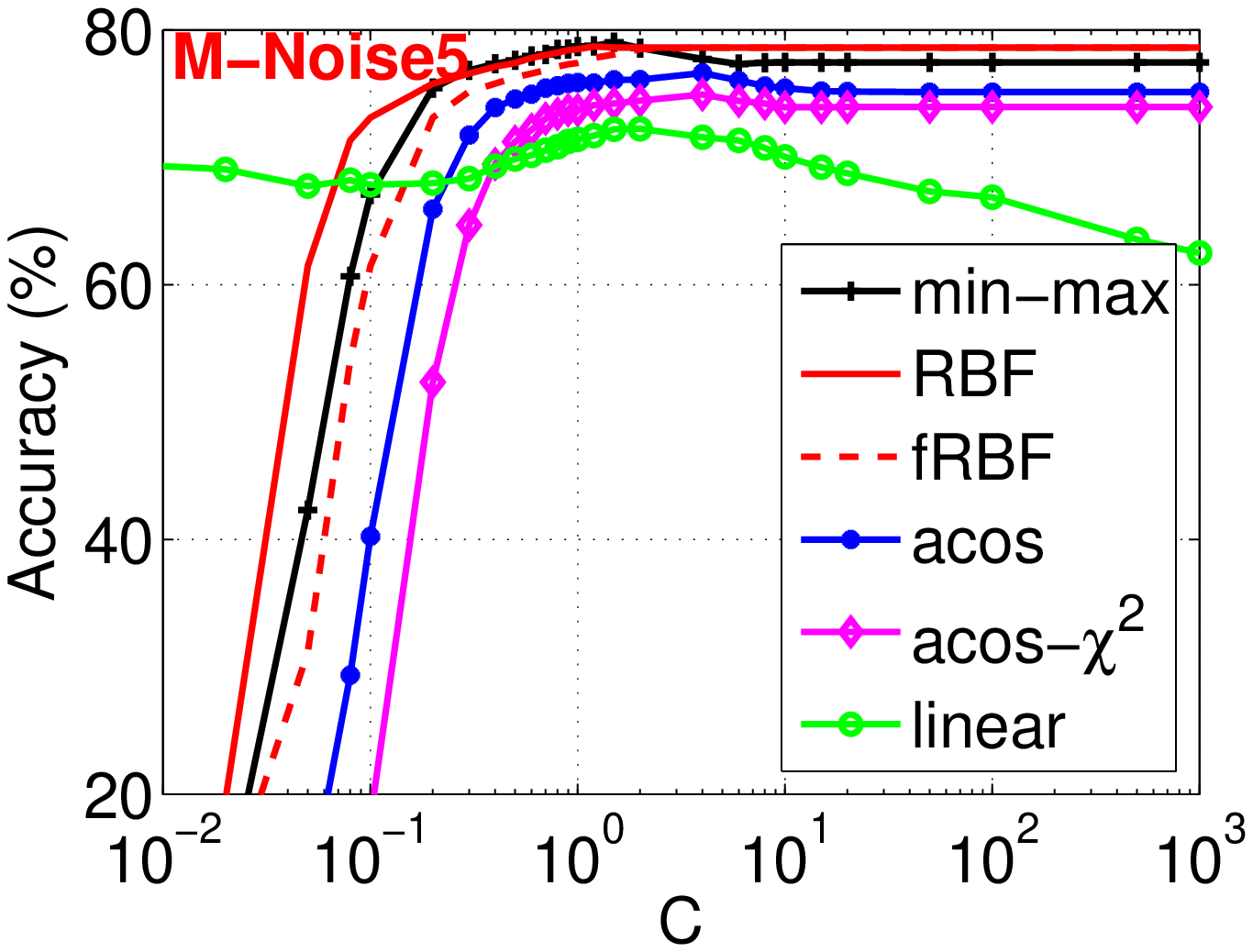}\hspace{0.3in}
\includegraphics[width=2.2in]{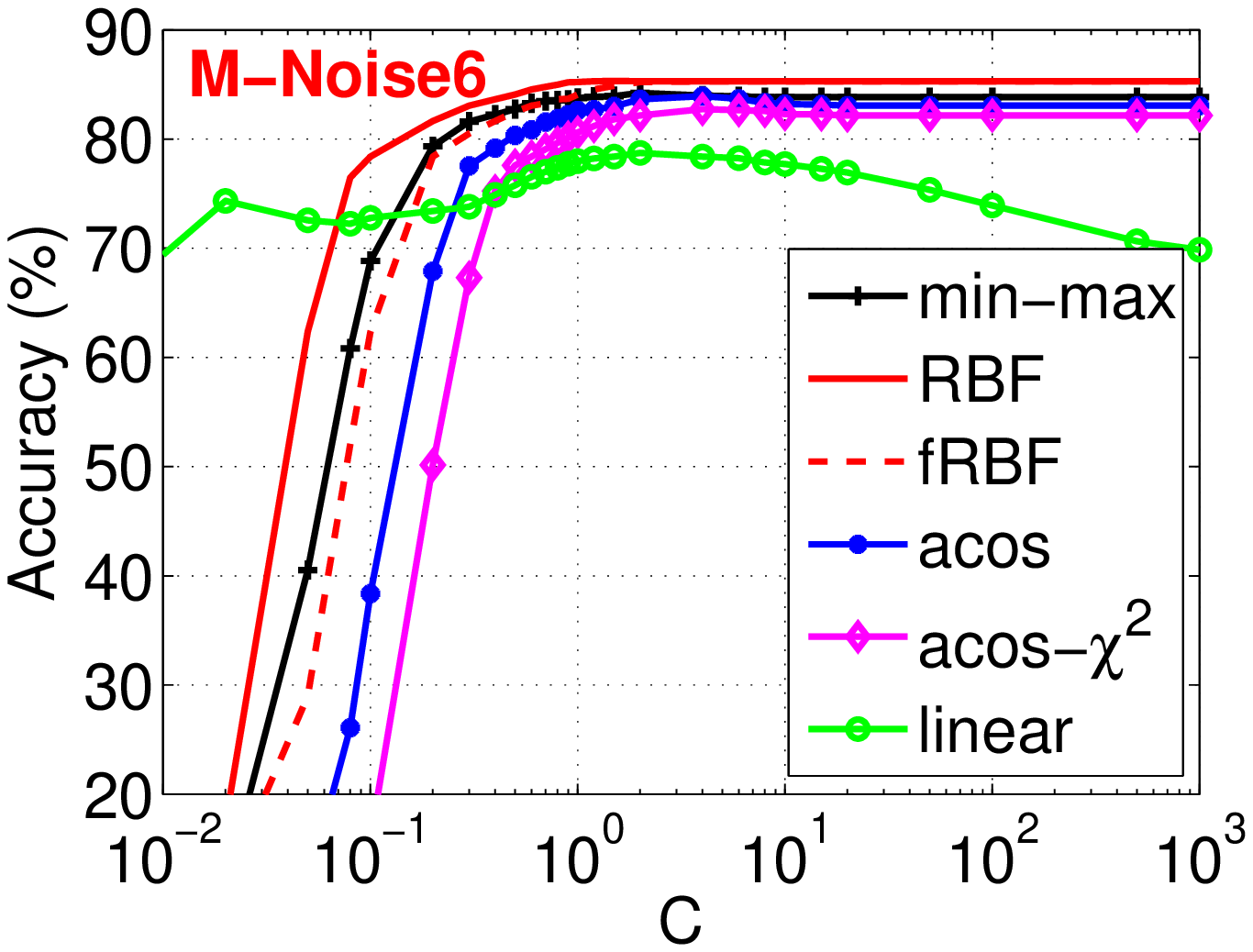}
}

\vspace{-0.041in}

\hspace{-0in}\mbox{
\includegraphics[width=2.2in]{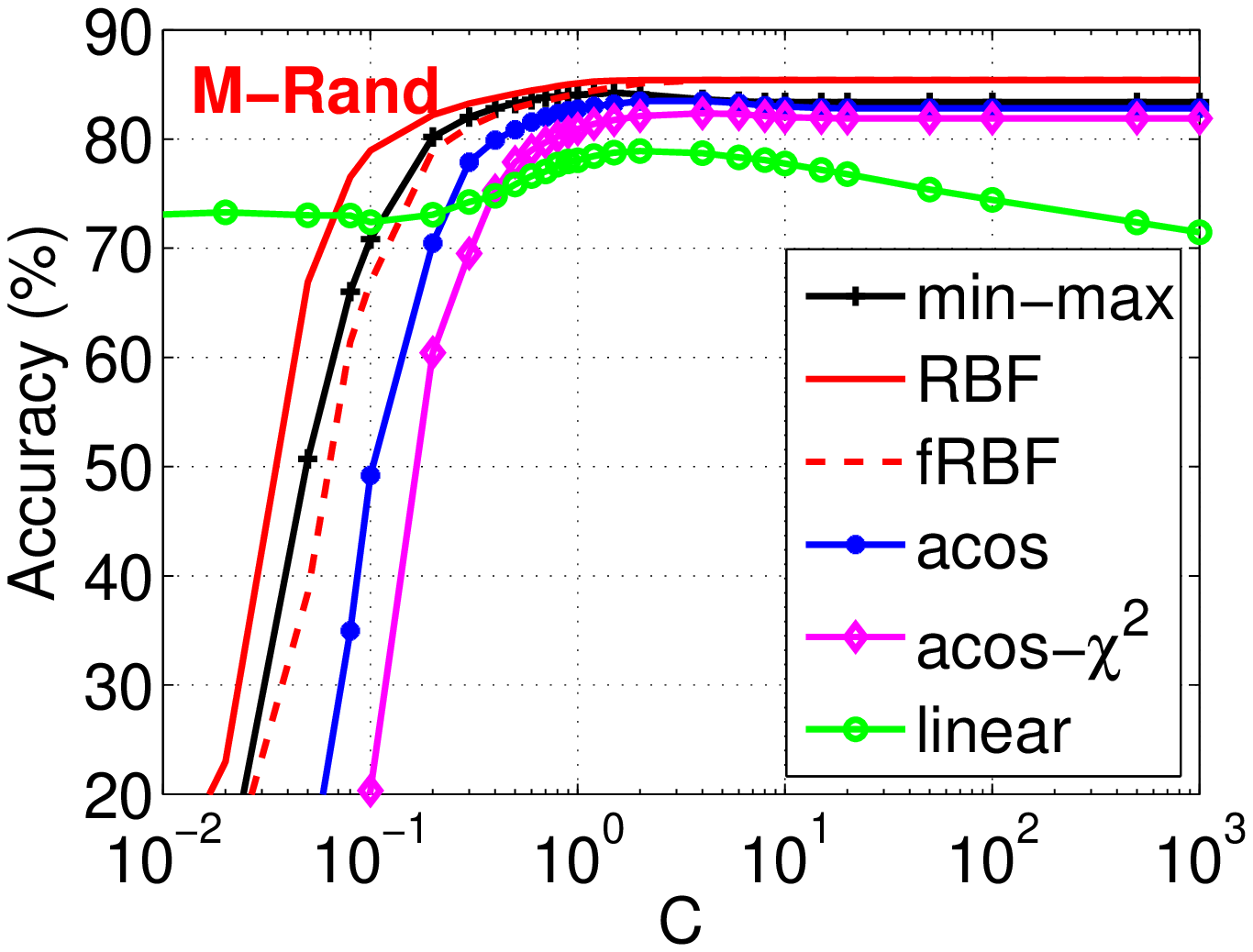}\hspace{0.3in}
\includegraphics[width=2.2in]{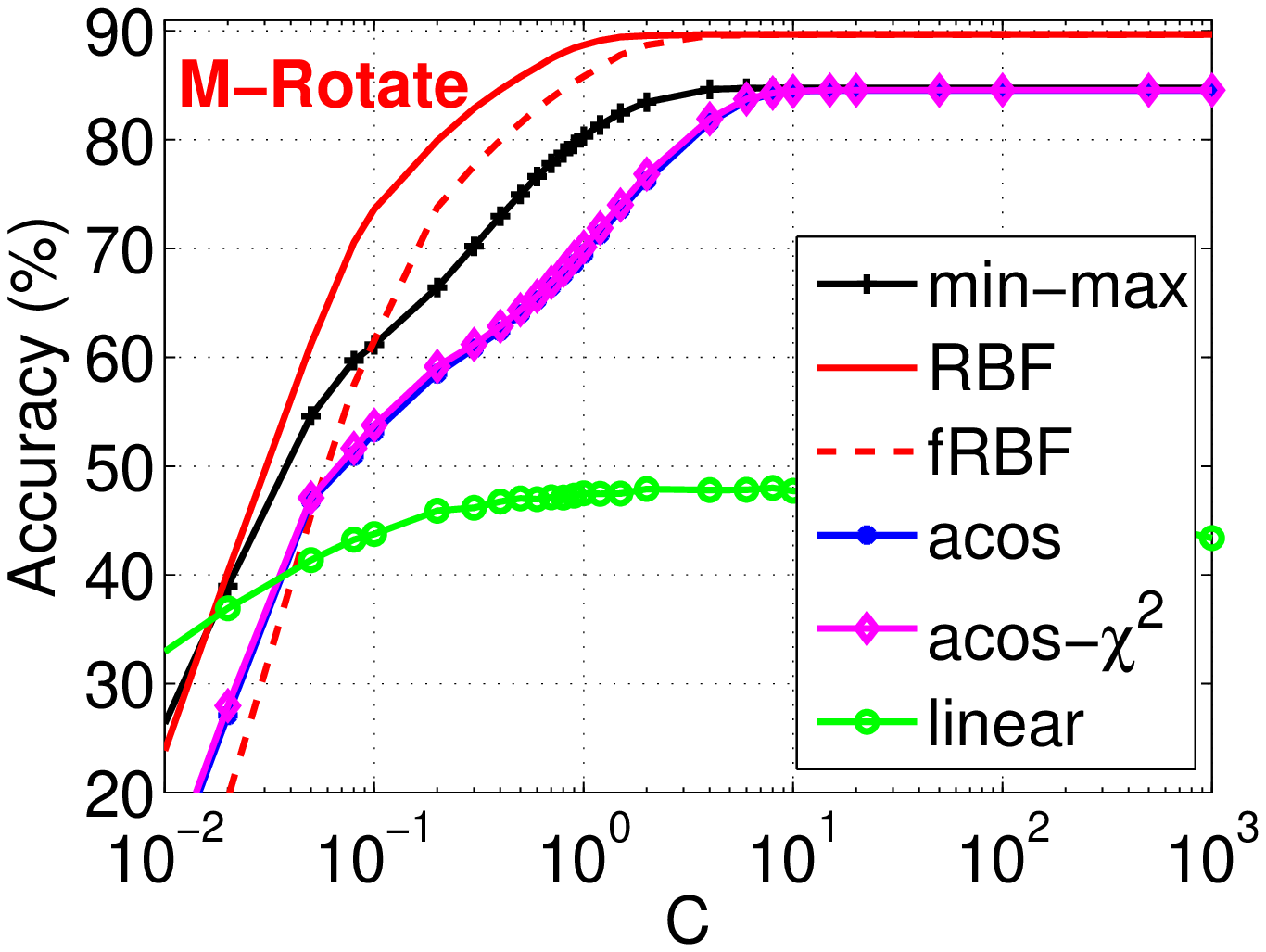}
}

\vspace{-0.041in}

\hspace{-0in}\mbox{
\includegraphics[width=2.2in]{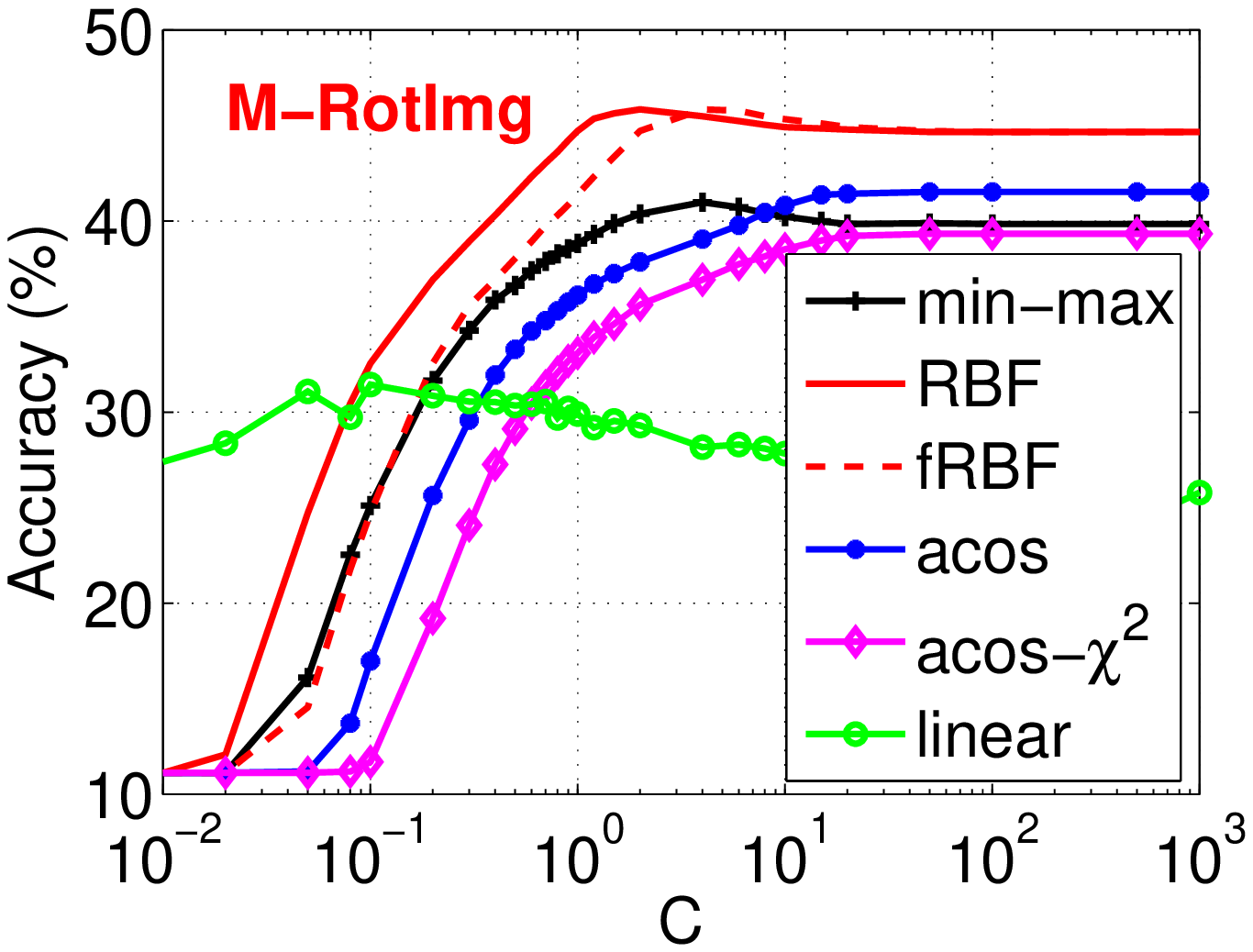}\hspace{0.3in}
\includegraphics[width=2.2in]{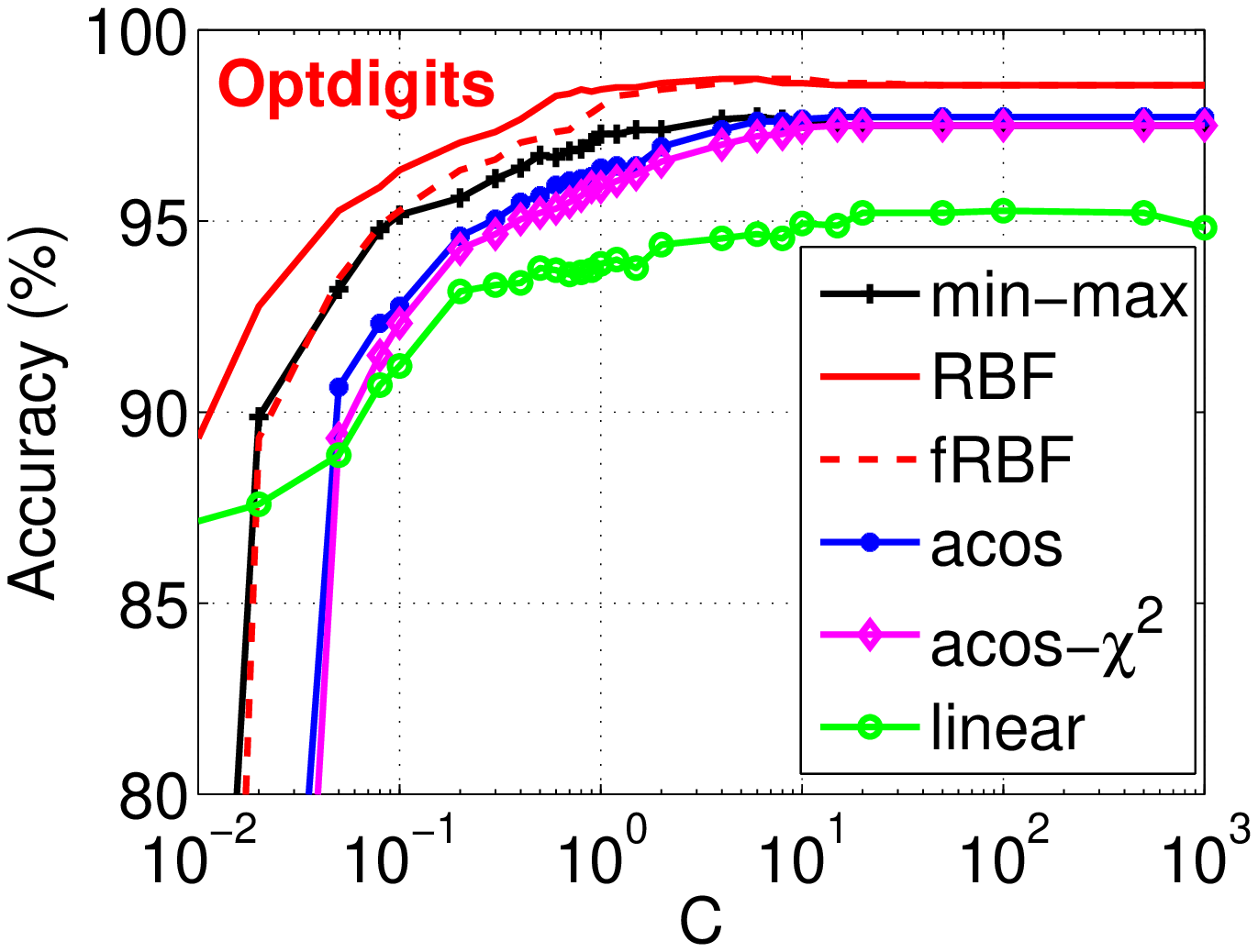}
}

\vspace{-0.041in}

\hspace{-0in}\mbox{
\includegraphics[width=2.2in]{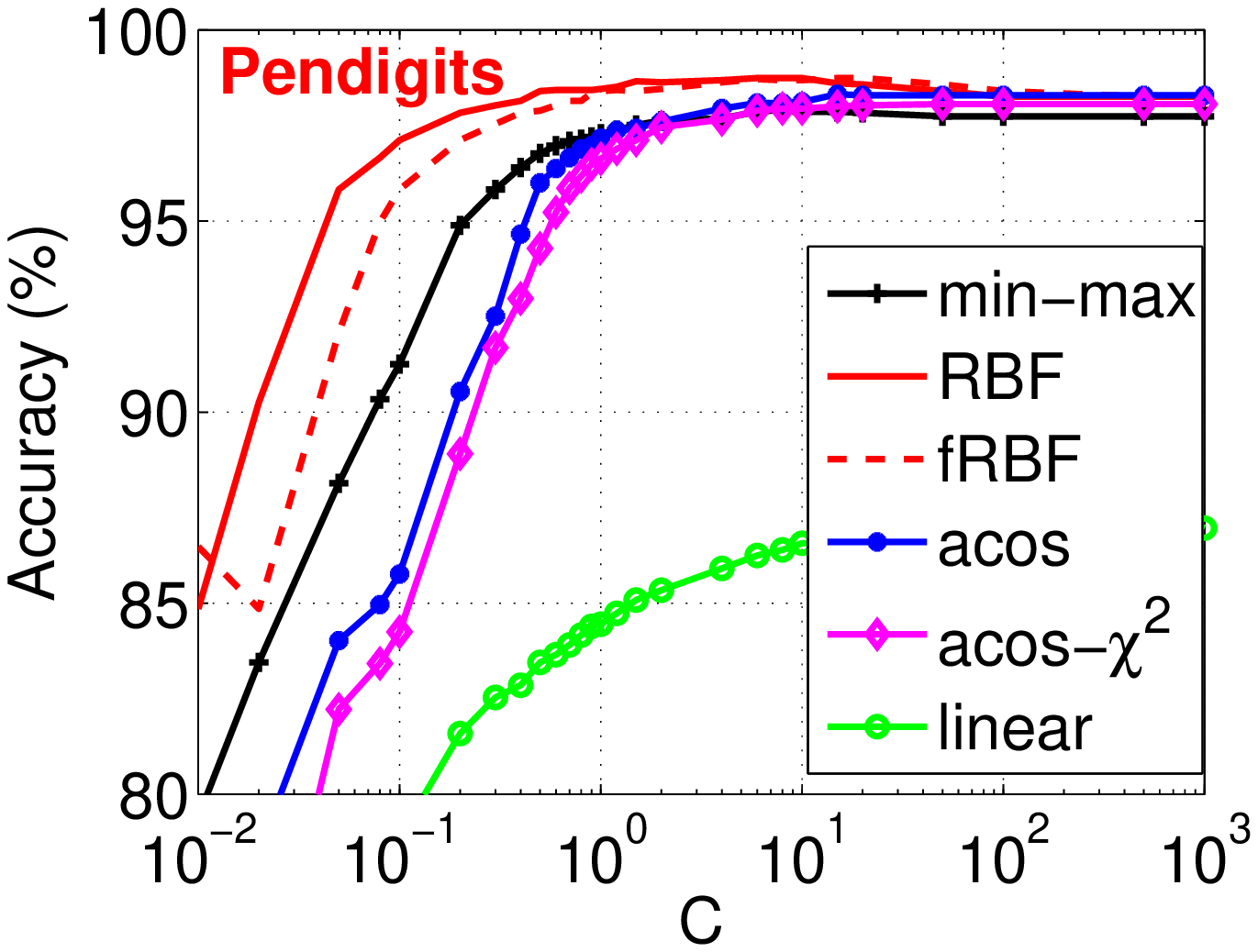}\hspace{0.3in}
\includegraphics[width=2.2in]{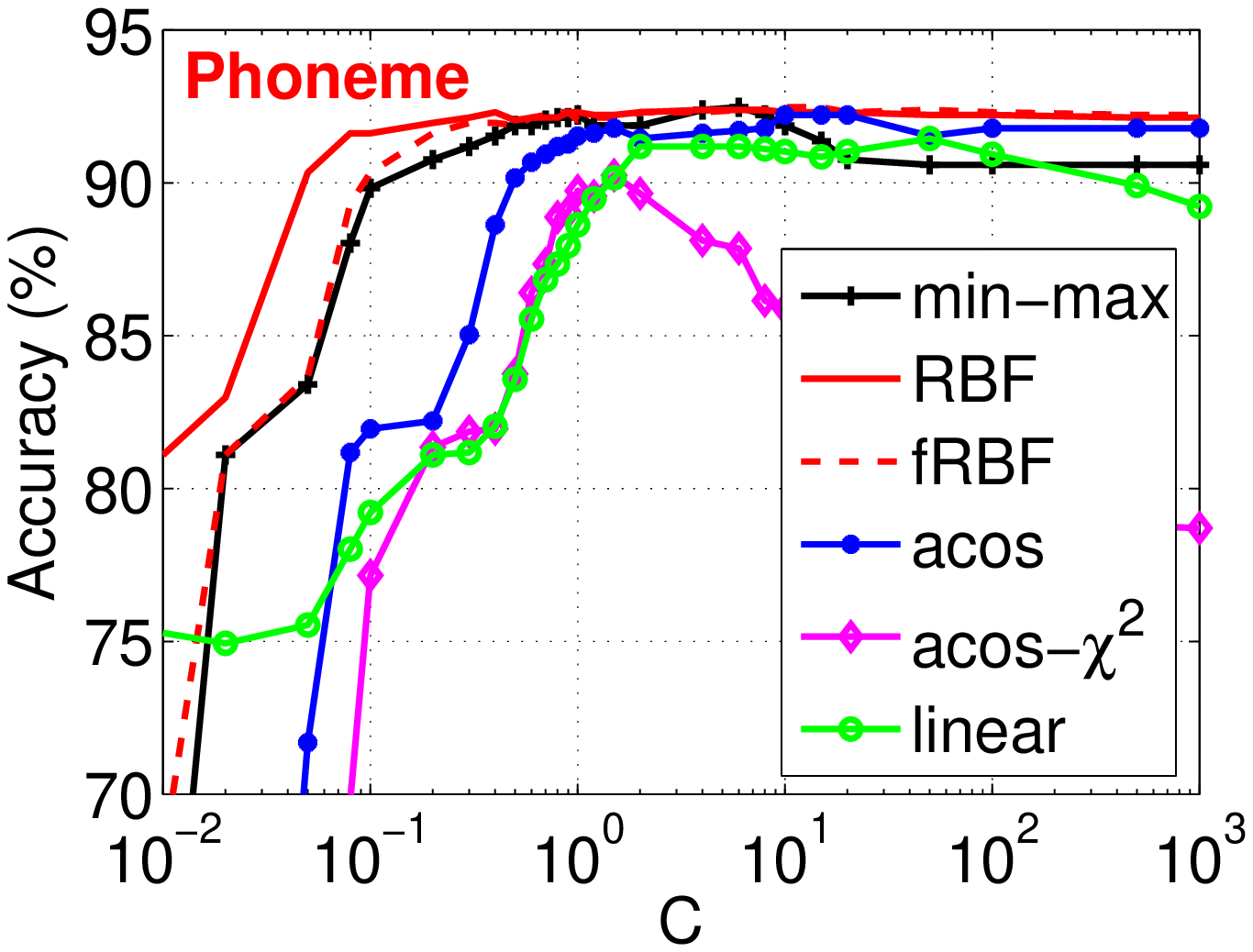}
}

\end{center}
\vspace{-0.4in}
\caption{Test classification accuracies  for  5 nonlinear kernels using $l_2$-regularized  SVM.}\label{fig_KernelSVM2}
\end{figure}

\begin{figure}[h!]
\begin{center}

\hspace{-0in}\mbox{
\includegraphics[width=2.2in]{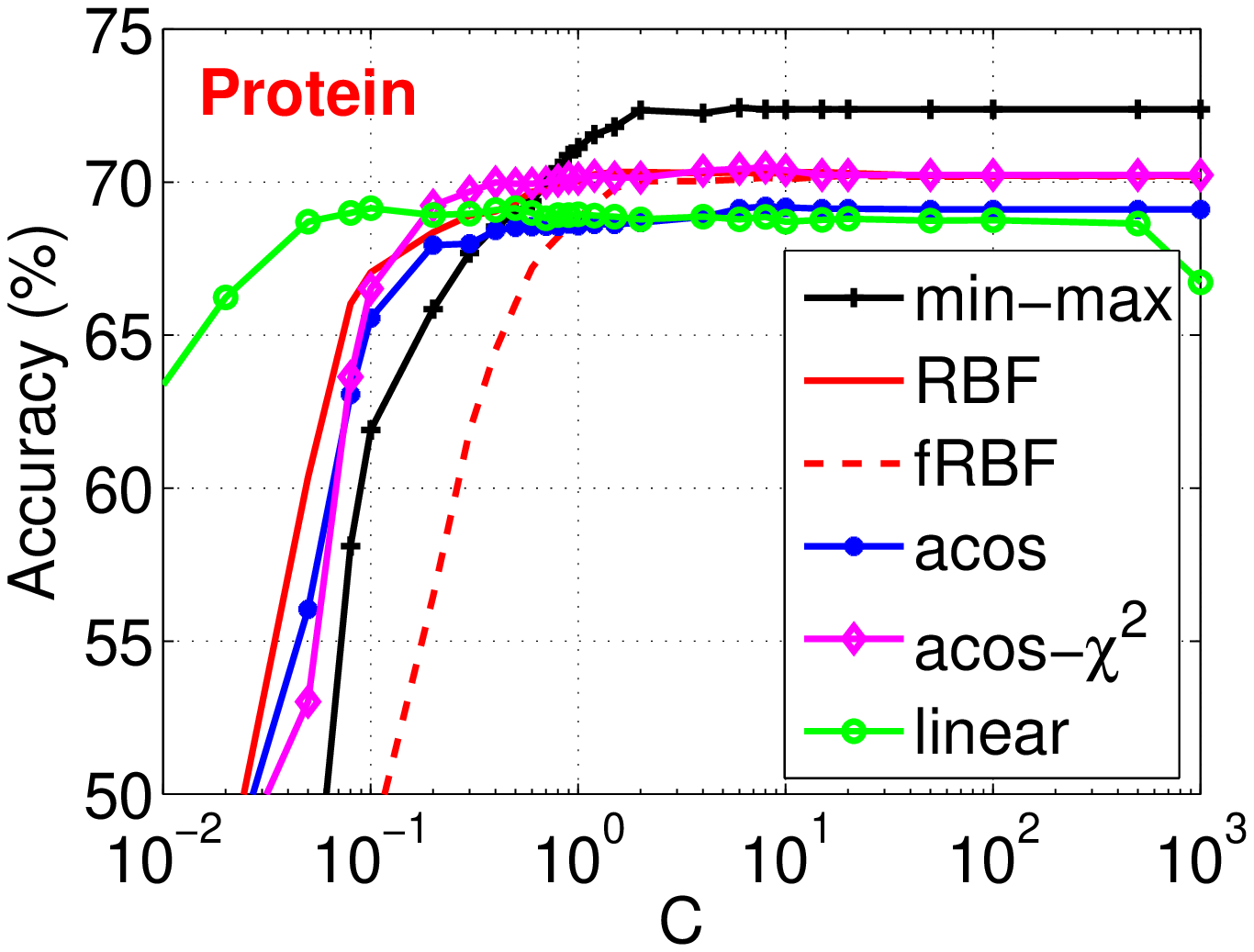}\hspace{0.3in}
\includegraphics[width=2.2in]{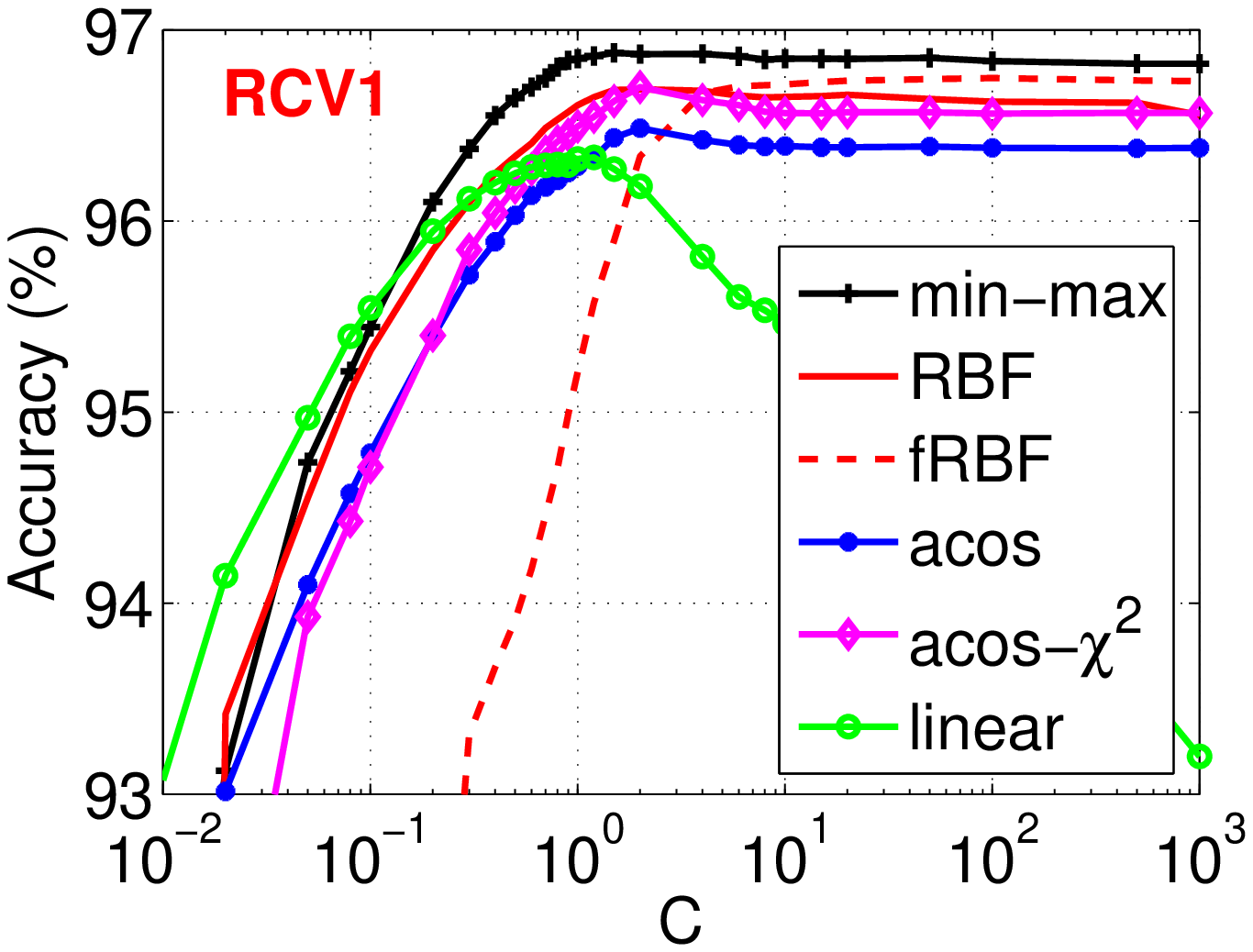}
}

\vspace{-0.041in}

\hspace{-0in}\mbox{
\includegraphics[width=2.2in]{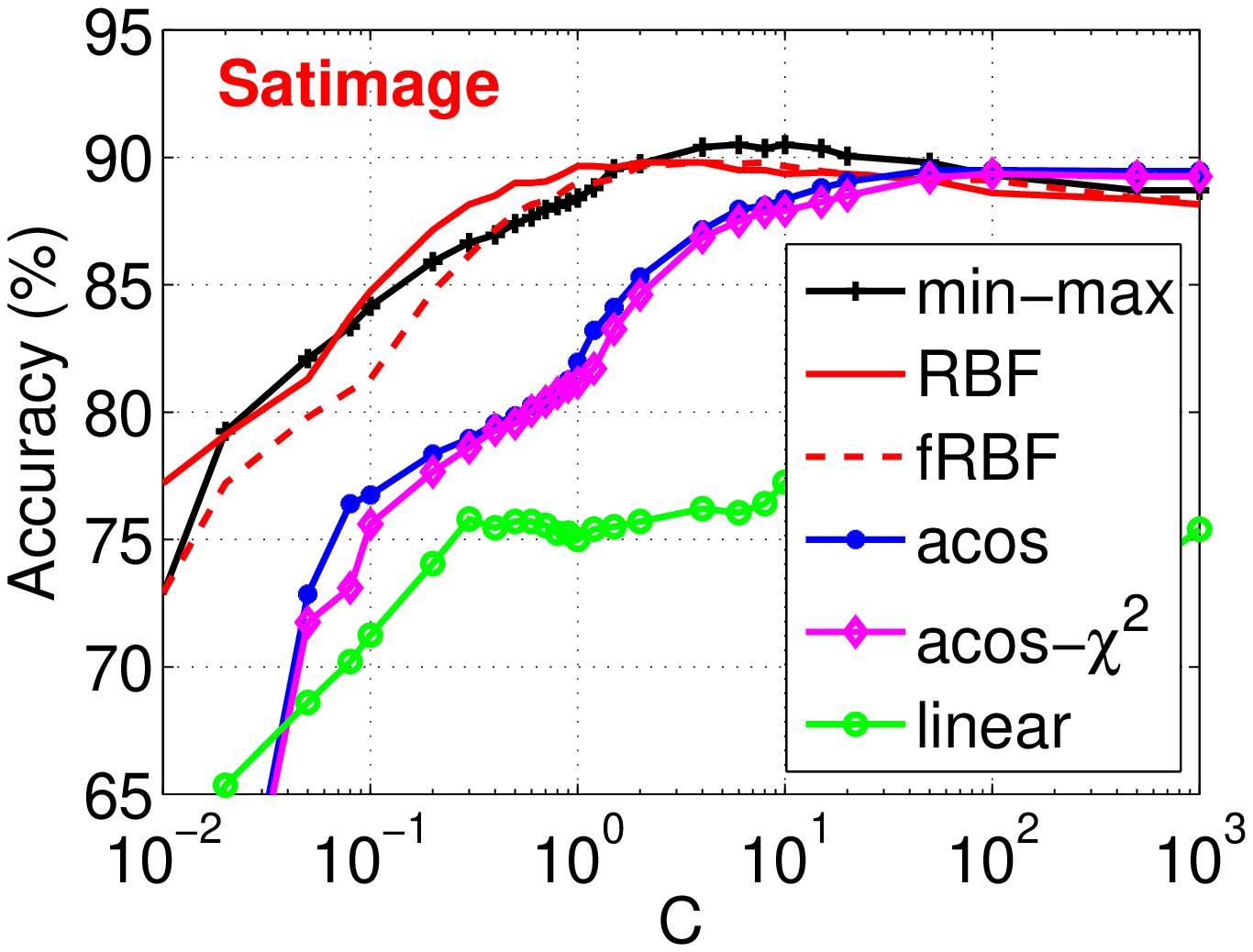}\hspace{0.3in}
\includegraphics[width=2.2in]{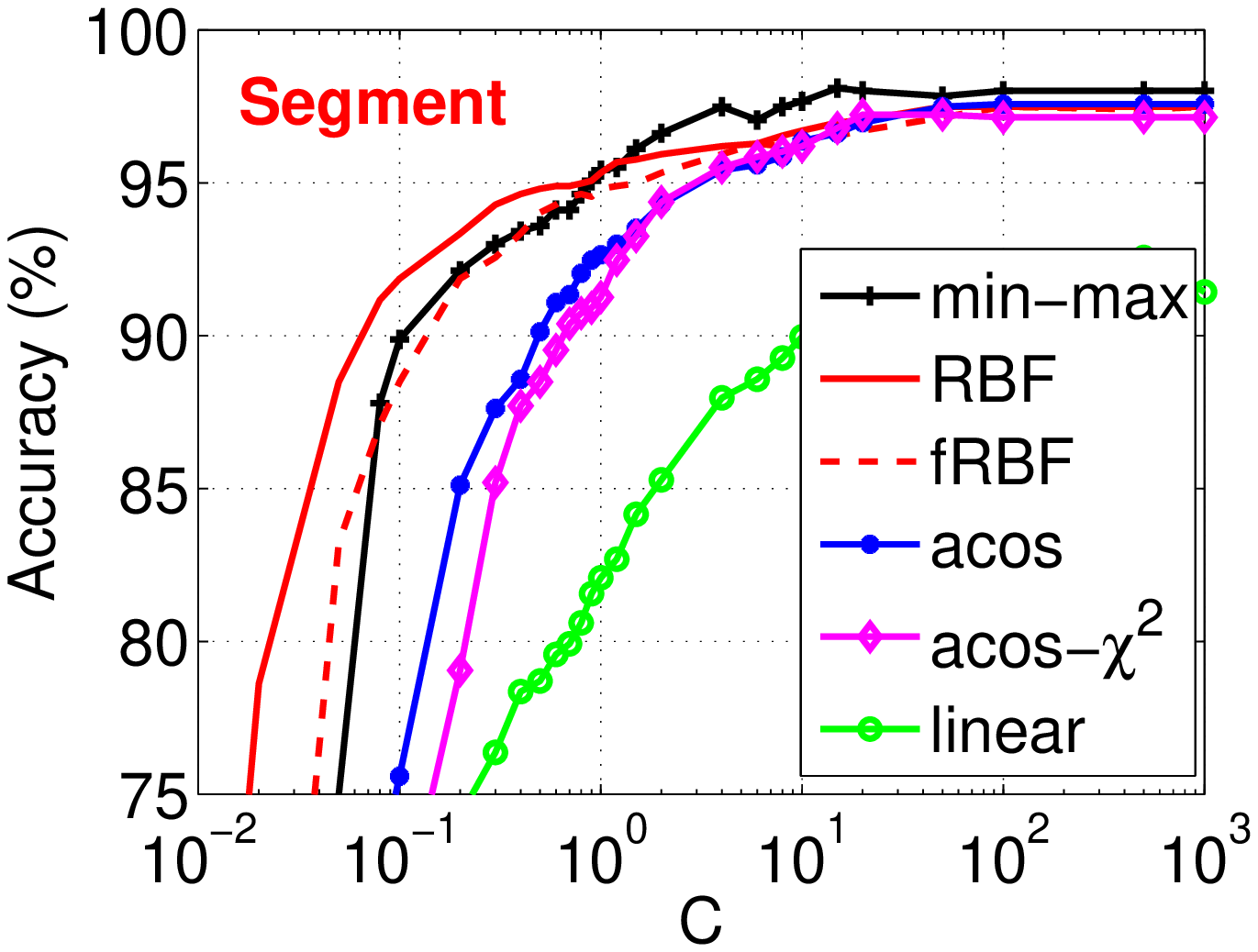}
}

\vspace{-0.041in}

\hspace{-0in}\mbox{
\includegraphics[width=2.2in]{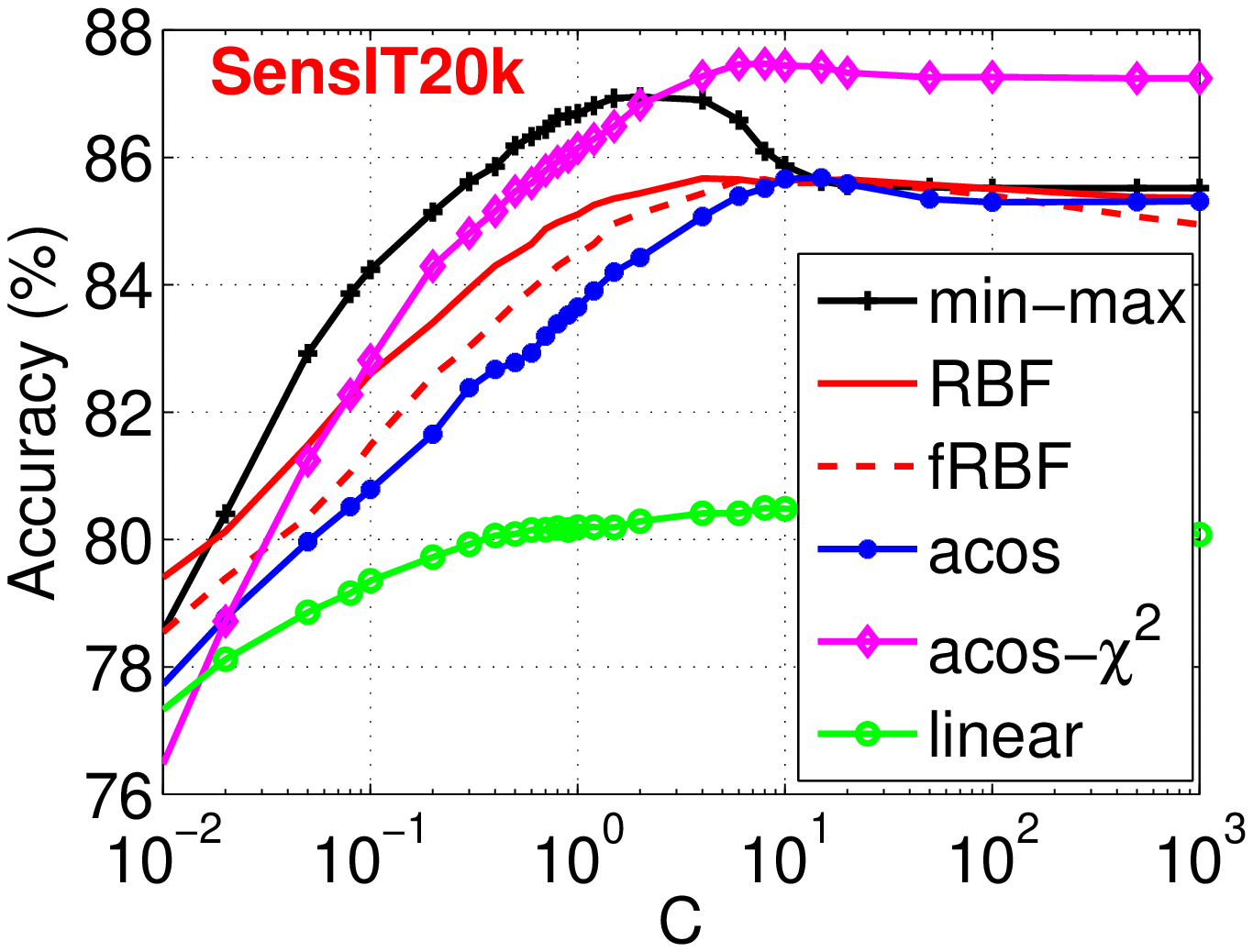}\hspace{0.3in}
\includegraphics[width=2.2in]{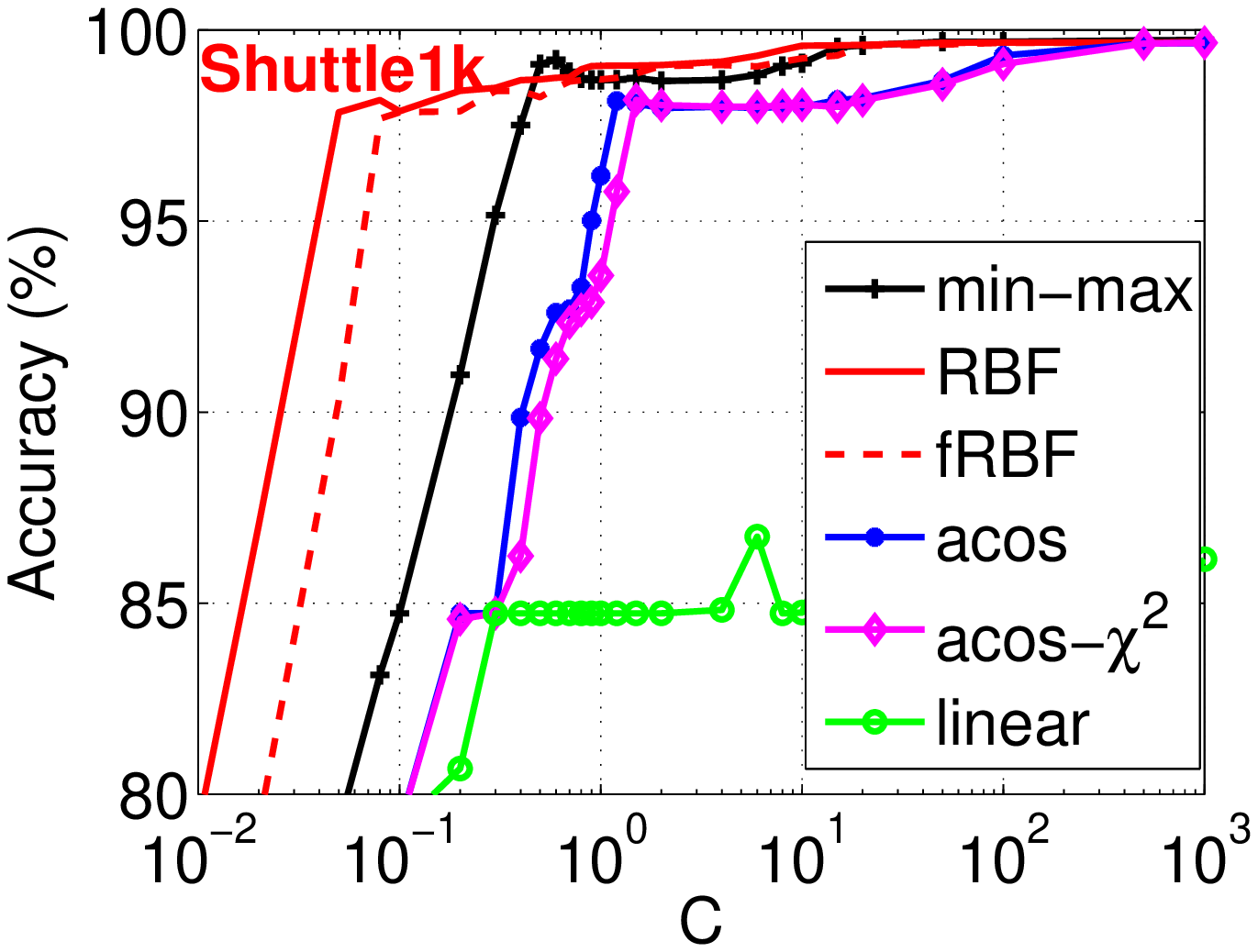}
}

\vspace{-0.041in}

\hspace{-0in}\mbox{
\includegraphics[width=2.2in]{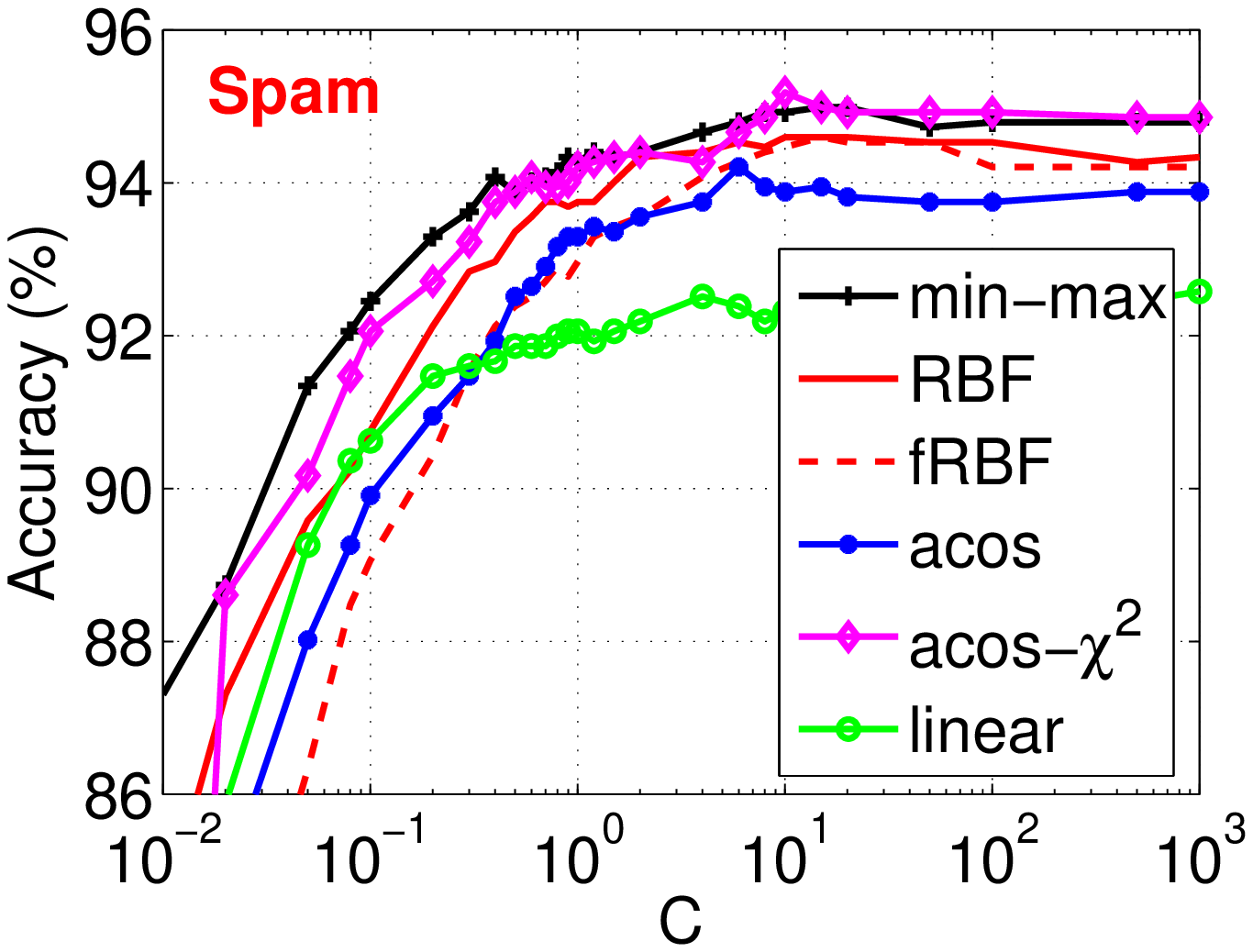}\hspace{0.3in}
\includegraphics[width=2.2in]{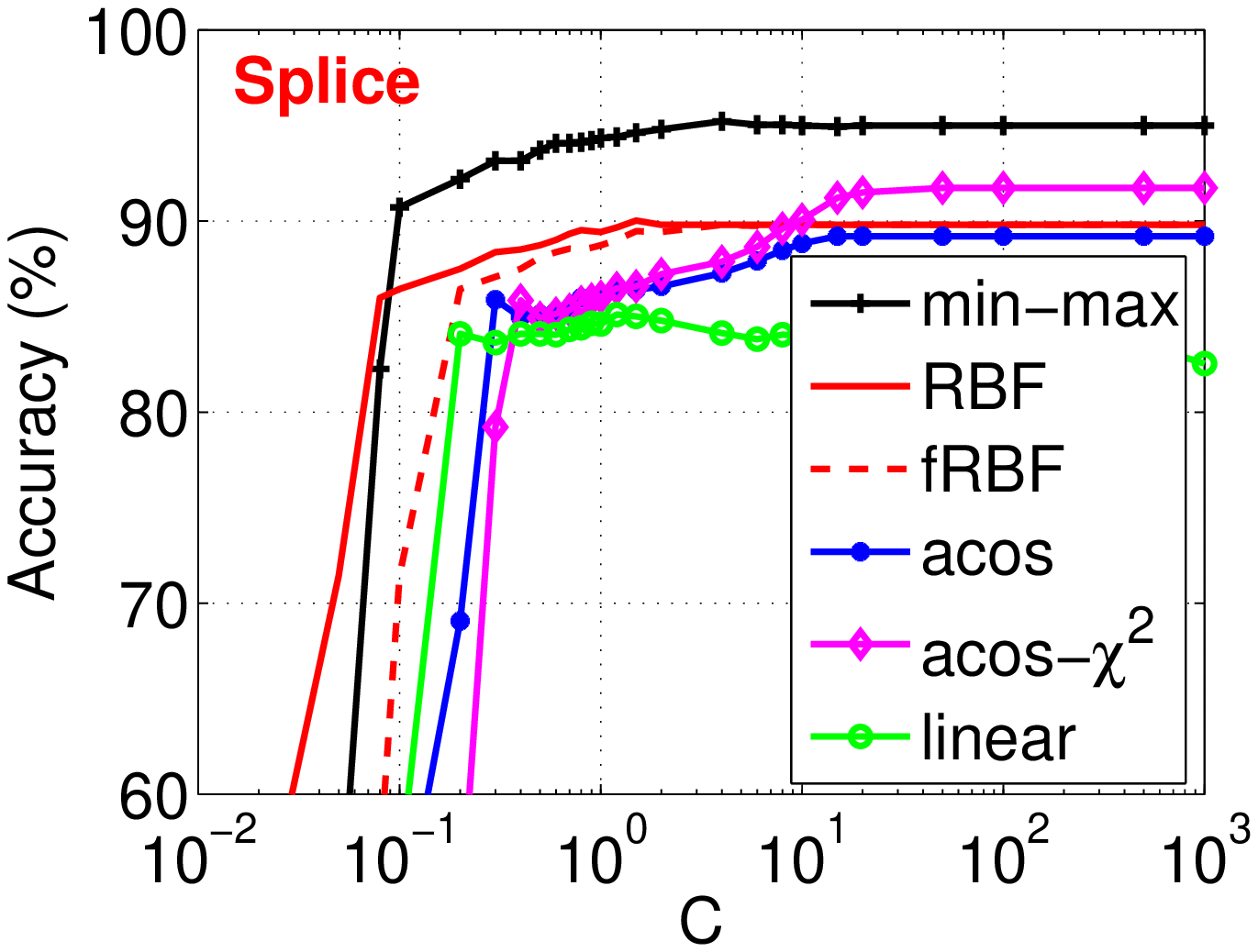}
}

\vspace{-0.041in}

\hspace{-0in}\mbox{
\includegraphics[width=2.2in]{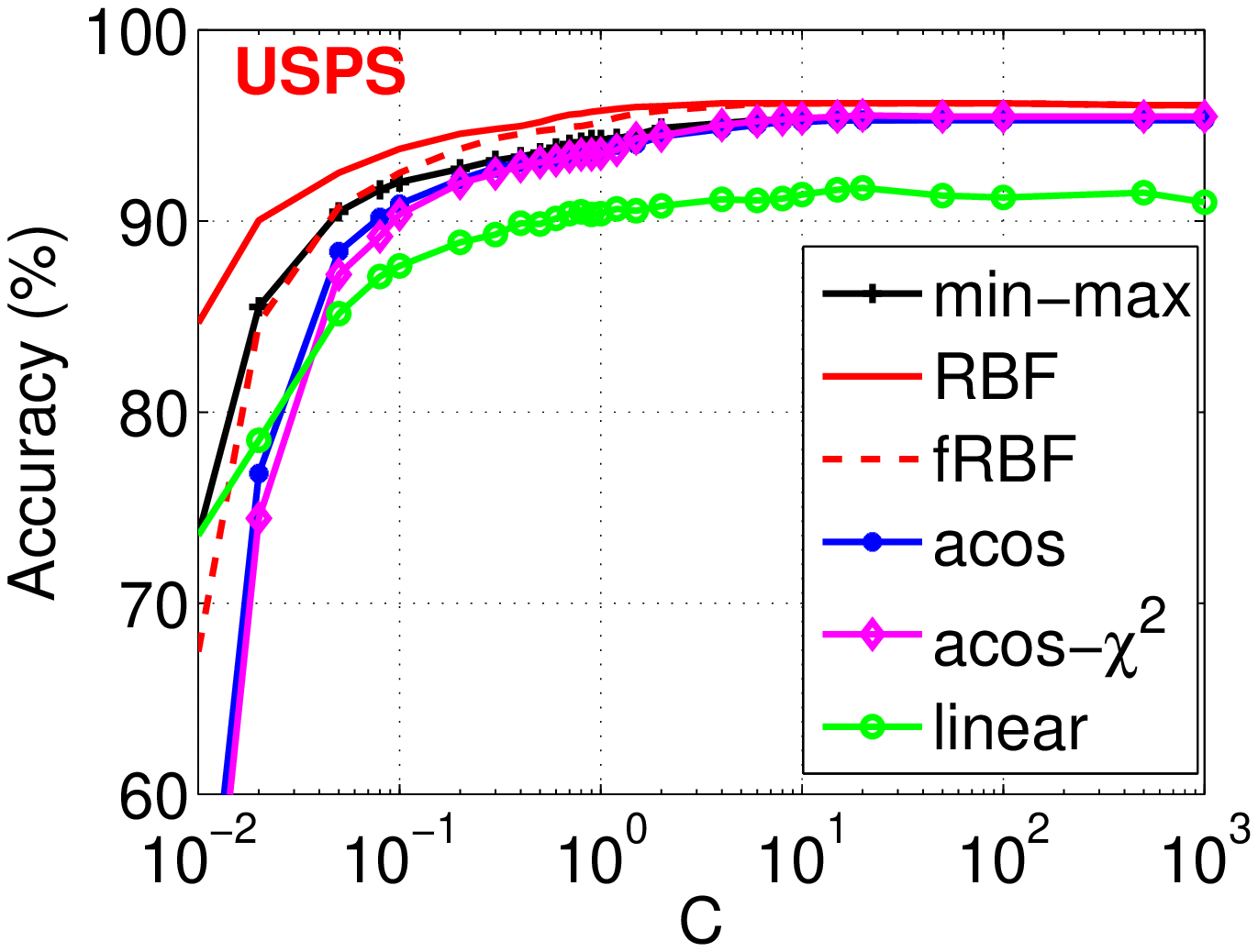}\hspace{0.3in}
\includegraphics[width=2.2in]{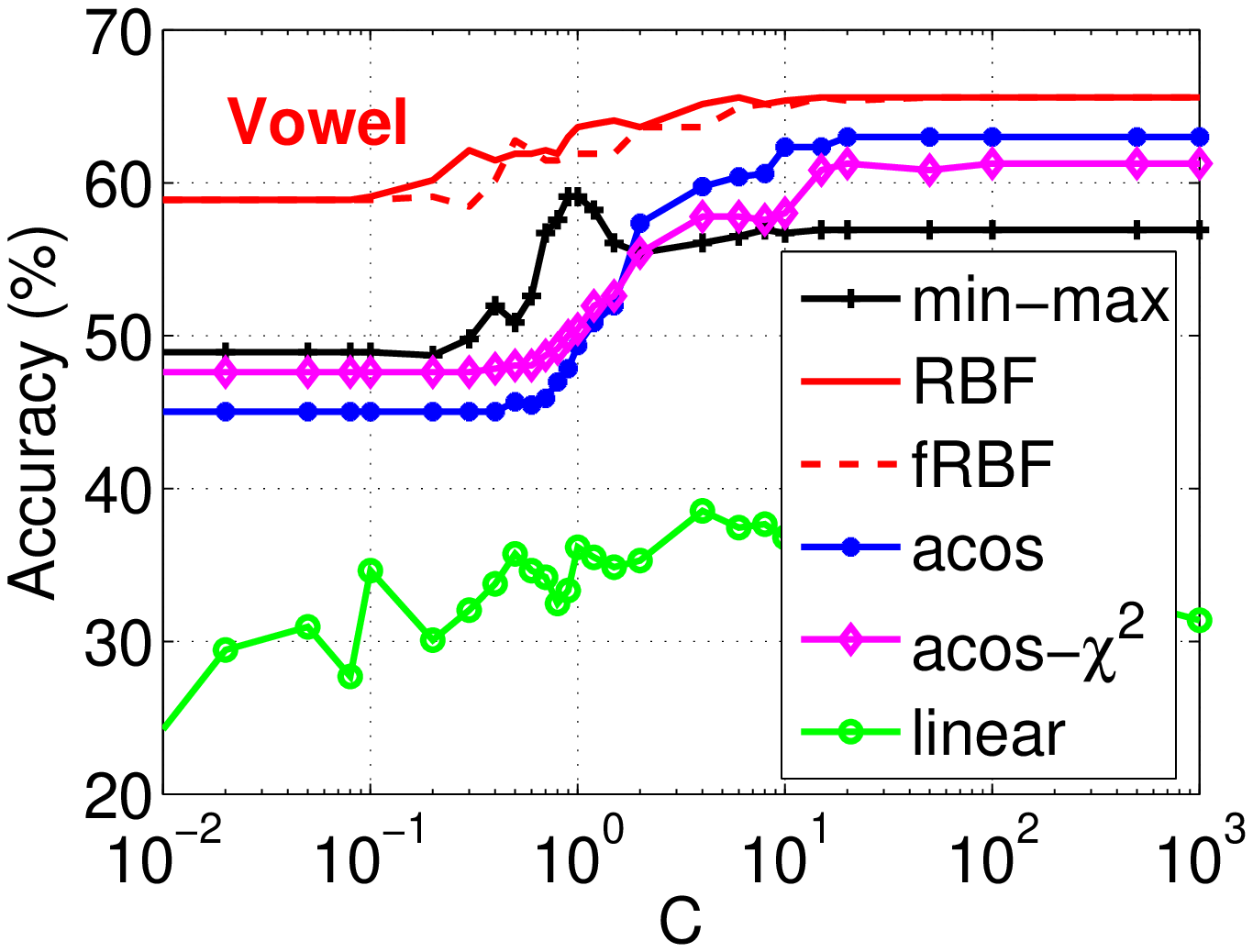}
}

\vspace{-0.041in}

\hspace{-0in}\mbox{
\includegraphics[width=2.2in]{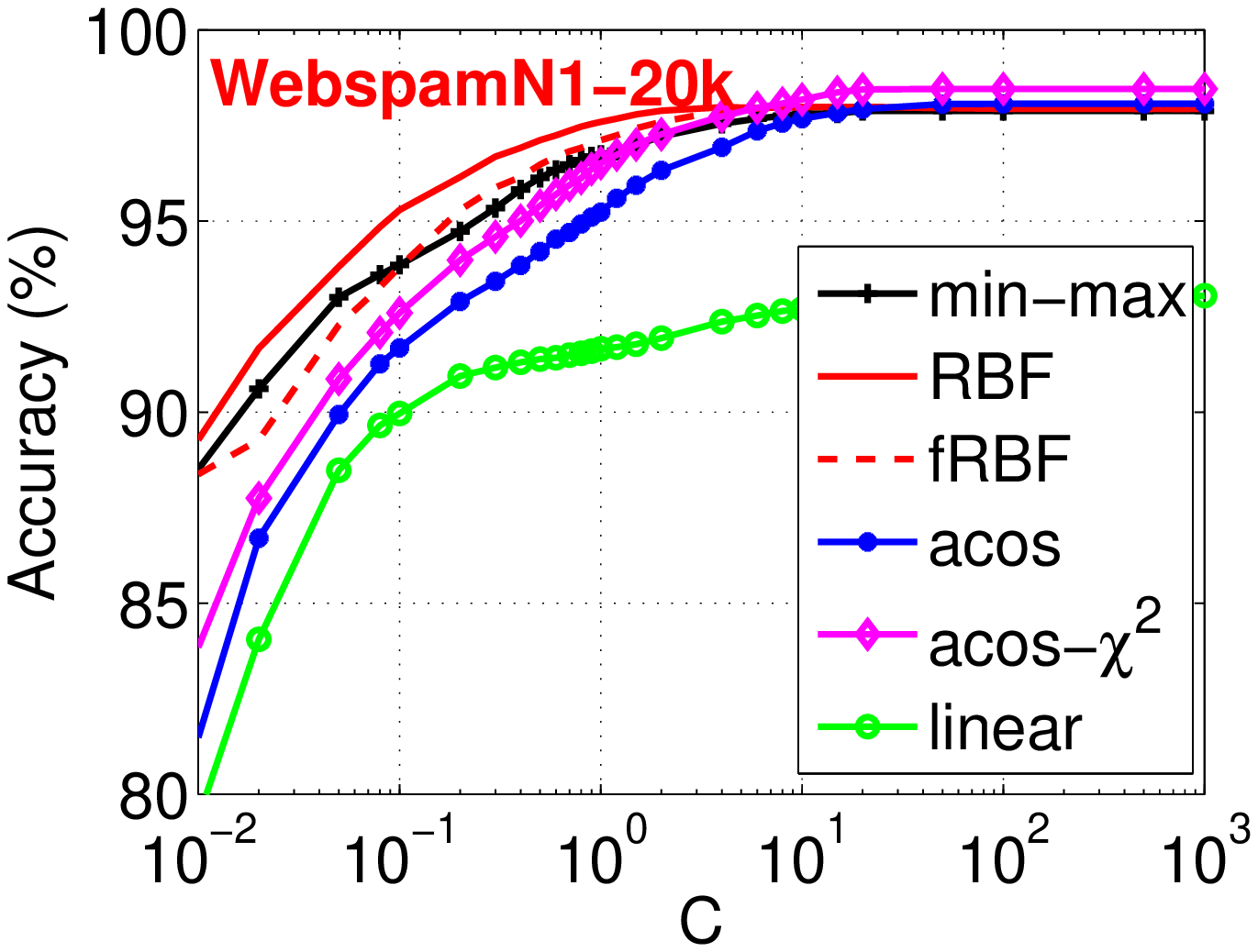}\hspace{0.3in}
\includegraphics[width=2.2in]{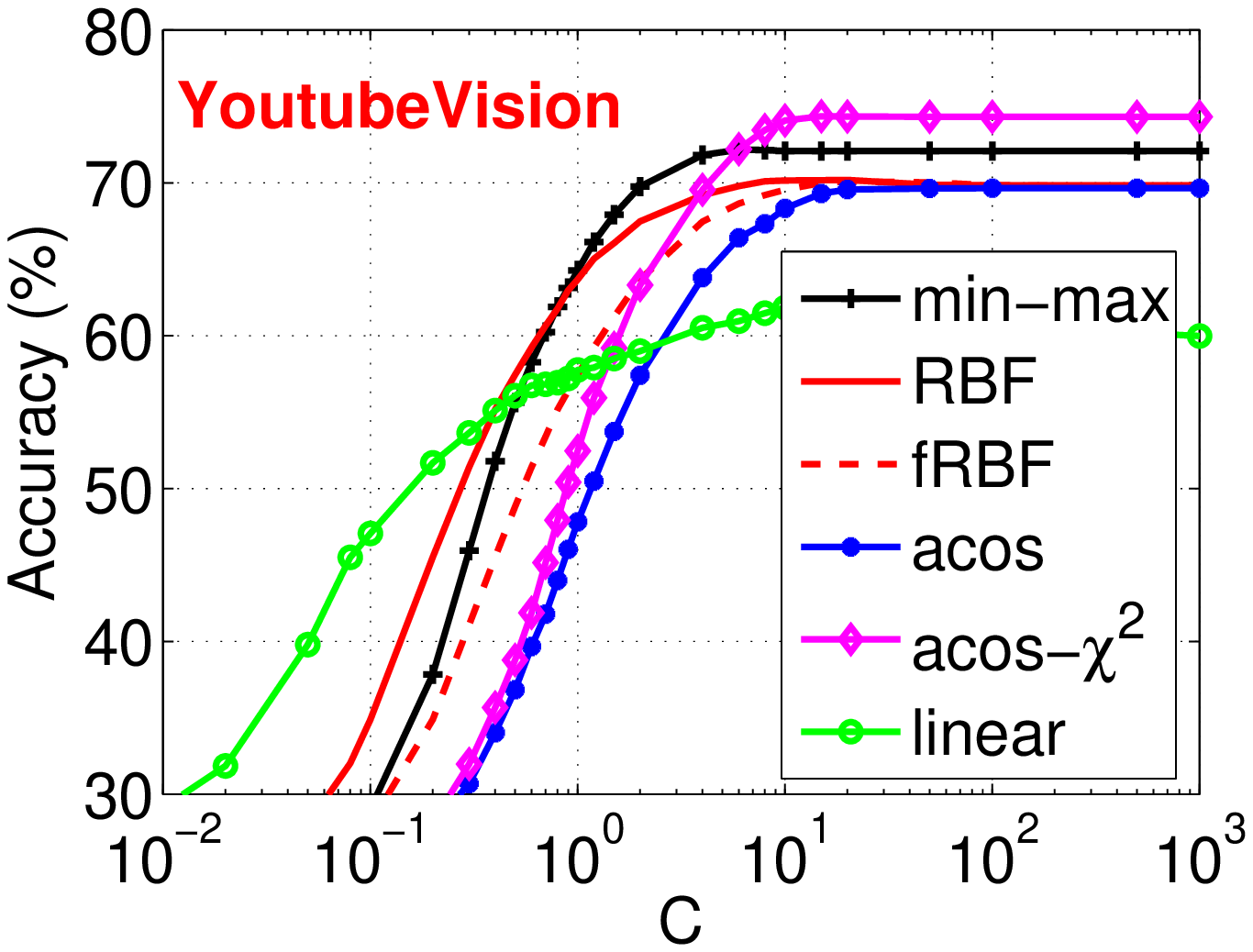}
}

\end{center}
\vspace{-0.4in}
\caption{Test classification accuracies  for  5 nonlinear kernels using $l_2$-regularized  SVM. }\label{fig_KernelSVM3}
\end{figure}

Table~\ref{tab_KernelSVM} summarizes the classification results using 5 different kernel SVMs:  the min-max kernel, the RBF kernel, the fRBF kernel, the acos kernel, and the acos-$\chi^2$ kernel. More detailed results (for all regularization $C$ values) are available in Figures~\ref{fig_KernelSVM1} to~\ref{fig_KernelSVM3}. To ensure repeatability, for all the kernels, we use the LIBSVM pre-computed kernel functionality. This also means we can not (easily) test nonlinear kernels on  larger datasets, for example, ``WebspamN1'' in the last row of Table~\ref{tab_data}.\\

For both RBF and fRBF kernels, we need to choose $\gamma$, the important tuning parameter. For all the datasets, we exhaustively experimented with 58 different values of $\gamma\in\{$0.001, 0.01, 0.1:0.1:2, 2.5, 3:1:20 25:5:50, 60:10:100, 120, 150, 200, 300, 500, 1000$\}$. Here, we adopt the MATLAB  notation that (e.g.,) 3:1:20 means all the numbers from 3 to 20 spaced at 1. Basically, Table~\ref{tab_KernelSVM} reports the best RBF/fRBF results among all $\gamma$ and $C$ values in our experiments.\\

\begin{table}[h!]
\caption{Classification accuracies (in \%) using 5 different kernels. We use  LIBSVM's ``pre-computed'' kernel functionality for training nonlinear $l_2$-regularized  kernel SVMs (with a  tuning parameter $C$). The reported test classification accuracies  are the best accuracies from a wide range of $C$ values; see Figures~\ref{fig_KernelSVM1} to~\ref{fig_KernelSVM3} for more details. In particular, for the RBF kernel and the fRBF kernel, we experimented with 58 different $\gamma$ values ranging from 0.001 to 1000 and the reported accuracies are the best values among all $\gamma$ (and all $C$). See Table~\ref{tab_data} for more information on the datasets. The numbers in parentheses are the best $\gamma$ values for RBF and fRBF. }
\begin{center}{\vspace{-0in}
{\begin{tabular}{l c l l c c c c}
\hline \hline
Dataset     &min-max &RBF &fRBF  &acos &acos-$\chi^2$ \\
\hline
Covertype10k    &80.4 &80.1 (120)   &80.1 (100)   &{\bf81.9}   &81.6 \\
Covertype20k    &83.3  &83.8 (150)   &83.8 (150)   &{\bf85.3}  &85.0 \\
IJCNN5k  &94.4  &{\bf98.0} (45)   &{\bf98.0} (40)  &96.9    &96.6 \\
IJCNN10k   &95.7   &{\bf98.3} (60)   &98.2 (50)   &97.5    &97.4 \\
Isolet   &96.4   &96.8 (6)   &{\bf96.9} (11)   &96.5   &96.1 \\
Letter   &96.2   &{\bf97.6} (100)   &{\bf97.6} (100)  &97.0   &97.0 \\
Letter4k    &91.4   &94.0 (40)   &{\bf94.1} (50)  & 93.3  & 93.3 \\
M-Basic      &96.2   &{\bf97.2} (5)  &{\bf97.2} (5)  &95.7   &95.8 \\
M-Image  &{\bf80.8}   &77.8 (16)   &77.8 (16) &76.2   &75.2\\
MNIST10k   &95.7   &96.8 (5)  &{\bf96.9} (5)  &95.2   &95.2 \\
M-Noise1 &{\bf71.4}   &66.8 (10)   &66.8 (10)   &65.0  & 64.0 \\
M-Noise2  &{\bf72.4}  &69.2 (11)   &69.2 (11)  &66.9   &65.7 \\
M-Noise3  &{\bf73.6}  &71.7 (11)   &71.7 (11)  &69.0   &68.0 \\
M-Noise4  &{\bf76.1}  &75.3 (14)   &75.3 (14)  &73.1   &71.1 \\
M-Noise5   &{\bf79.0}  &78.7 (12)  &78.6 (11)   &76.6   &74.9 \\
M-Noise6 &84.2  &{\bf85.3} (15)  &{\bf85.3} (15)   &83.9   &82.8\\
M-Rand  &84.2   &{\bf85.4} (12)  &{\bf85.4} (12)   &83.5   &82.3 \\
M-Rotate   &84.8   &{\bf89.7} (5)   &{\bf89.7} (5)   &84.5  &84.6\\
M-RotImg  &41.0   &{\bf45.8} (18)   &{\bf45.8} (18)   &41.5  &39.3 \\
Optdigits  &97.7   &{\bf98.7} (8)   &{\bf98.7} (8)   &97.7  &97.5 \\
Pendigits  &97.9   &{\bf98.7} (13)  &{\bf98.7} (11) & 98.3  &98.1 \\
Phoneme  &{\bf92.5}   &92.4 (10)   &{\bf92.5} (9)   &92.2   &90.2 \\
Protein  &{\bf72.4}   &70.3 (4)  &70.2 (4)   &69.2   &70.5 \\
RCV1  &{\bf96.9}   &96.7 (1.7)  &96.7 (0.3)  &96.5   &96.7 \\
Satimage &{\bf90.5}  &89.8 (150)   &89.8 (150)  &89.5   &89.4\\
Segment  &{\bf98.1}  &97.5 (15)   &97.5 (15)  &97.6   &97.2 \\
SensIT20k  &86.9   &85.7 (4)   &85.7 (4)  &85.7  &{\bf87.5} \\
Shuttle1k  &{\bf99.7}   &{\bf99.7} (10)   &{\bf99.7} (15)  &{\bf99.7}   &{\bf99.7} \\
Spam  & 95.0 & 94.6 (1.2)  &94.6 (1.7)  &94.2   &{\bf95.2} \\
Splice  &{\bf95.2}  &90.0 (15)  &89.8 (16)  & 89.2   &91.7  \\
USPS   &95.3  &{\bf96.2} (11)  &{\bf96.2} (11)   &95.3   &95.5  \\
Vowel  &59.1   &{\bf65.6} (20)   &{\bf65.6} (20)  &63.0    &61.3 \\
WebspamN1-20k   &97.9   &98.0 (35)  &98.0 (35)   &98.1   &{\bf98.5}   \\
YoutubeVision   &72.2   &70.2 (3)   &70.1 (4)   &69.6  &{\bf74.4} \\
\hline\hline
\end{tabular}}
}
\end{center}\label{tab_KernelSVM}

\end{table}

Table~\ref{tab_KernelSVM} shows that RBF kernel and fRBF kernel perform very similarly. Interestingly, even with the best tuning parameters, RBF/fRBF kernels do not always achieve the highest classification accuracies. In fact, for about $40\%$ of the datasets, the min-max kernel (which is tuning-free) achieves the highest accuracies. It is  also interesting that the acos kernel and the acos-$\chi^2$ kernel perform reasonably well compared to the RBF/fRBF kernels.\\

Overall, it appears that the RBF/fRBF kernels tend to perform well in very low dimensional datasets. One interesting future study is to develop new kernel functions based on the min-max kernel, the acos kernel, or the acos-$\chi^2$ kernel, to improve the accuracies. The new kernels could be the original kernels equipped with a tuning parameter via a nonlinear transformation. One challenge is that, for any new (and tunable) kernel, we must also be able to find a randomization algorithm to linearize the kernel; otherwise, it would not be too meaningful for large-scale applications.\vspace{-0.15in}

\newpage\clearpage

\clearpage\newpage

\section{Linearization of Nonlinear Kernels}

It is known that a straightforward  implementation of nonlinear kernels can be difficult for large datasets~\cite{Book:Bottou_07}.  As mentioned earlier, for  a small dataset with merely $60,000$ data points, the $60,000 \times 60,000$ kernel matrix  has $3.6\times 10^9$ entries.  In practice, being able to linearize nonlinear kernels becomes very beneficial, as that would allow us to easily apply efficient linear algorithms in particular online learning~\cite{URL:Bottou_SGD}. Randomization is a popular tool for kernel linearization. \\

Since LIBSVM did not implement most of the nonlinear kernels in our study, we simply used the LIBSVM pre-computed kernel functionality in our experimental study as reported in Table~\ref{tab_KernelSVM}. While this strategy ensures repeatability, it requires very large memory.

In the introduction, we have explained how to linearize these 5 types of nonlinear kernels. From practitioner's perspective, while results in Table~\ref{tab_KernelSVM} are informative, they are not sufficient for guiding  the choice of kernels. For example, as we will show, for some datasets, even though the RBF/fRBF kernels perform better than the min-max kernel in the kernel SVM experiments, their linearization algorithms require many more samples (i.e., large $k$) to reach the same accuracy as the linearization method  (i.e., 0-bit CWS) for  the min-max kernel.

\subsection{RBF Kernel versus  fRBF Kernel}

We have explained how to linearize both the RBF kernel and the fRBF kernel in Introduction. For two normalized vectors $u,v\in\mathbb{R}^D$, we generate i.i.d. samples $r_i \sim N(0,1)$ and independent $w\sim uniform(0,2\pi)$. Let $x =\sum_{i=1}^D u_i r_i$ and $y = \sum_{i=1}^D vi r_i$. Then we have
\begin{align}\notag
&E\left(\cos(\sqrt{\gamma}x +w)\cos(\sqrt{\gamma}y +w)\right) = RBF(u,v;\gamma)\\\notag
&E\left(\cos(\sqrt{\gamma}x)\cos(\sqrt{\gamma}y)\right) = fRBF(u,v;\gamma)
\end{align}
In order to approximate the expectations with sufficient accuracies, we need to generate the samples $(x,y)$ many (say $k$) times. Typically $k$ has to be  large. In our experiments, even though we use $k$ as large as 4096, it looks we will have to further increase $k$  in order to reach the accuracy of the original RBF/fRBF kernels (as in Table~\ref{tab_KernelSVM}).\\

\begin{figure}[h!]
\begin{center}

\hspace{-0in}\mbox{
\includegraphics[width=2.2in]{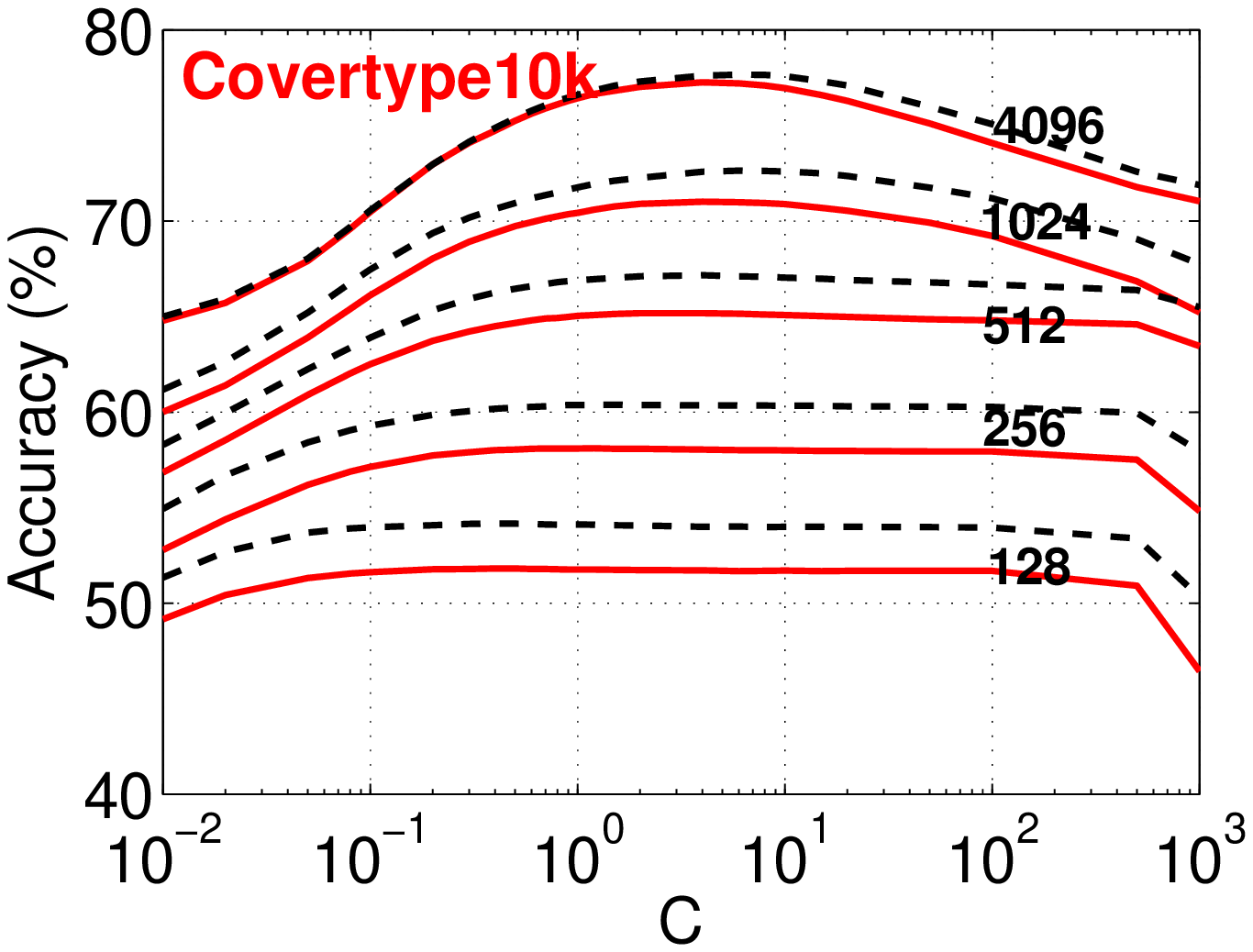}\hspace{0.3in}
\includegraphics[width=2.2in]{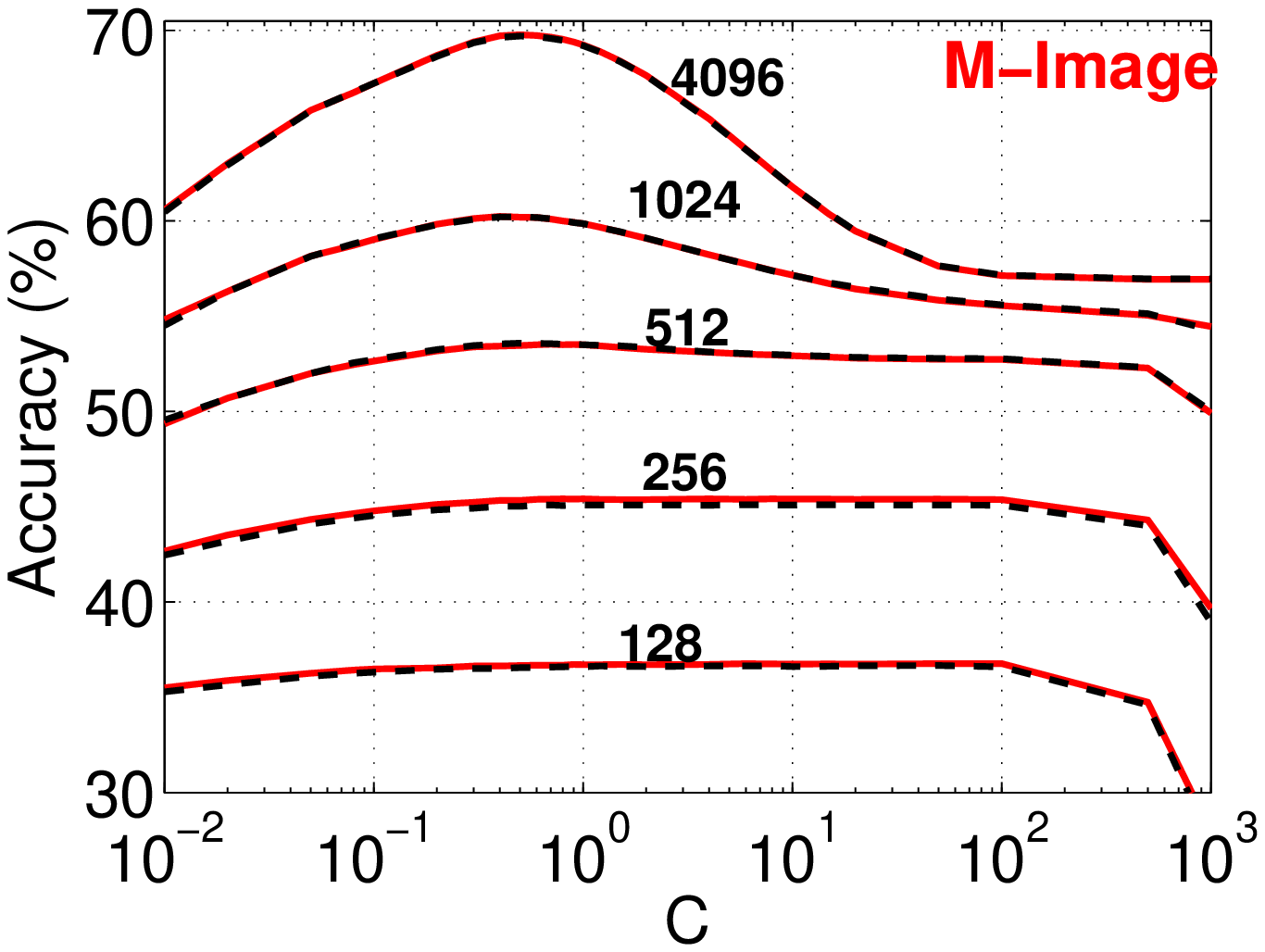}
}

\vspace{-0.038in}

\hspace{-0in}\mbox{
\includegraphics[width=2.2in]{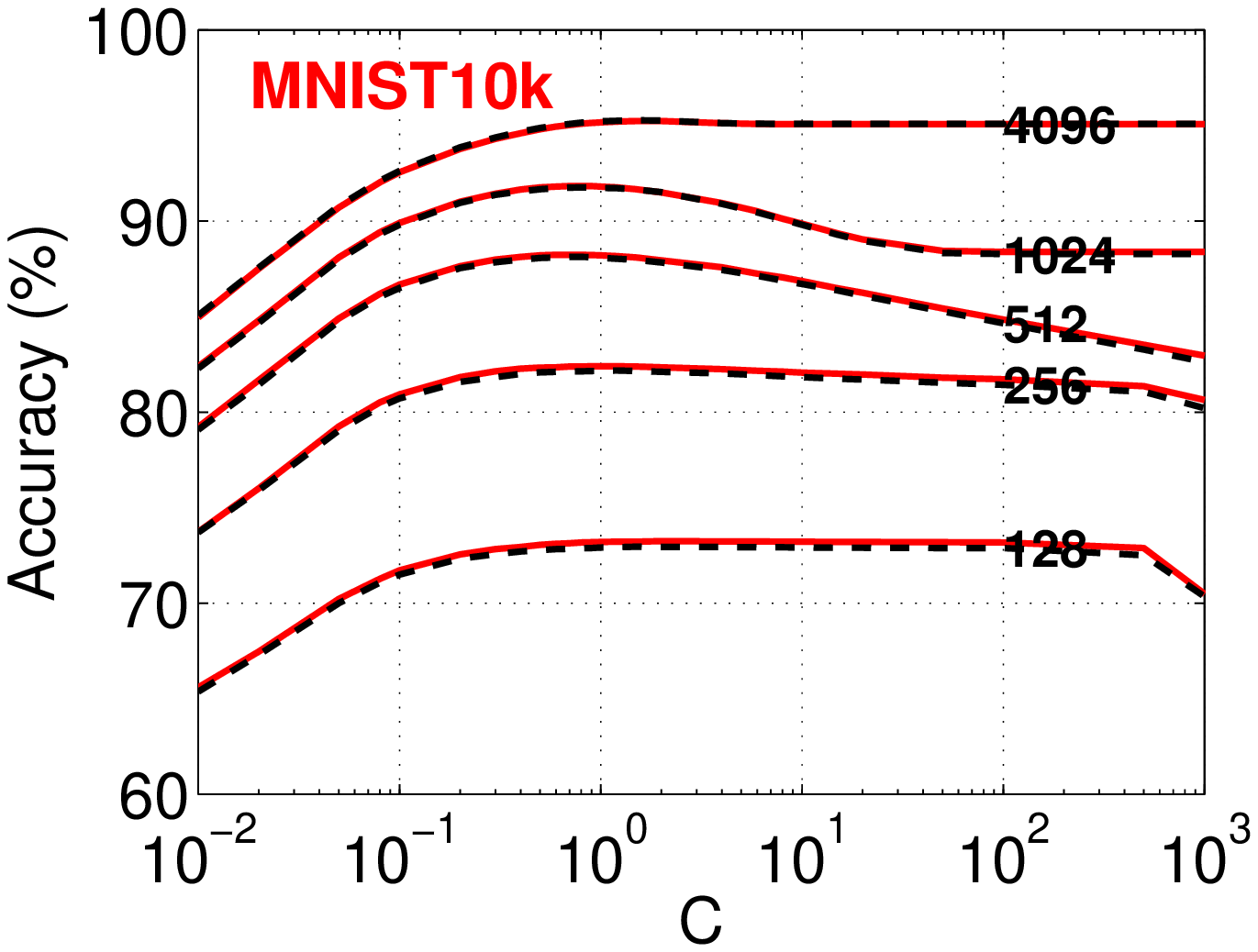}\hspace{0.3in}

\includegraphics[width=2.2in]{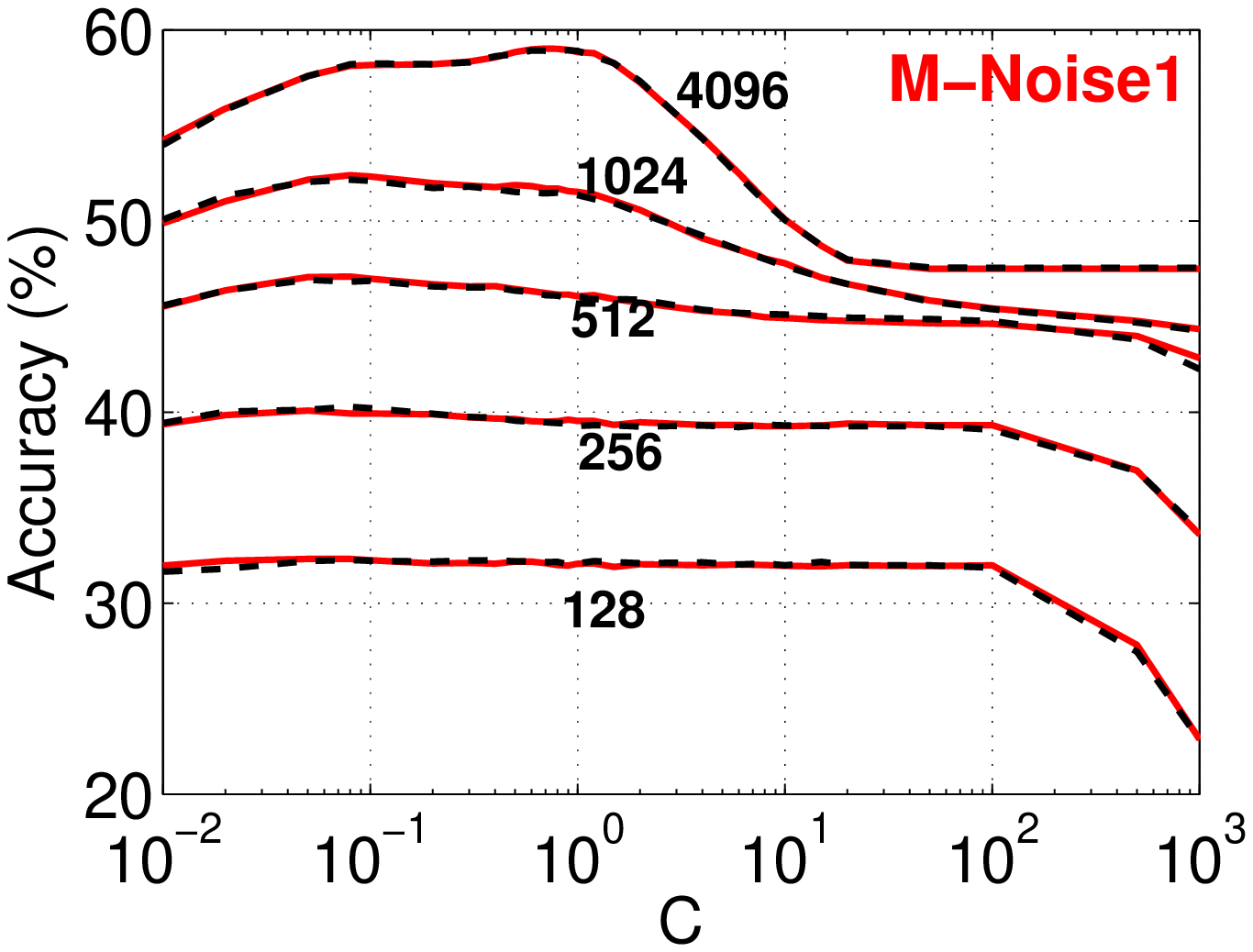}

}

\vspace{-0.038in}

\hspace{-0in}\mbox{
\includegraphics[width=2.2in]{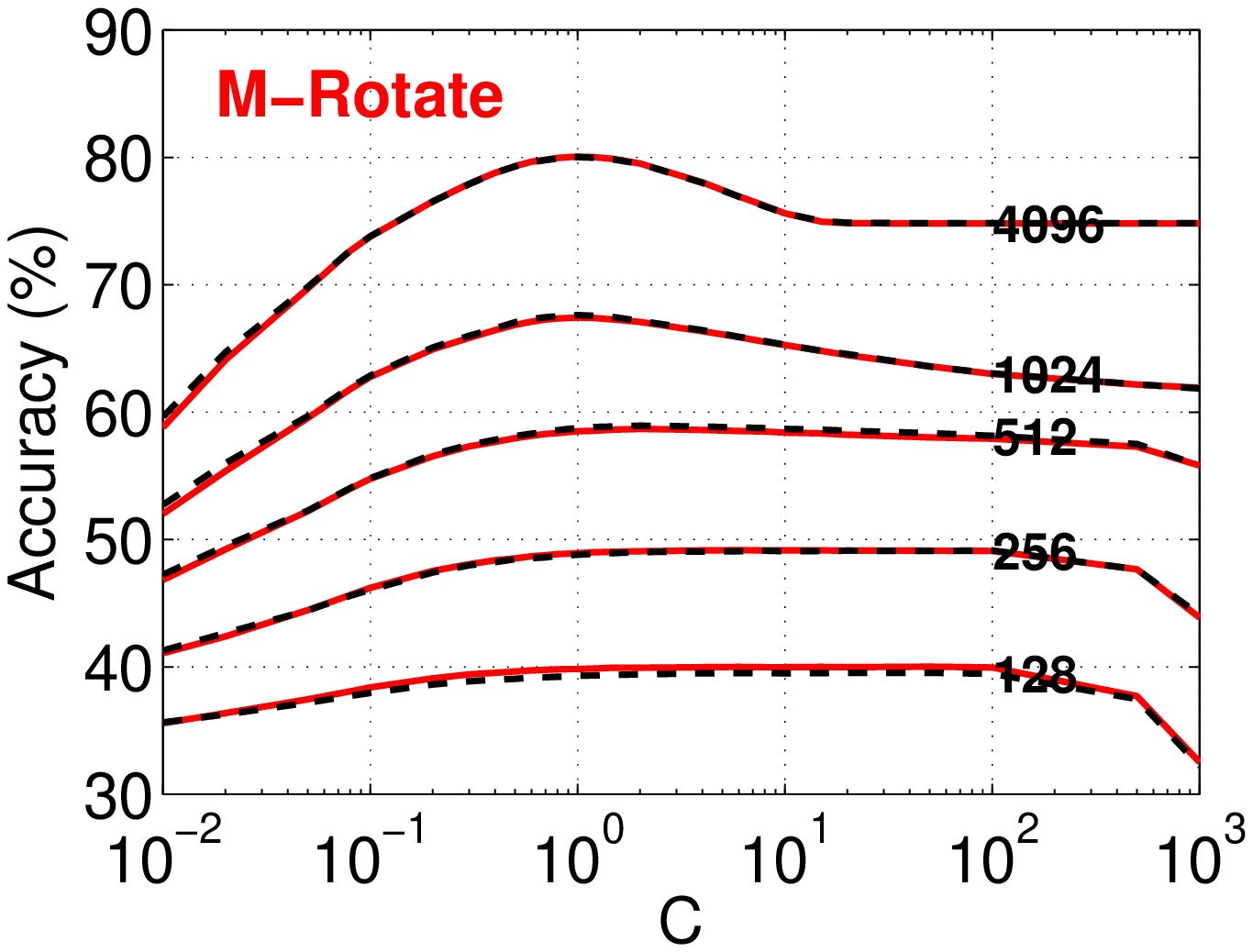}\hspace{0.3in}
\includegraphics[width=2.2in]{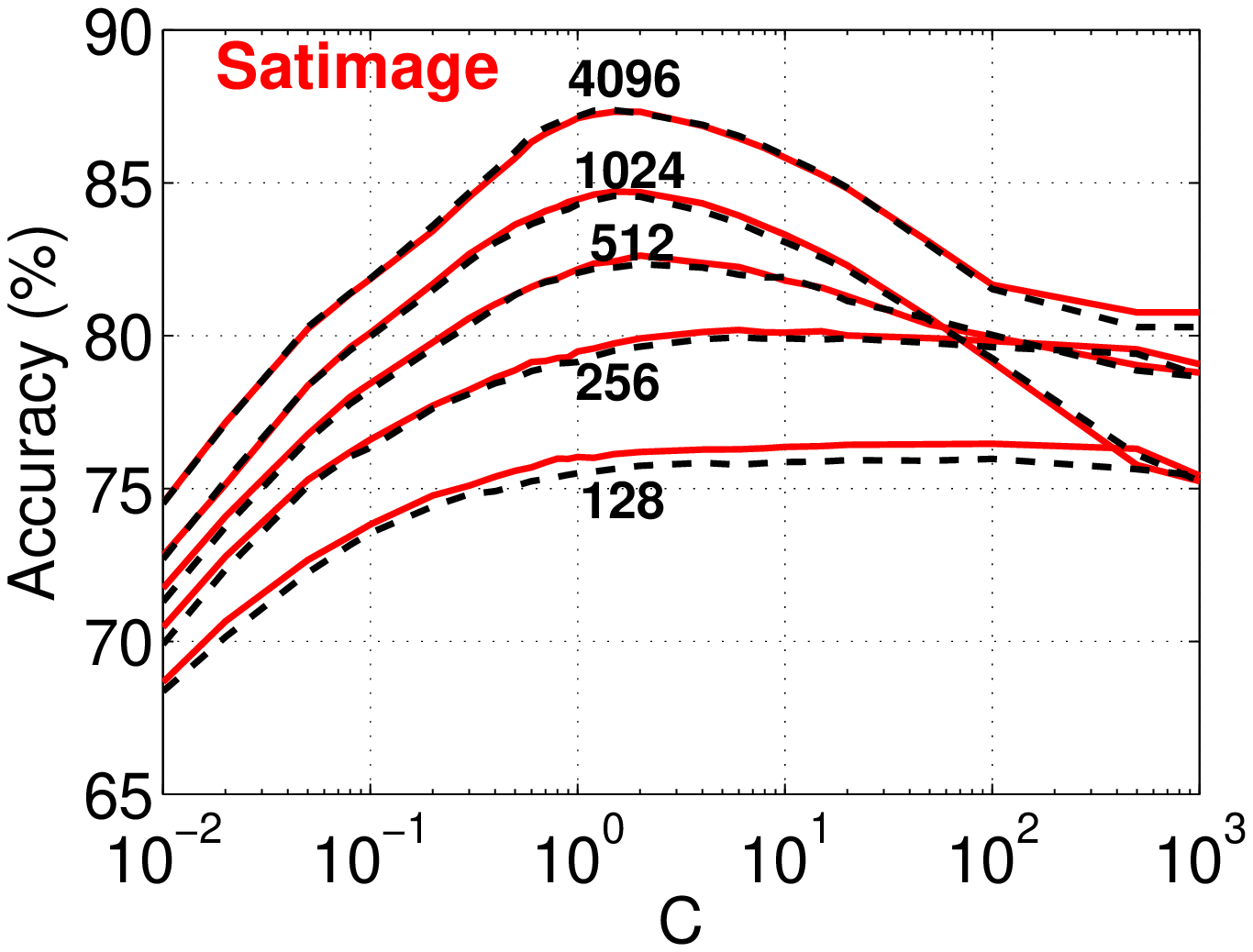}

}

\vspace{-0.038in}

\hspace{-0in}\mbox{
\includegraphics[width=2.2in]{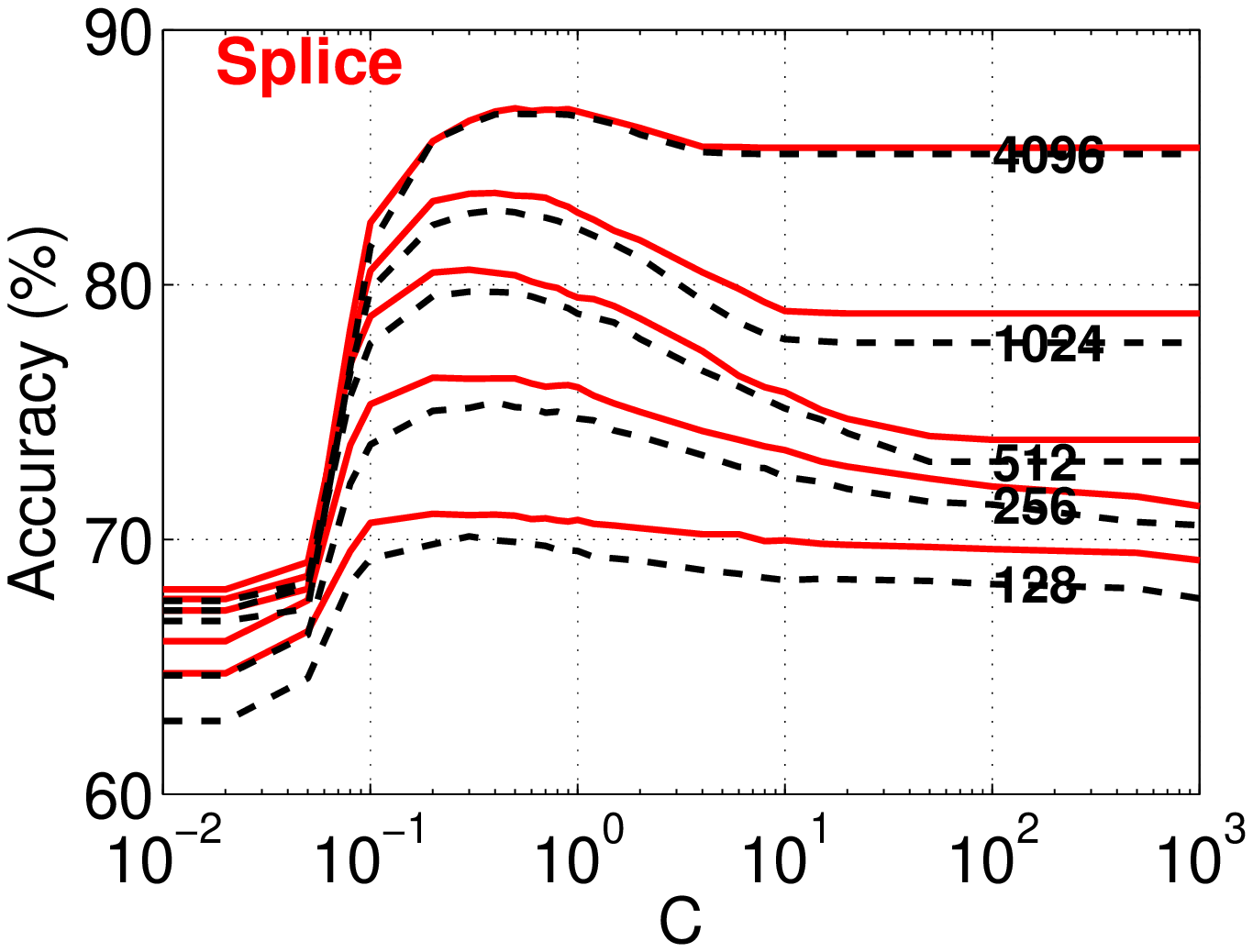}\hspace{0.3in}
\includegraphics[width=2.2in]{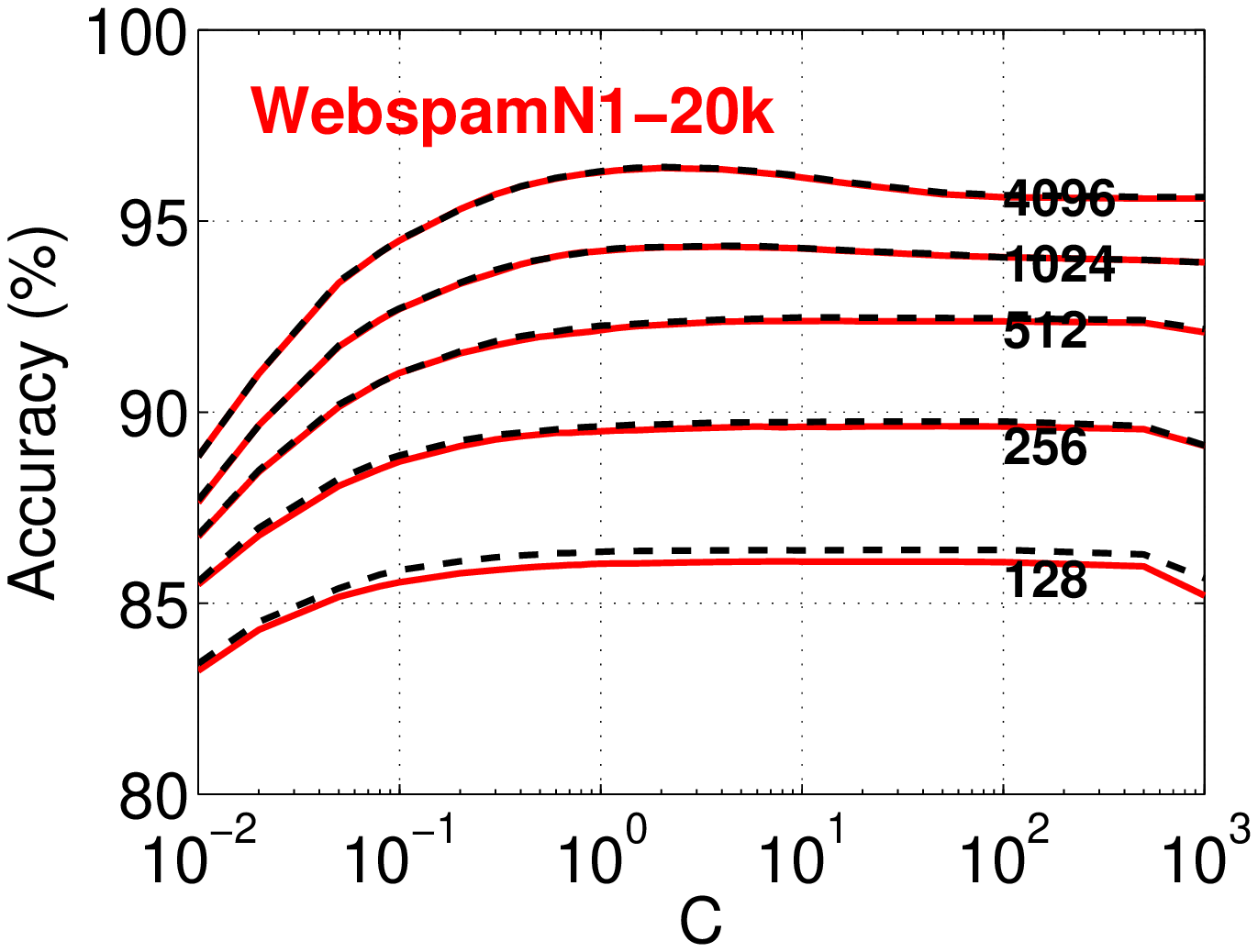}
}

\vspace{-0.038in}

\hspace{-0in}\mbox{
\includegraphics[width=2.2in]{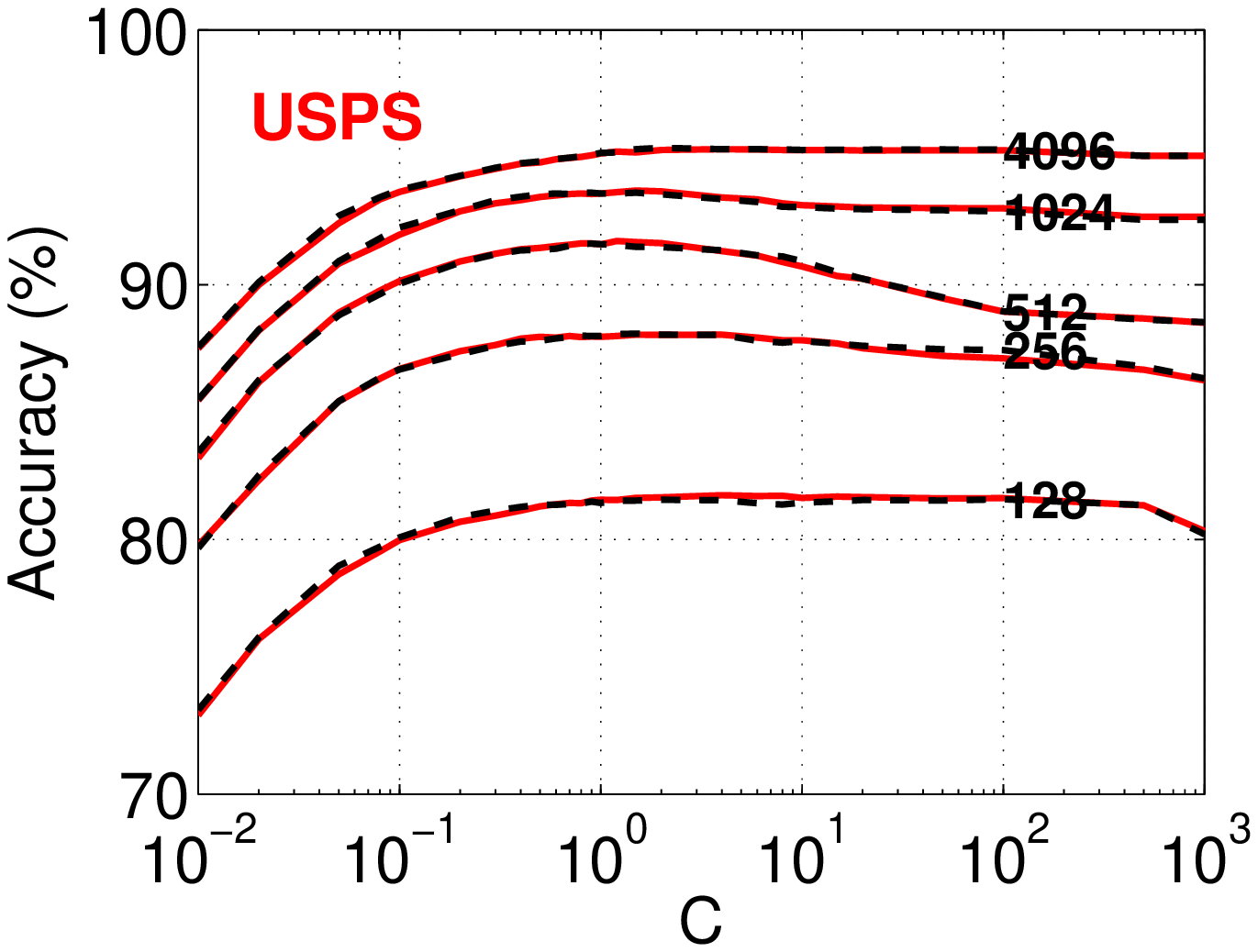}\hspace{0.3in}
\includegraphics[width=2.2in]{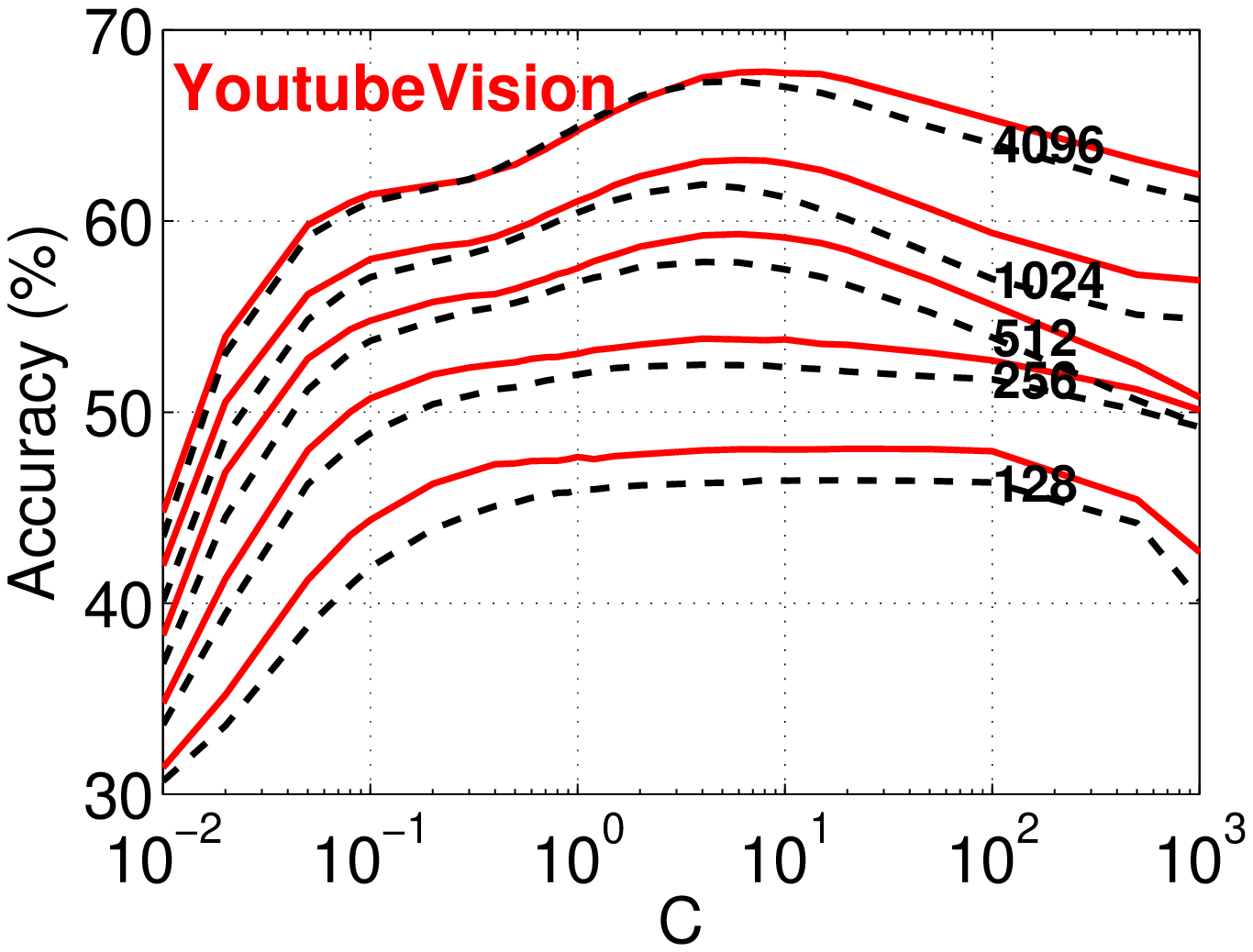}
}

\end{center}
\vspace{-0.3in}
\caption{Classification accuracies of the linearized RBF (solid curves) and linearized fRBF (dashed curves) kernels, using LIBLINEAR. We report the results on 5 different $k$ (sample size) values: 128, 256, 512, 1024, 4096. For most datasets, both RBF and fRBF perform almost identically. }\label{fig_RBF/fRBF}
\end{figure}

Figure~\ref{fig_RBF/fRBF} reports the linear SVM experiments on the linearized data for 10 datasets, for $k$ as large as 4096.  We can see that, for most datasets, the linearized RBF and linearized fRBF kernels perform almost identically. For a few datasets, there are visible discrepancies but the differences are small. We repeat the experiments 10 times and the reported results are the averages.  Note that we always use the best $\gamma$ values as provided in Table~\ref{tab_KernelSVM}.\\

Together with the results on Table~\ref{tab_KernelSVM}, the results as shown in Figure~\ref{fig_RBF/fRBF} allow us to conclude that the fRBF kernel can replace the RBF kernel and we can simplify the linearization algorithm by removing the additional random variable $w$.

\clearpage\newpage

\subsection{Min-max Kernel versus  RBF/fRBF Kernels}

Table~\ref{tab_KernelSVM} has shown that for  quite a few datasets, the RBF/fRBF kernels outperform the min-max kernel. Now we compare their corresponding linearization algorithms. We adopt the 0-bit CWS~\cite{Proc:Li_KDD15} strategy and use at most 8 bits for storing each sample. See the Introduction and Appendix~\ref{app_CWS} for more details on consistent weighted sampling (CWS).\\

\begin{figure}[h!]
\begin{center}

\hspace{-0in}\mbox{
\includegraphics[width=2.2in]{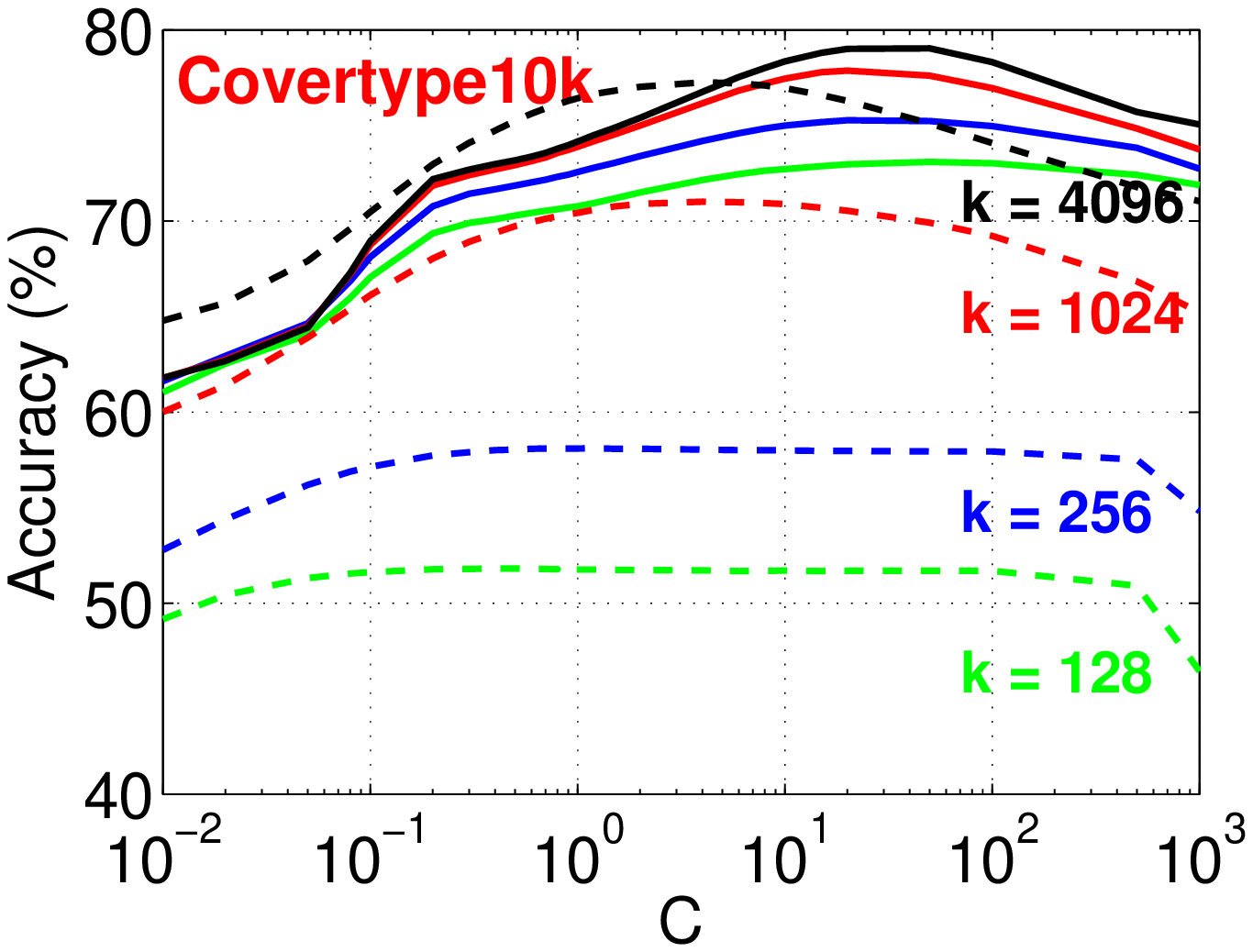}\hspace{0.3in}
\includegraphics[width=2.2in]{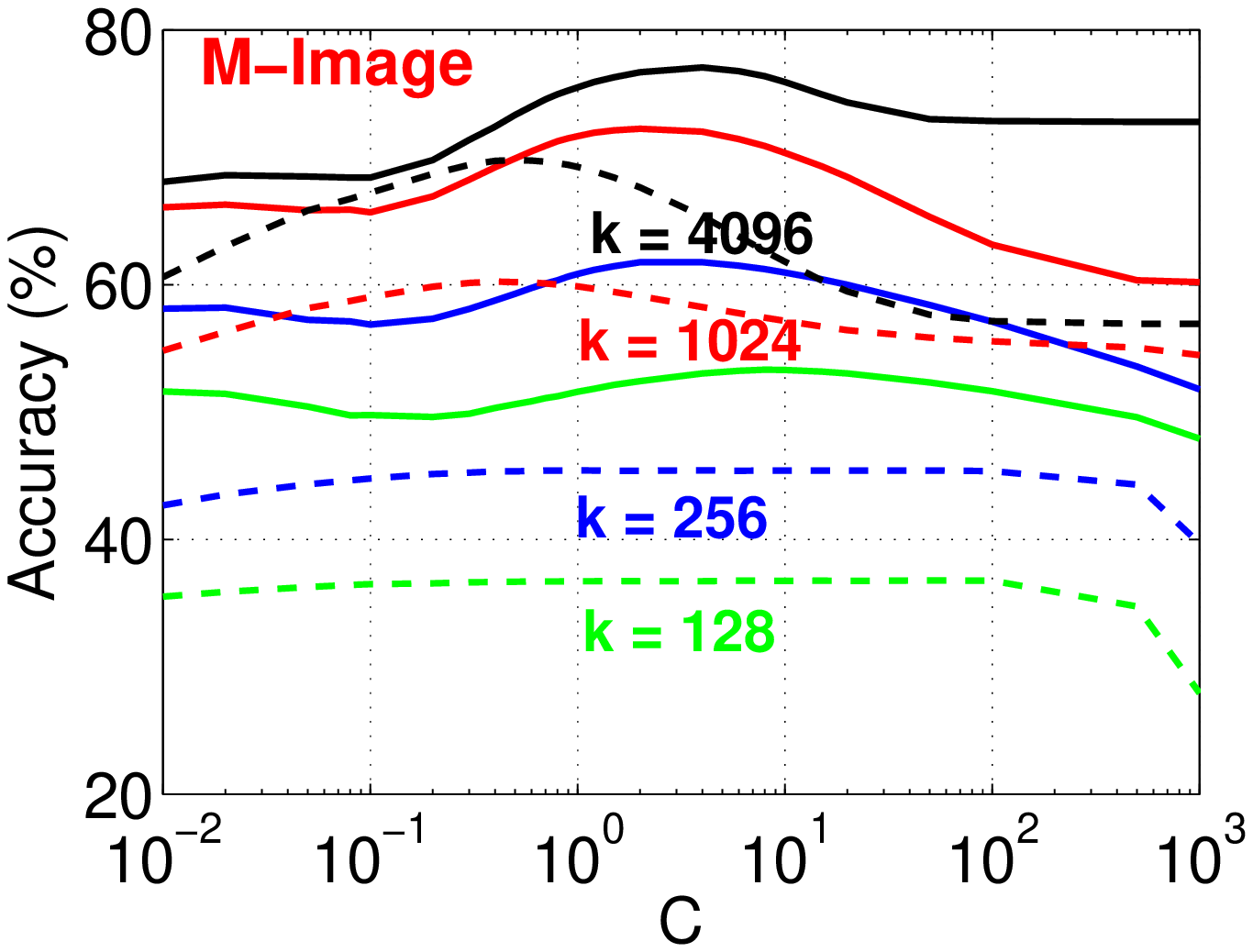}
}

\vspace{-0.038in}

\hspace{-0in}\mbox{
\includegraphics[width=2.2in]{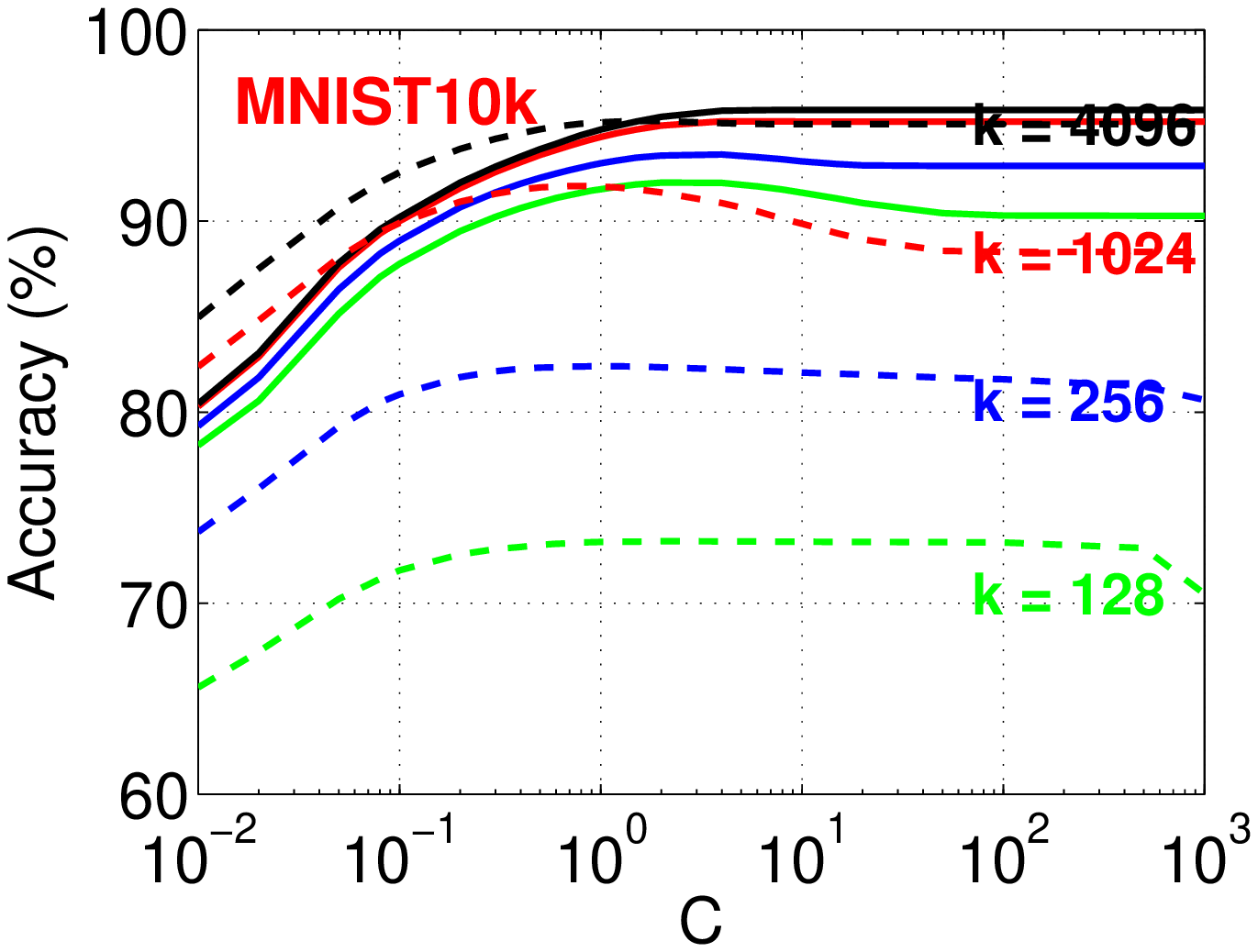}\hspace{0.3in}
\includegraphics[width=2.2in]{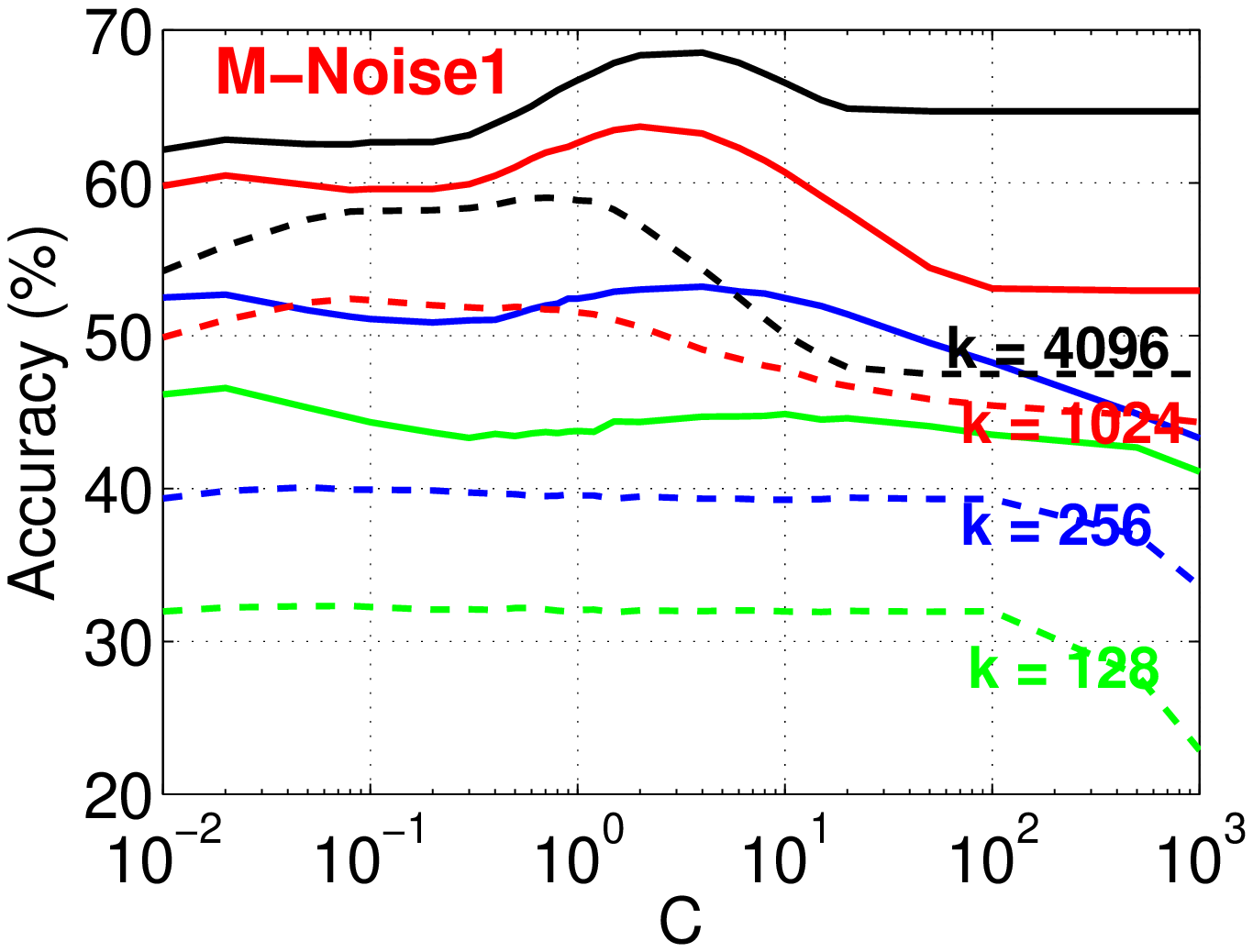}
}

\vspace{-0.038in}

\hspace{-0in}\mbox{
\includegraphics[width=2.2in]{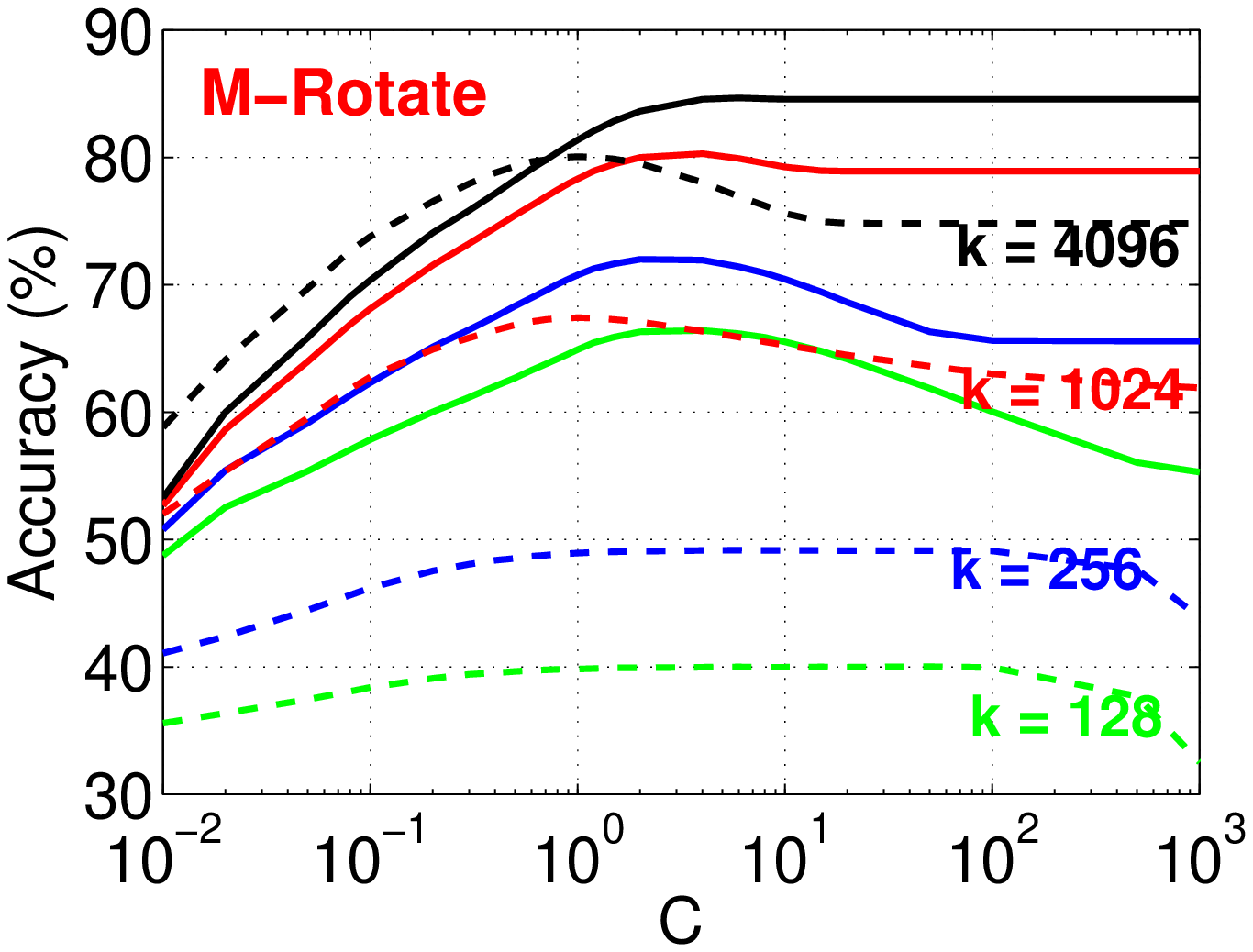}\hspace{0.3in}

\includegraphics[width=2.2in]{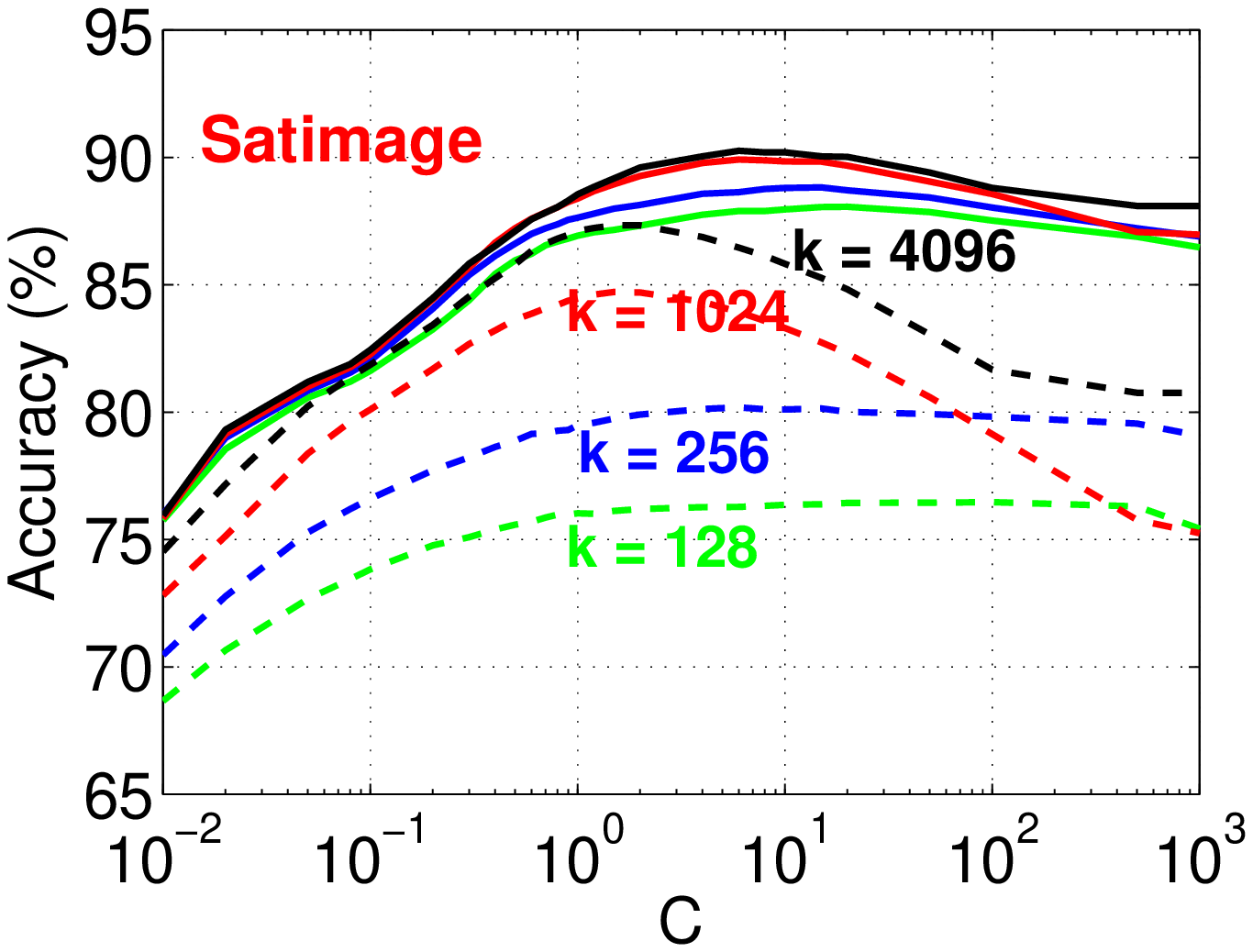}

}

\vspace{-0.038in}

\hspace{-0in}\mbox{
\includegraphics[width=2.2in]{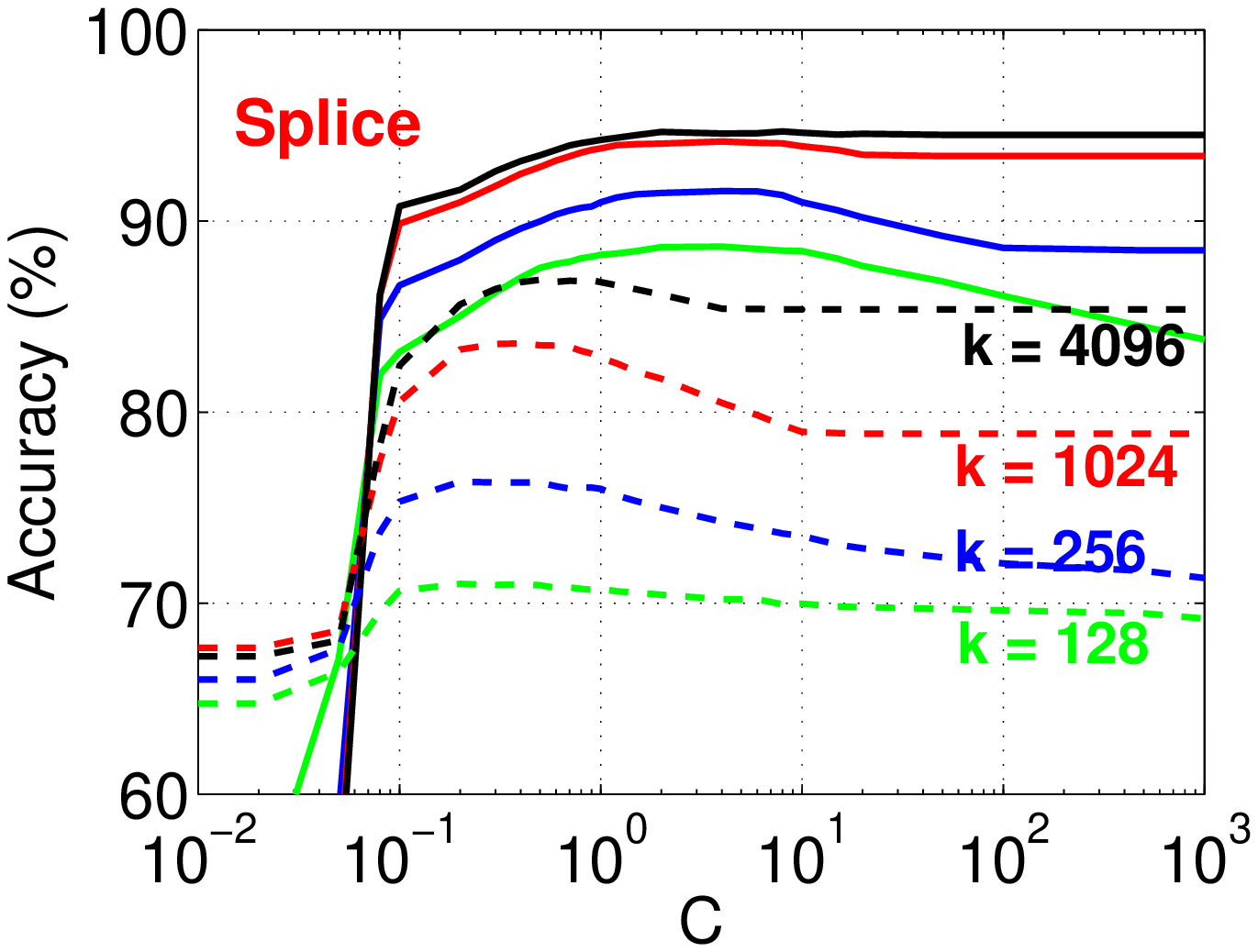}\hspace{0.3in}
\includegraphics[width=2.2in]{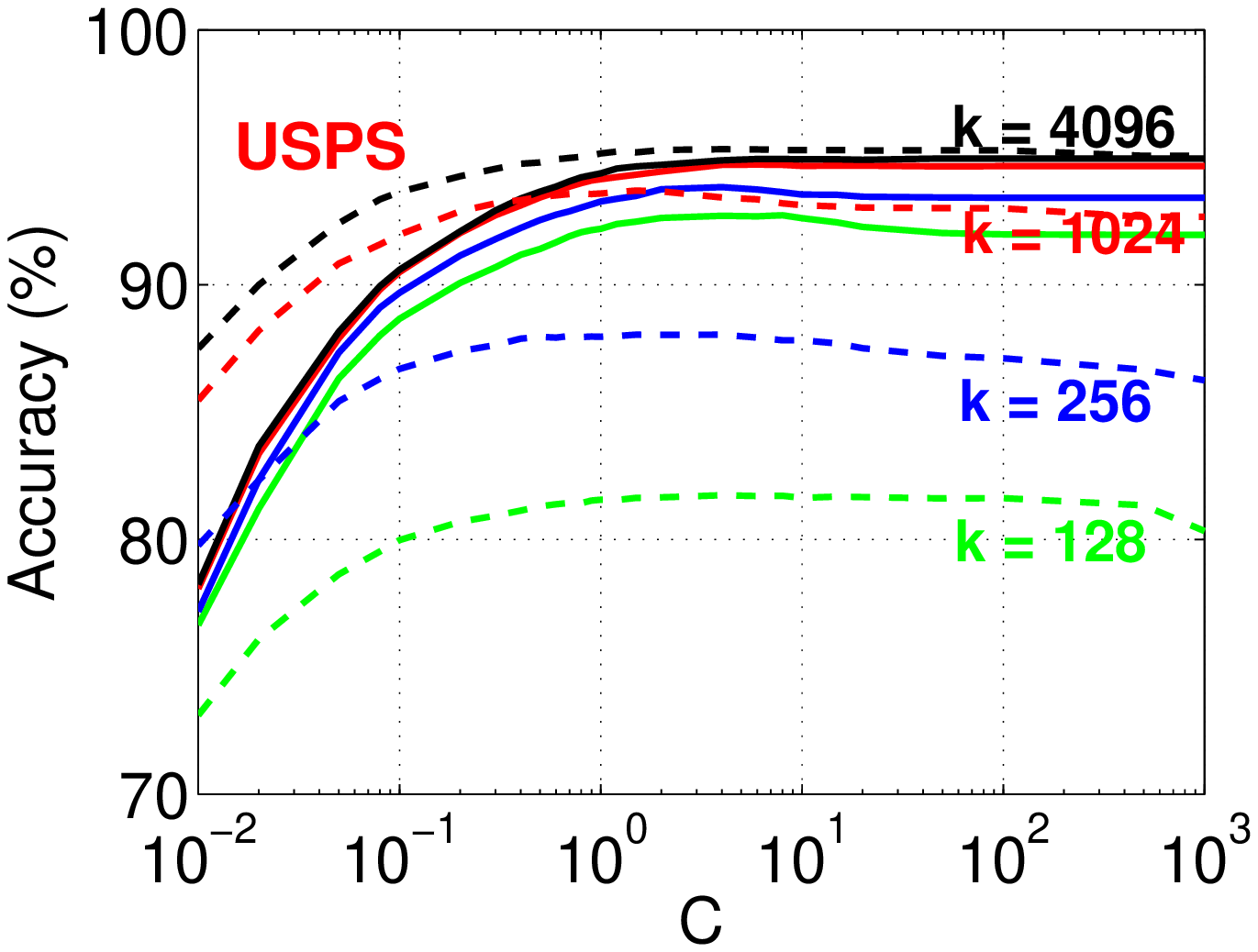}

}

\vspace{-0.038in}

\hspace{-0in}\mbox{
\includegraphics[width=2.2in]{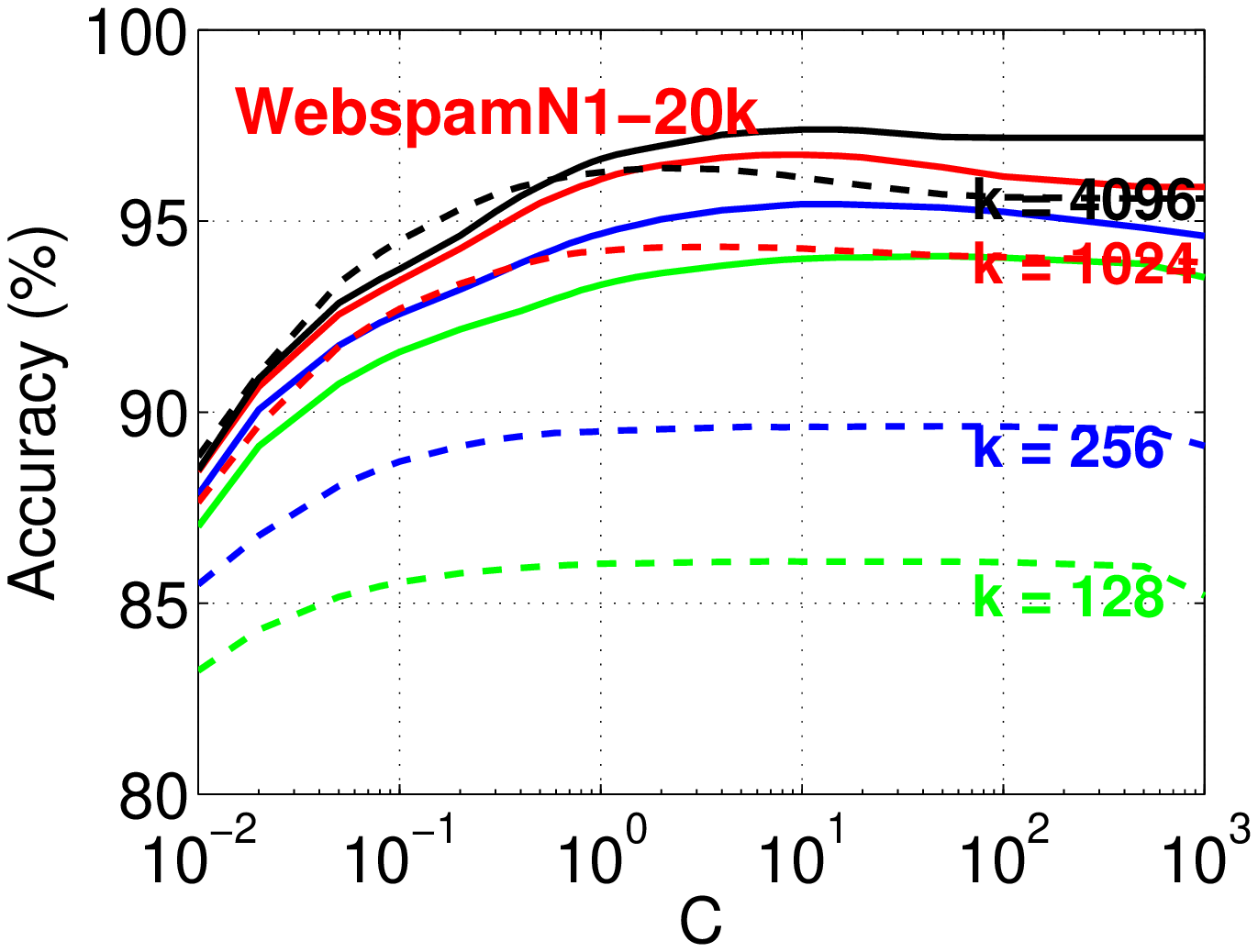}\hspace{0.3in}
\includegraphics[width=2.2in]{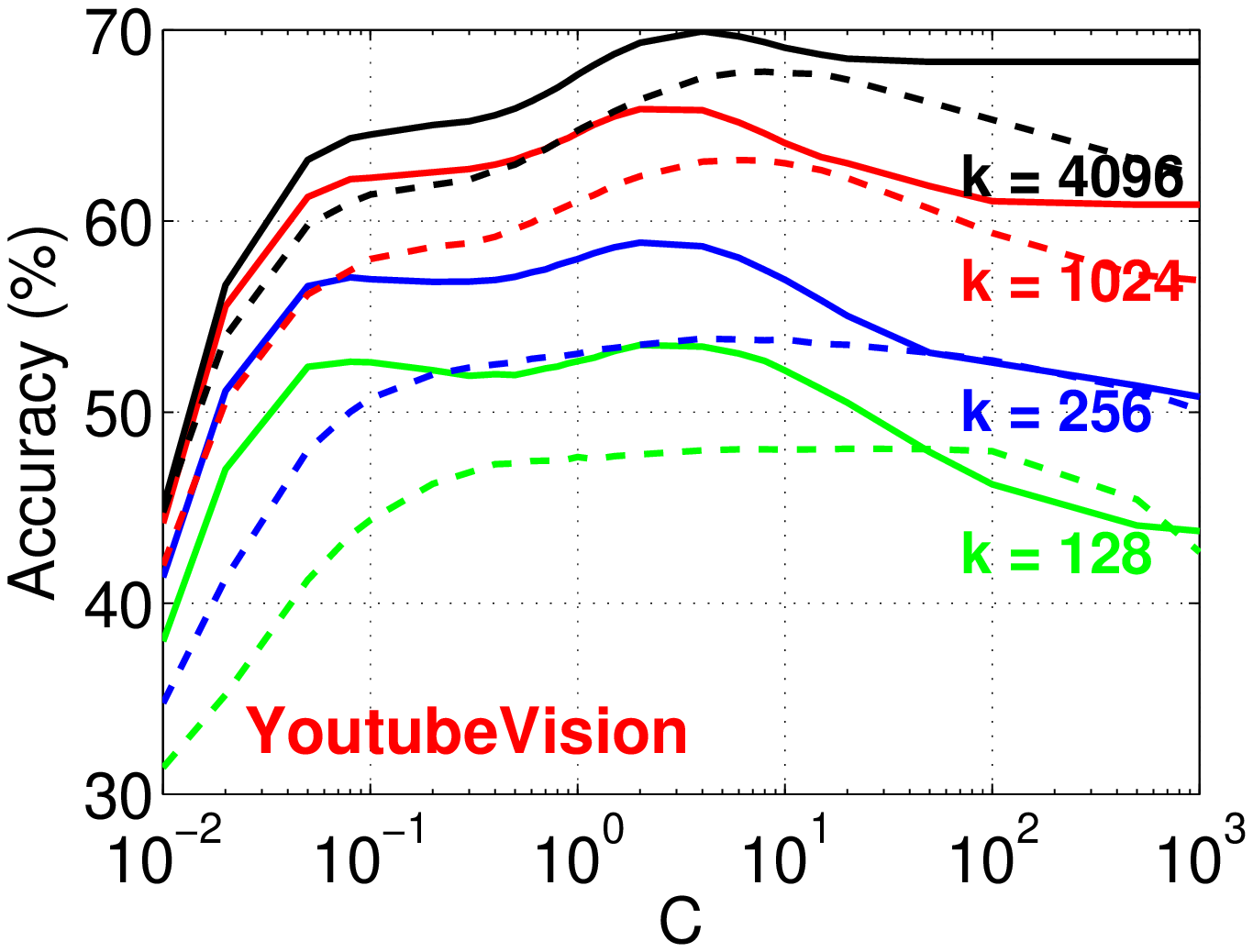}

}

\vspace{-0.3in}
\end{center}
\caption{Classification accuracies of the linearized min-max kernel  (solid curves) and  linearized RBF (dashed curves) kernel, using LIBLINEAR. We report the results on 4 different $k$ (sample size) values: 128, 256, 1024, 4096. We only label the dashed curves. We can see that linearized RBF would require substantially more samples in order to reach the same accuracies as the linearized min-max method.}\label{fig_CWS/RBF}
\vspace{-0.15in}
\end{figure}

Figure~\ref{fig_CWS/RBF} compares the linearization results of the min-max kernel with the results of the RBF kernel. We can see that the linearization algorithm for RBF performs very poorly when the sample size $k$ is small (e.g., $k< 1024$). Even with $k=4096$, the accuracies still do not reach the accuracies using the original RBF kernel as reported in Table~\ref{tab_KernelSVM}.\\

There is an interesting example. For the ``M-Rotate'' dataset, the original RBF kernel notably outperforms the original min-max kernel ($89.7\%$ versus $84.8\%$). However, as shown in Figure~\ref{fig_CWS/RBF}, even with 4096 samples, the accuracy of the linearized RBF kernel is still substantially lower than the accuracy of the linearized min-max kernel.\\

These observations motivate  a useful future research: Can we develop an improved linearization algorithm for RBF/fRBF kernels which would require much fewer samples to reach good accuracies?

\subsection{Min-max Kernel versus acos and acos-$\chi^2$ Kernels}

\begin{figure}[h!]
\begin{center}

\hspace{-0in}\mbox{
\includegraphics[width=2.2in]{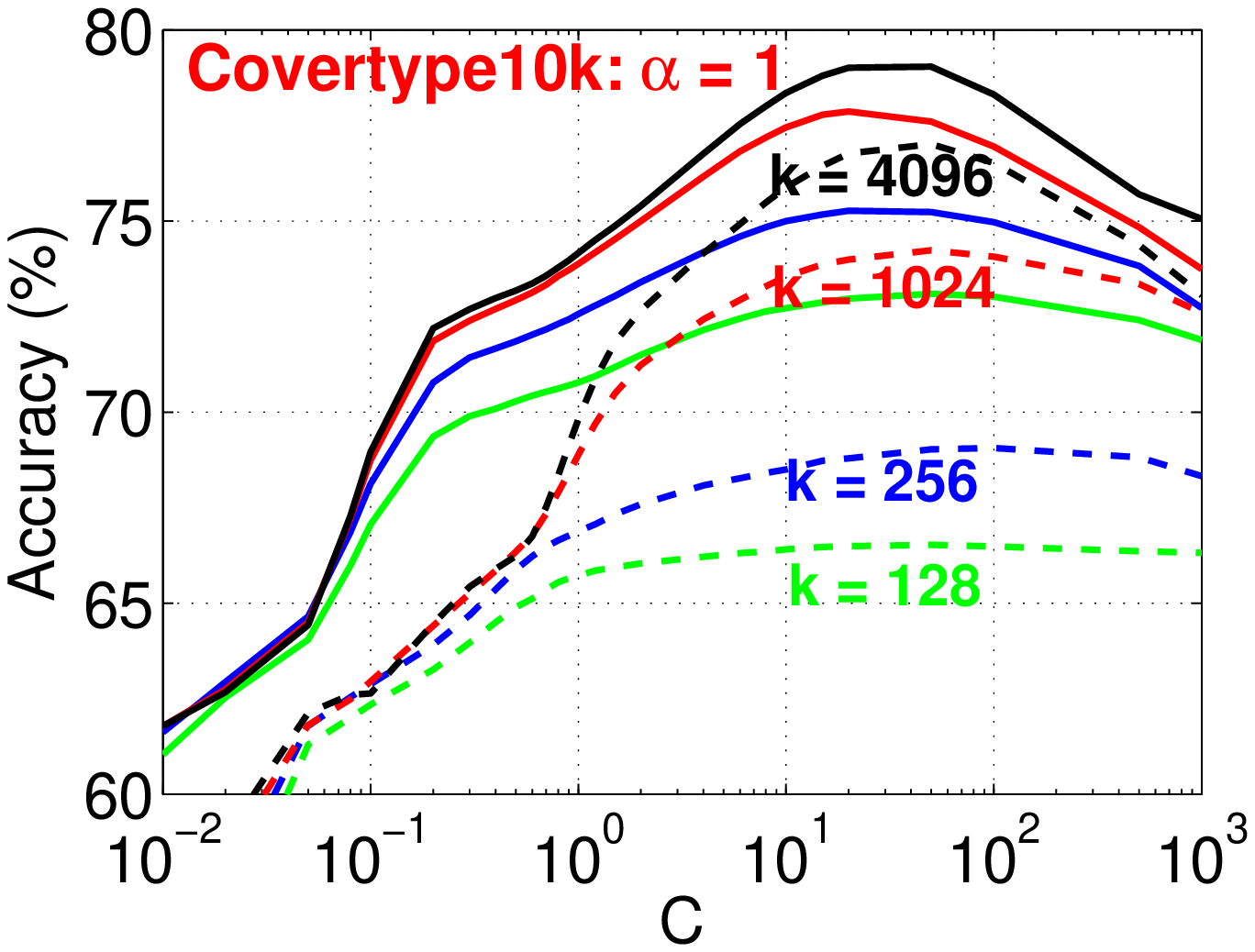}\hspace{0.3in}
\includegraphics[width=2.2in]{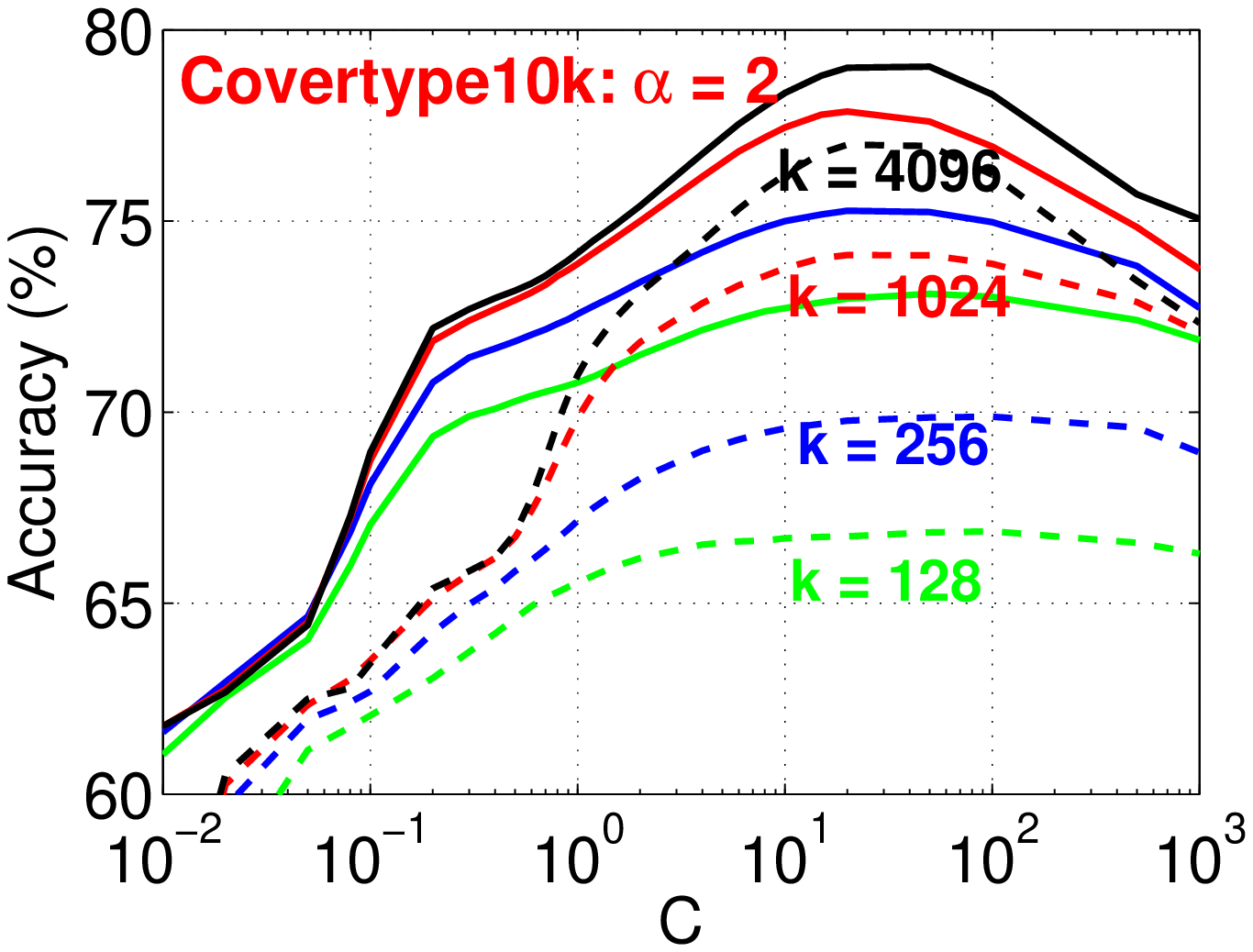}
}

\vspace{-0.038in}

\hspace{-0in}\mbox{
\includegraphics[width=2.2in]{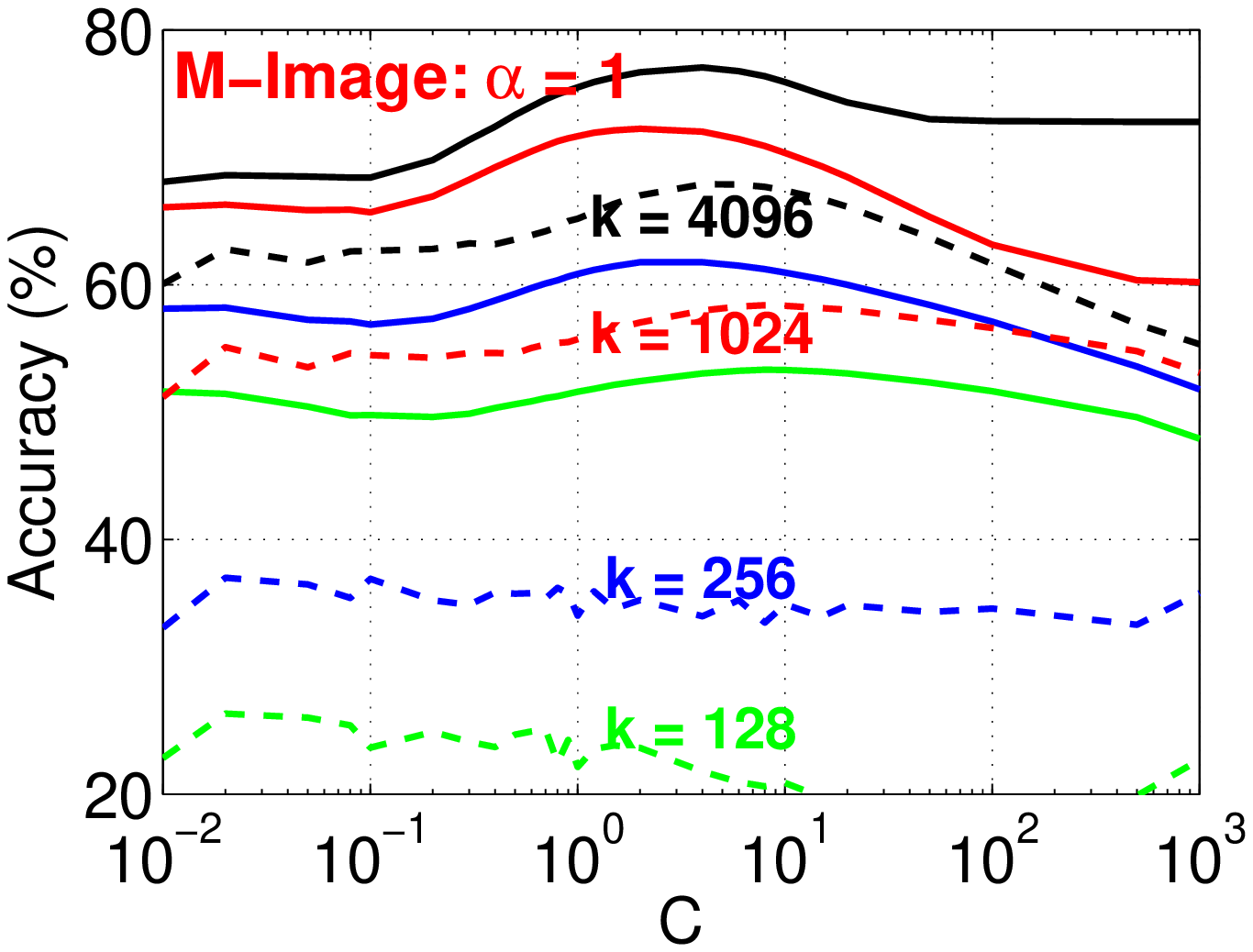}\hspace{0.3in}
\includegraphics[width=2.2in]{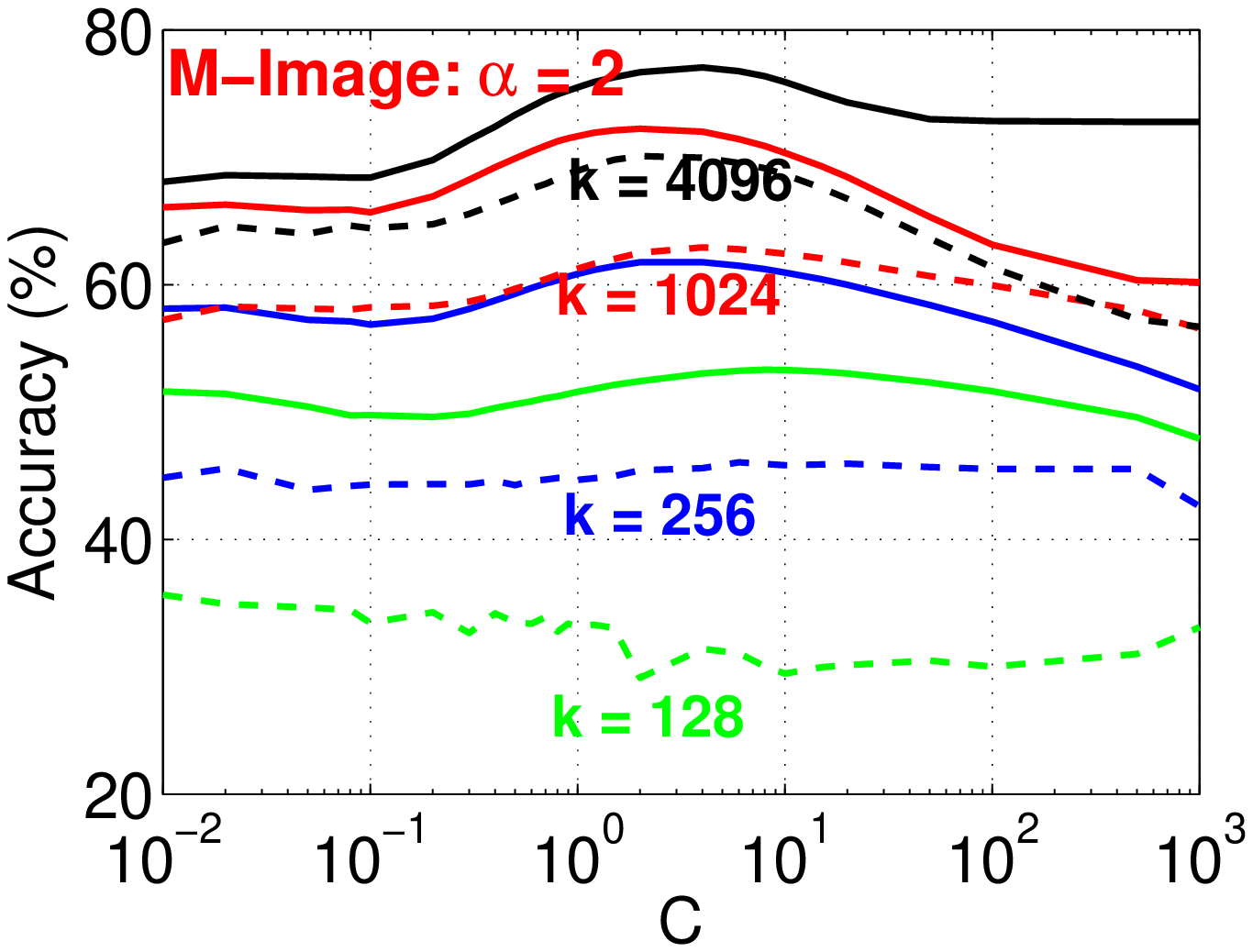}
}

\vspace{-0.038in}

\hspace{-0in}\mbox{
\includegraphics[width=2.2in]{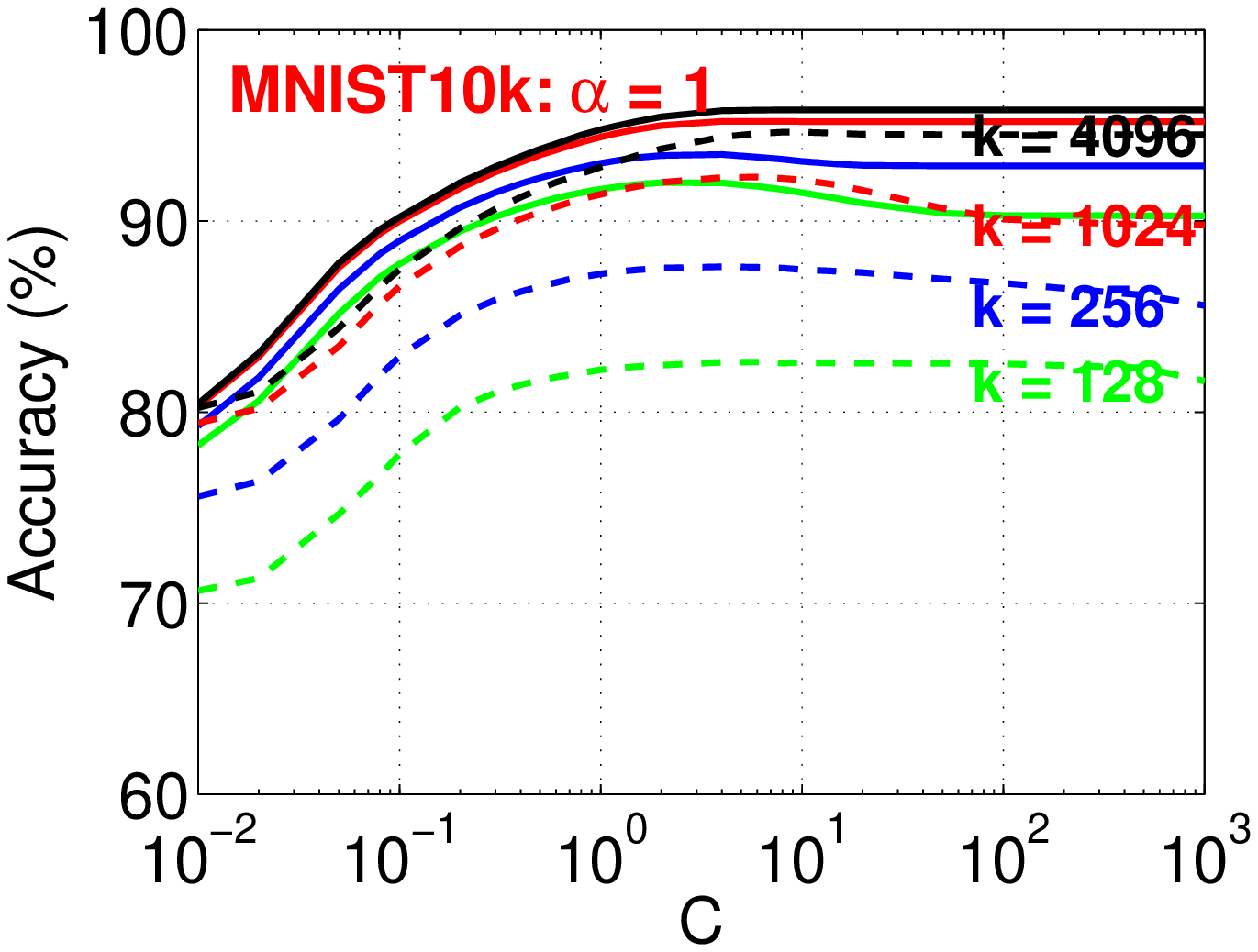}\hspace{0.3in}
\includegraphics[width=2.2in]{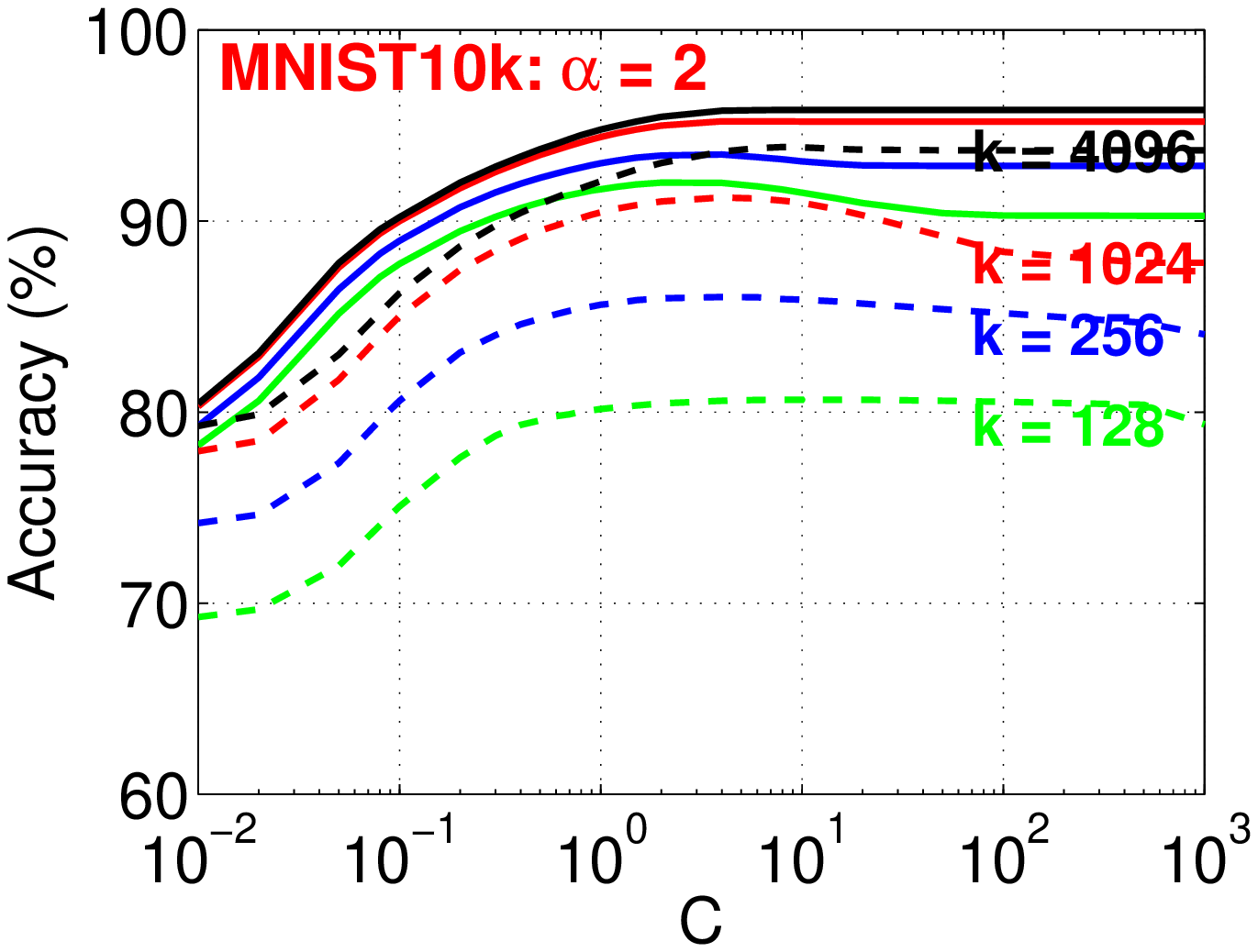}
}

\vspace{-0.038in}

\hspace{-0in}\mbox{
\includegraphics[width=2.2in]{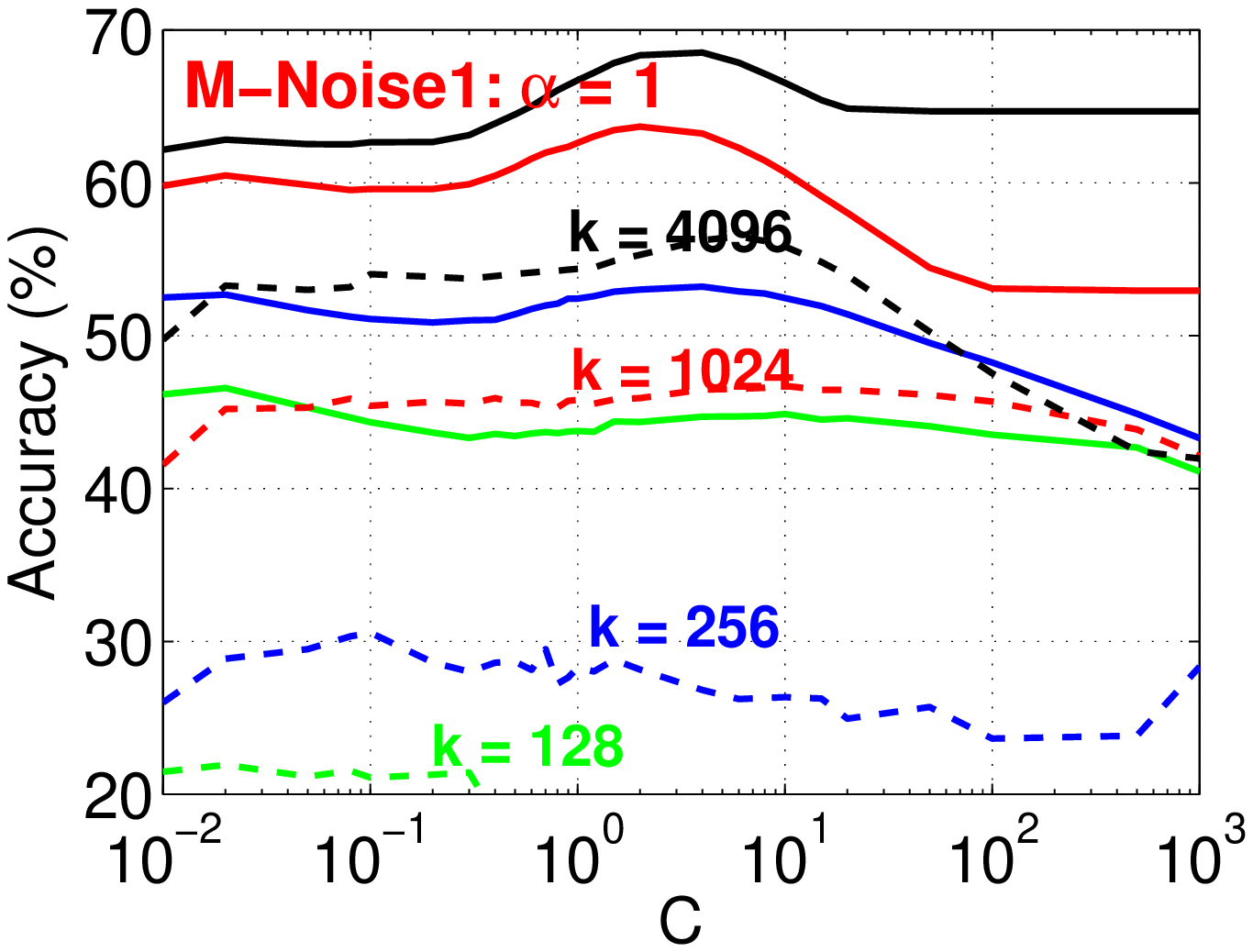}\hspace{0.3in}
\includegraphics[width=2.2in]{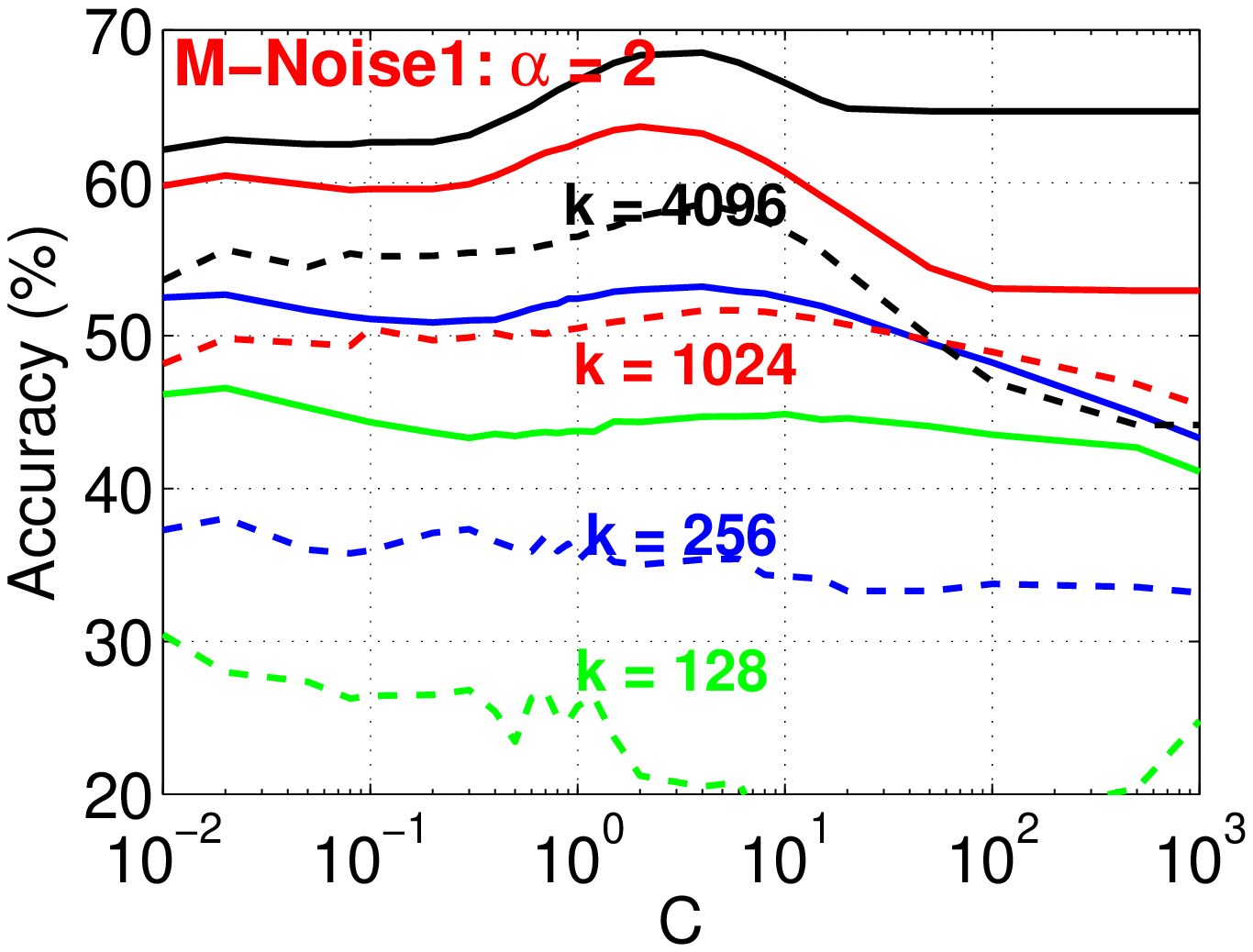}
}

\vspace{-0.038in}

\hspace{-0in}\mbox{
\includegraphics[width=2.2in]{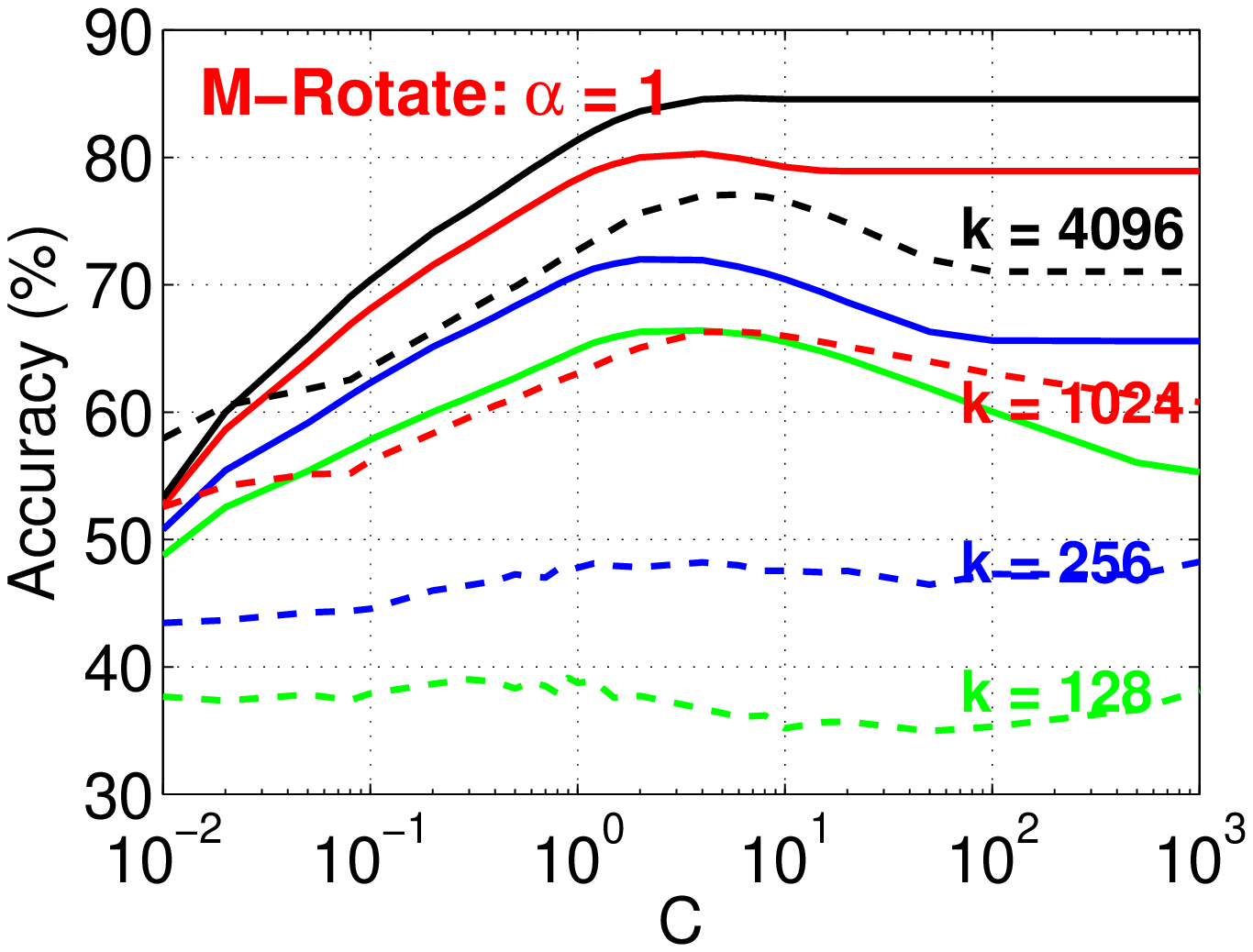}\hspace{0.3in}
\includegraphics[width=2.2in]{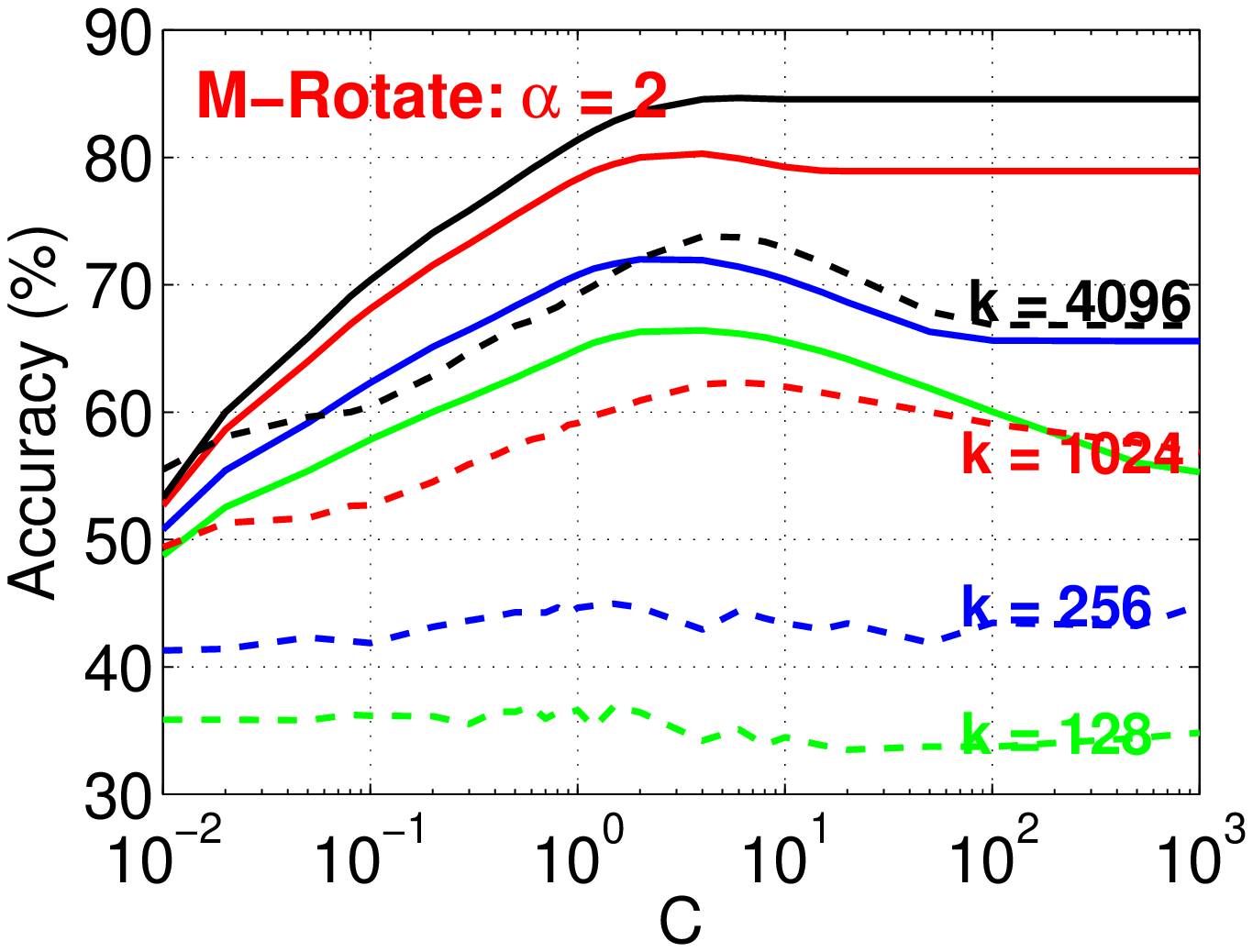}
}

\vspace{-0.3in}
\end{center}
\caption{Classification accuracies of the linearized min-max kernel  (solid curves) and acos (dashed curves) kernel (right panels, i.e., $\alpha=2$) and the acos-$\chi^2$ (dashed curves) kernel (left panels, i.e., $\alpha=1$), using LIBLINEAR. We report the results on 4 different $k$ (sample size) values: 128, 256, 1024, 4096. We only label the dashed curves. We can see that linearized acos and acos-$\chi^2$ kernels require substantially more samples in order to reach the same accuracies as the linearized min-max method.}\label{fig_CWS/acos1}
\vspace{-0in}
\end{figure}

\begin{figure}[h!]
\begin{center}

\hspace{-0in}\mbox{
\includegraphics[width=2.2in]{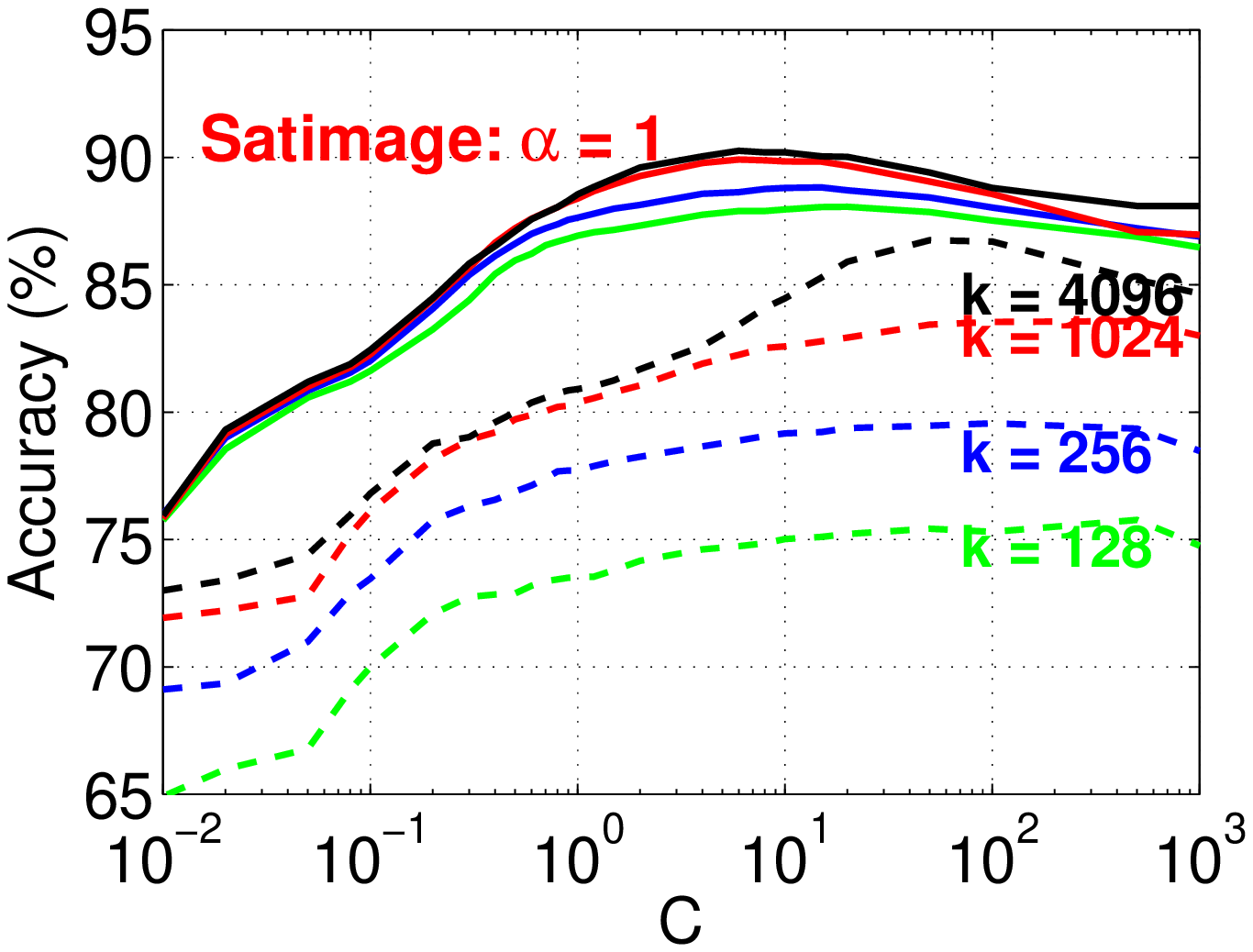}\hspace{0.3in}
\includegraphics[width=2.2in]{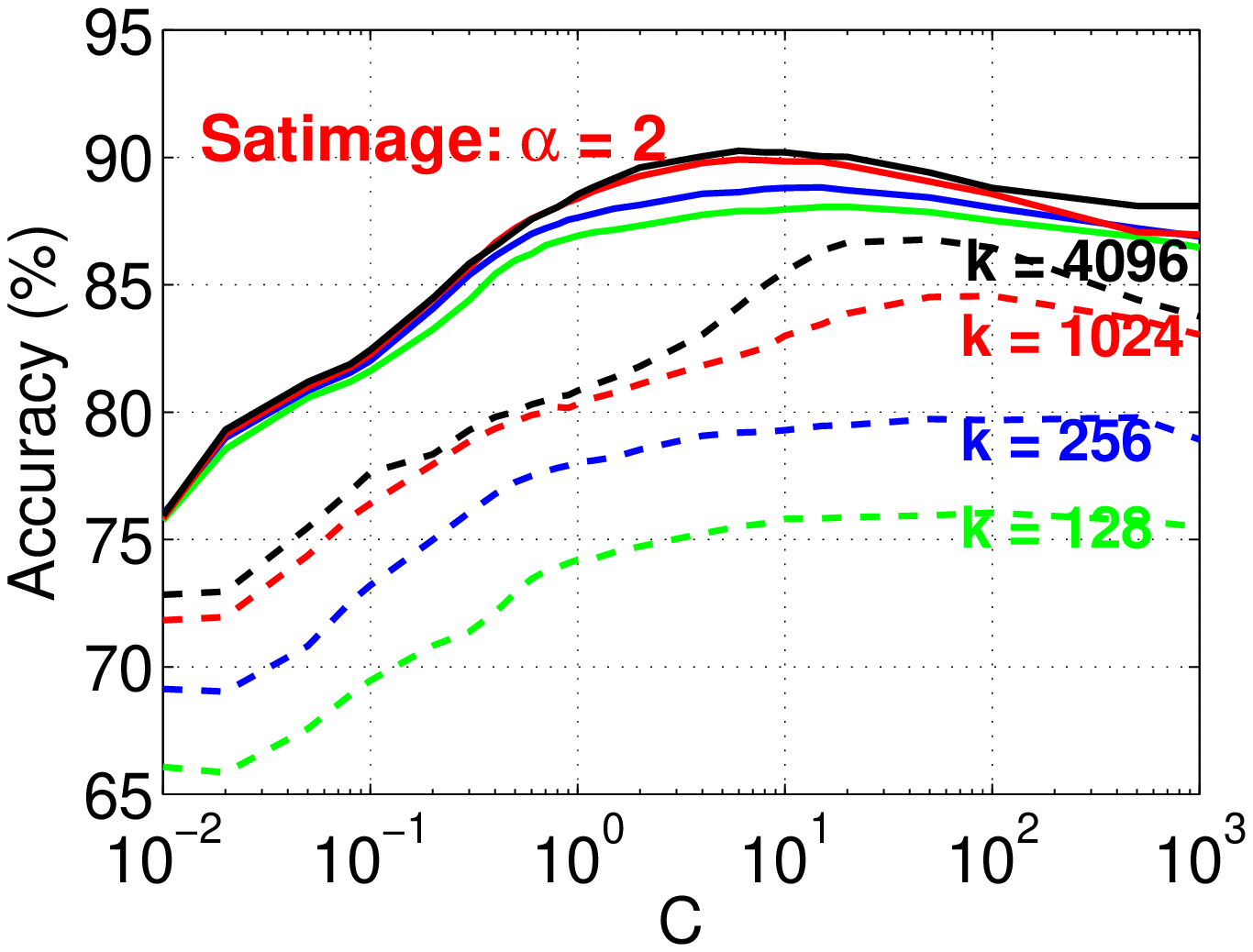}
}

\vspace{-0in}

\hspace{-0in}\mbox{
\includegraphics[width=2.2in]{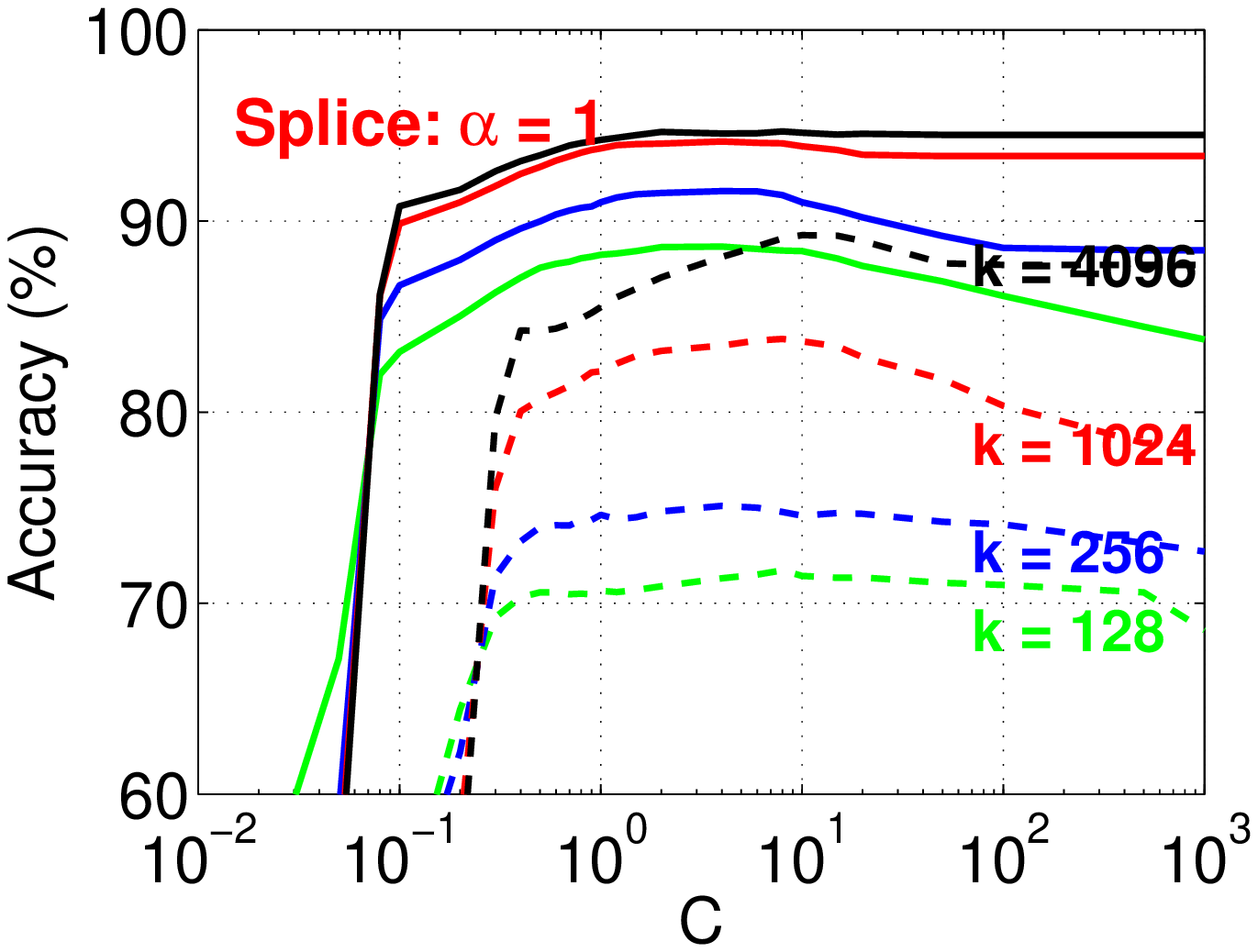}\hspace{0.3in}
\includegraphics[width=2.2in]{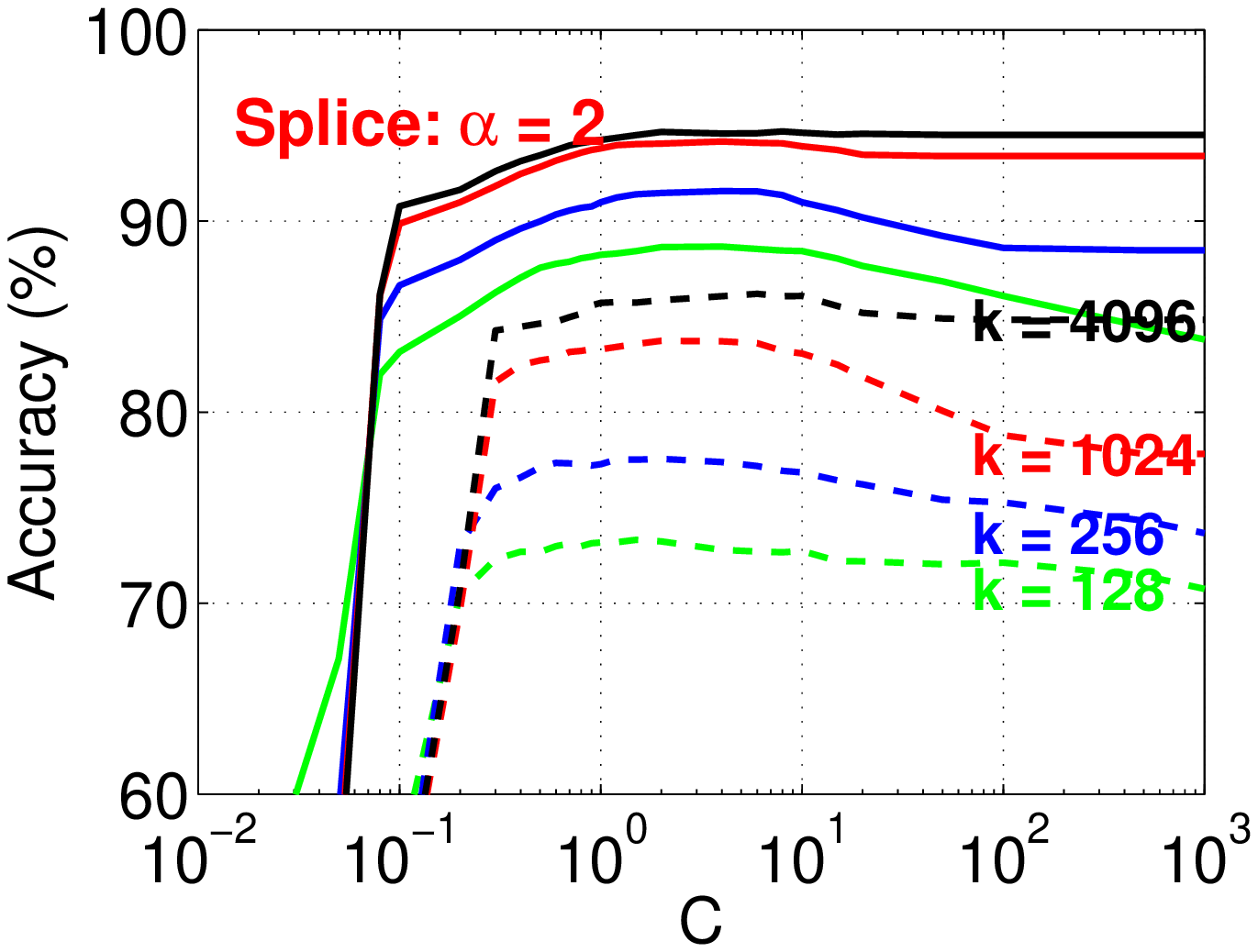}
}

\vspace{-0in}

\hspace{-0in}\mbox{
\includegraphics[width=2.2in]{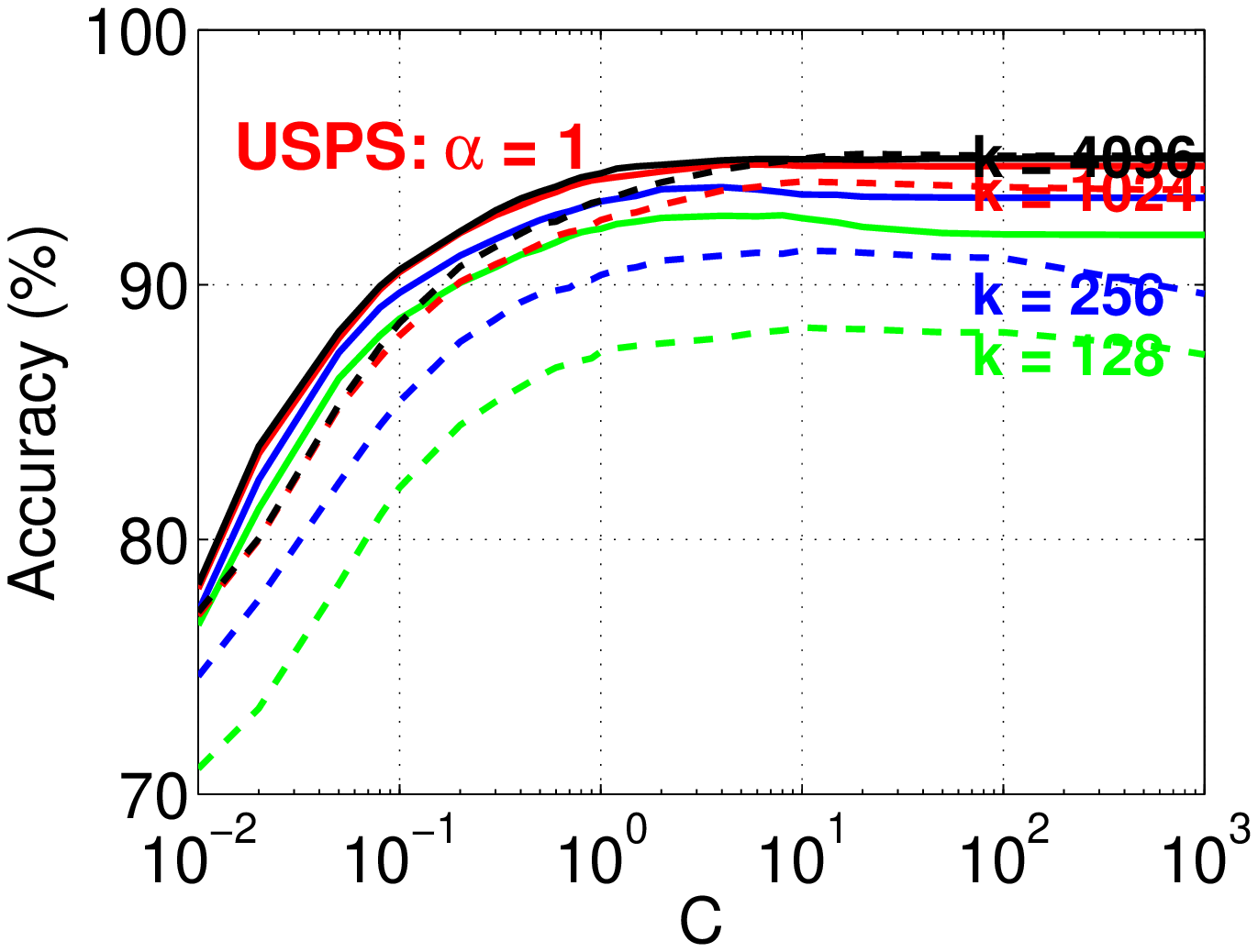}\hspace{0.3in}
\includegraphics[width=2.2in]{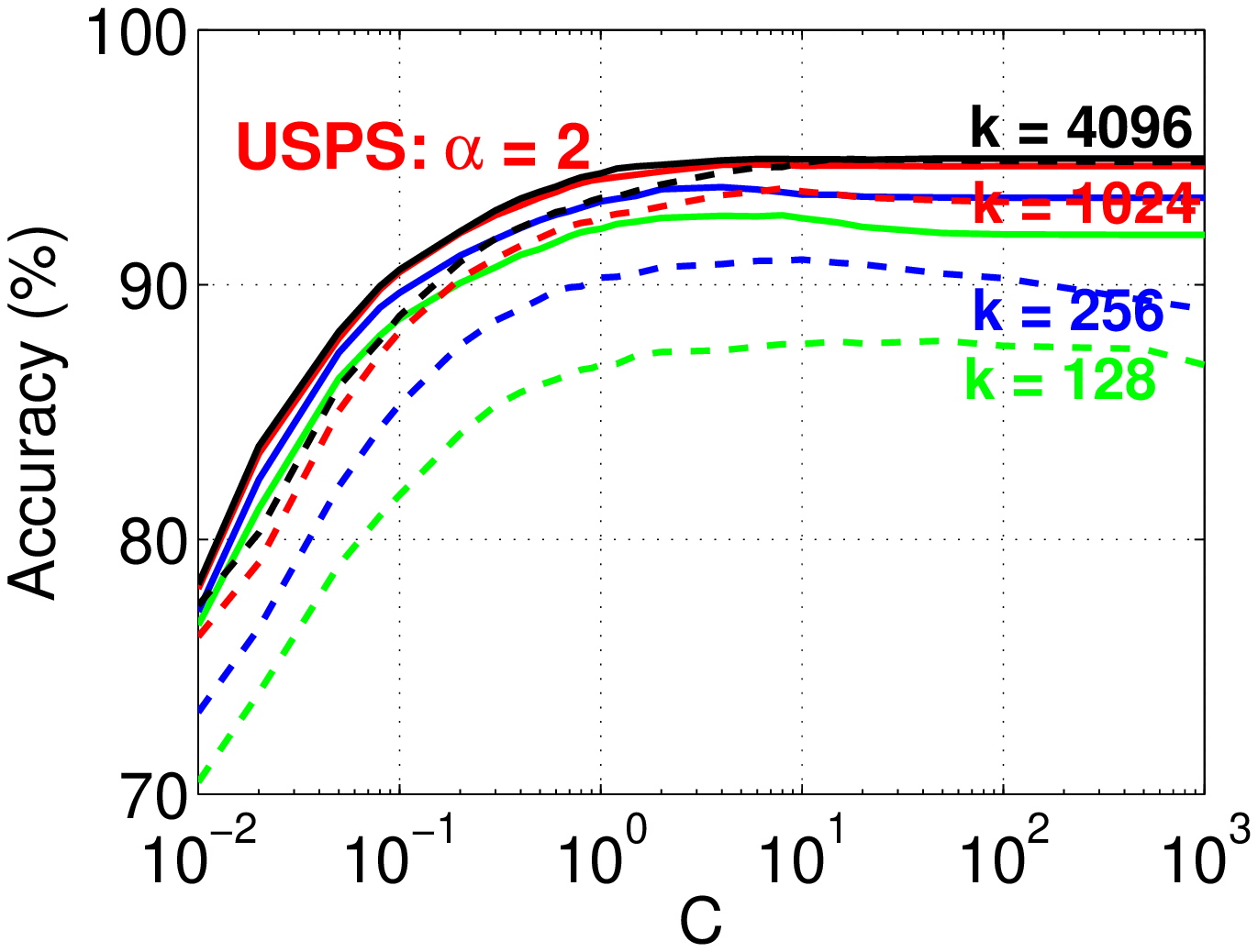}
}

\vspace{-0in}

\hspace{-0in}\mbox{
\includegraphics[width=2.2in]{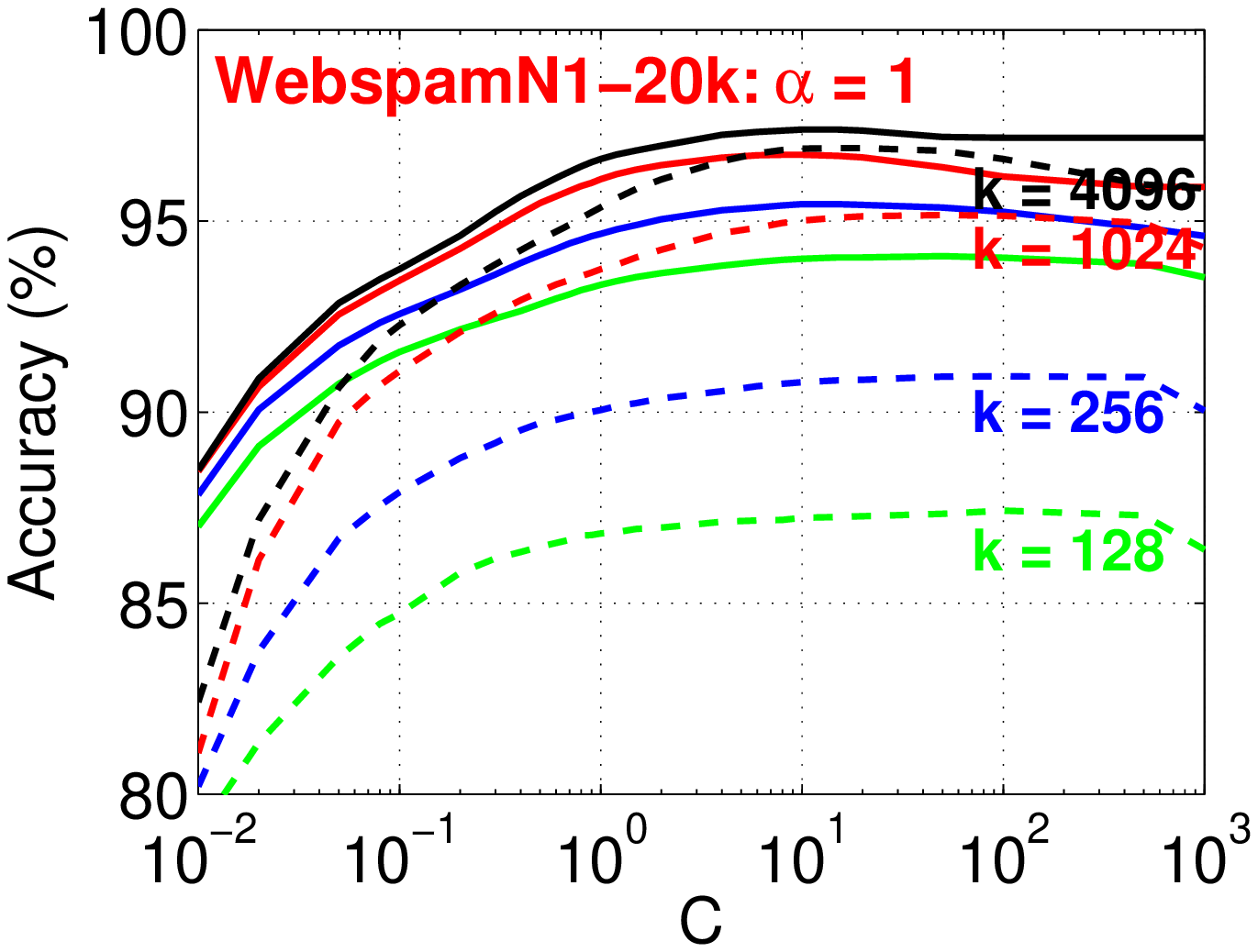}\hspace{0.3in}
\includegraphics[width=2.2in]{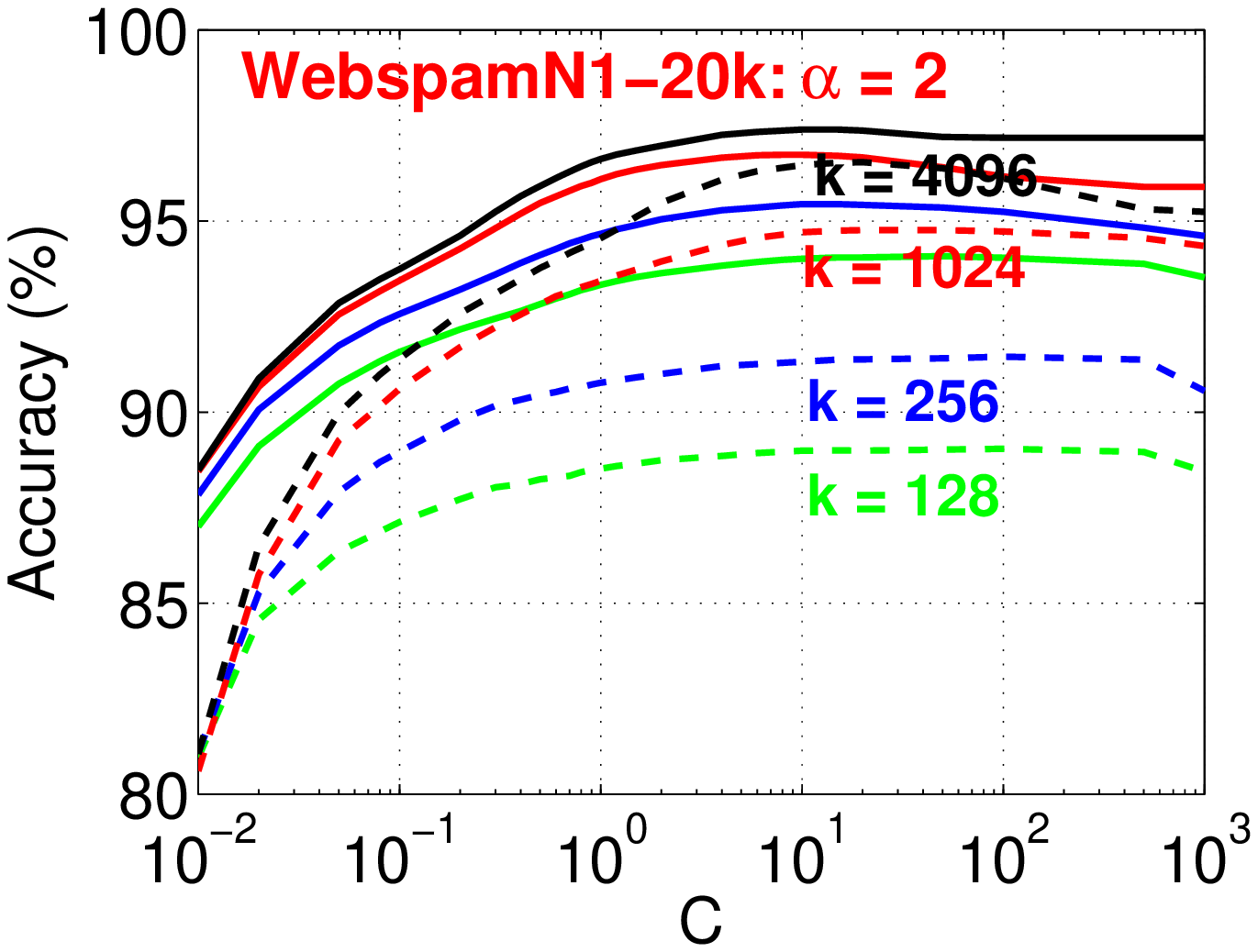}
}

\vspace{-0in}

\hspace{-0in}\mbox{
\includegraphics[width=2.2in]{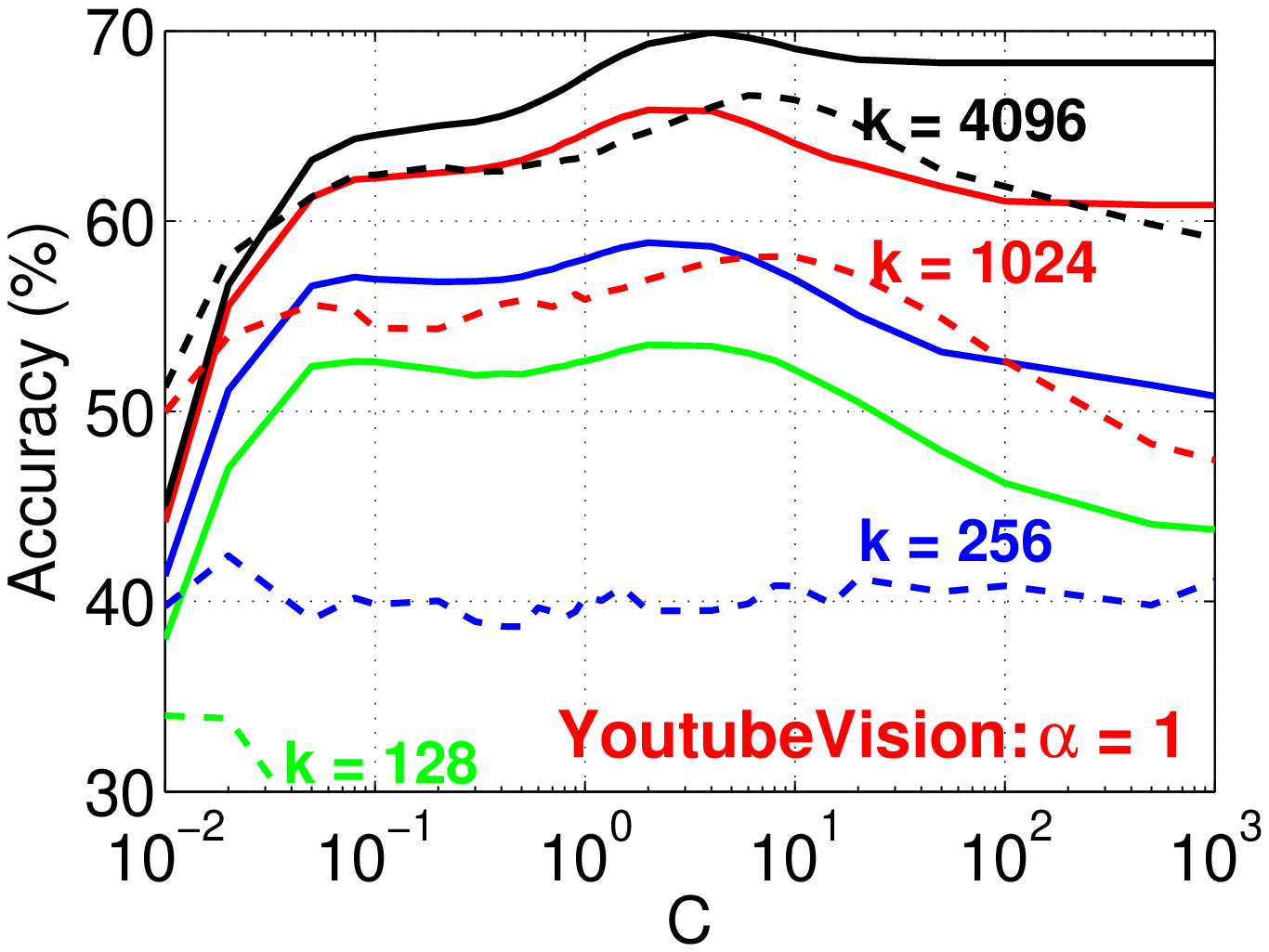}\hspace{0.3in}
\includegraphics[width=2.2in]{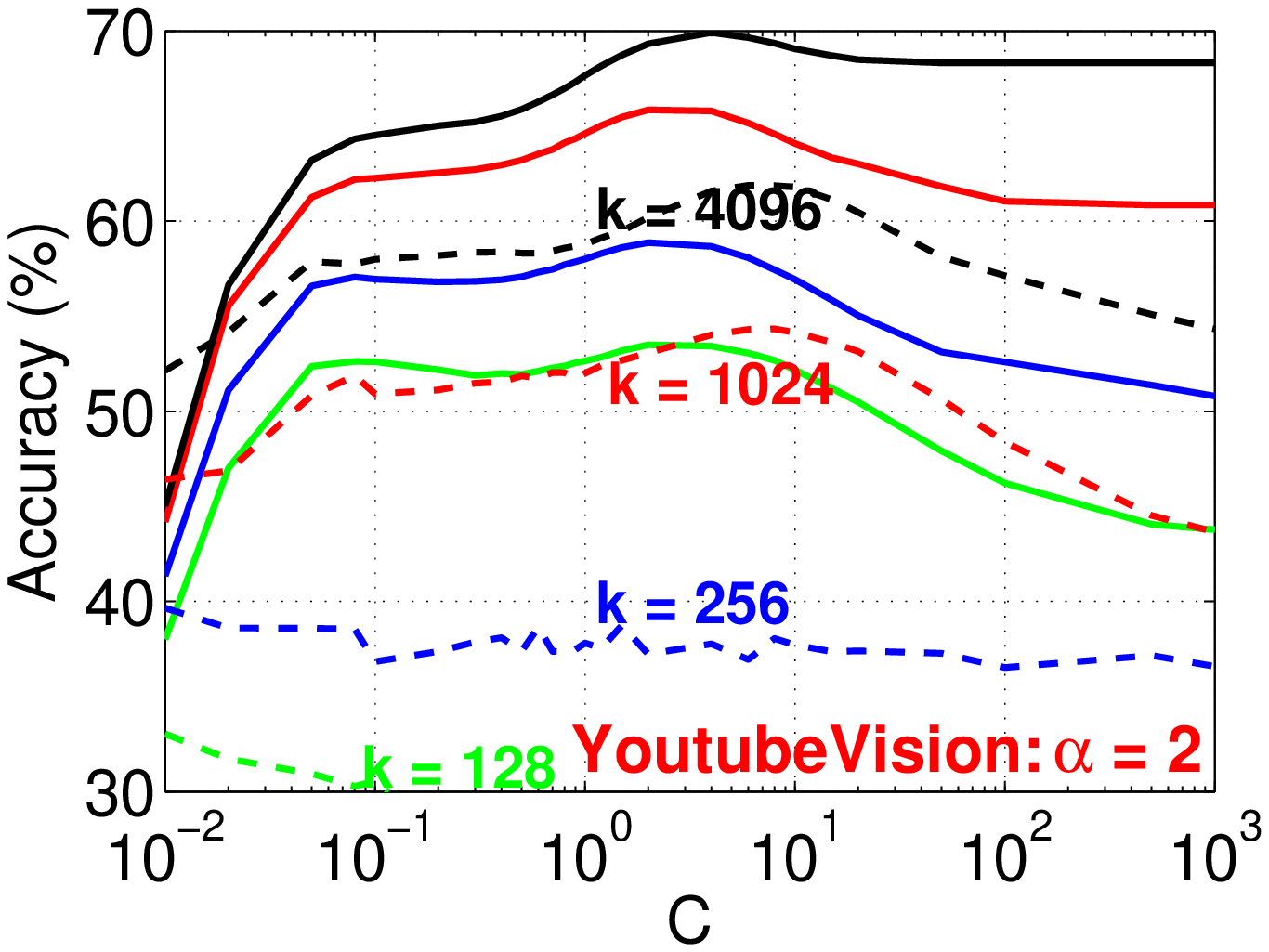}
}

\end{center}
\vspace{-0.3in}
\caption{Classification accuracies of the linearized min-max kernel  (solid curves) and acos (dashed curves) kernel (right panels, i.e., $\alpha=2$) and the acos-$\chi^2$ (dashed curves) kernel (left panels, i.e., $\alpha=1$), using LIBLINEAR. Again, we can see that linearized acos and acos-$\chi^2$ kernels require substantially more samples in order to reach the same accuracies as the linearized min-max method.}\label{fig_CWS/acos2}\vspace{-0in}
\end{figure}

As introduced at the beginning of the paper, sign Gaussian random projections and sign Cauchy random projections are the linearization methods for the acos kernel and the acos-$\chi^2$ kernel, respectively.  Figures~\ref{fig_CWS/acos1} and~\ref{fig_CWS/acos2} compare them with 0-bit CWS, where we use ``$\alpha=2$'' for sign Gaussian  projections and ``$\alpha=1$'' for sign Cauchy  projections.\\

Again, like in Figure~\ref{fig_CWS/RBF}, we can see that the linearization method for the min-max kernel requires substantially few samples than the linearization methods for the acos and acos-$\chi^2$ kernels. Since both kernels show reasonably good performance (without linearization), This should also motivate us to pursue improved linearization algorithms for the acos and acos-$\chi^2$ kernels, as future research. \vspace{-0.15in}

\clearpage\newpage

\clearpage\newpage

\subsection{Comparisons on a Larger Dataset}

\begin{figure}[h!]
\begin{center}

\hspace{-0in}\mbox{
\includegraphics[width=2.5in]{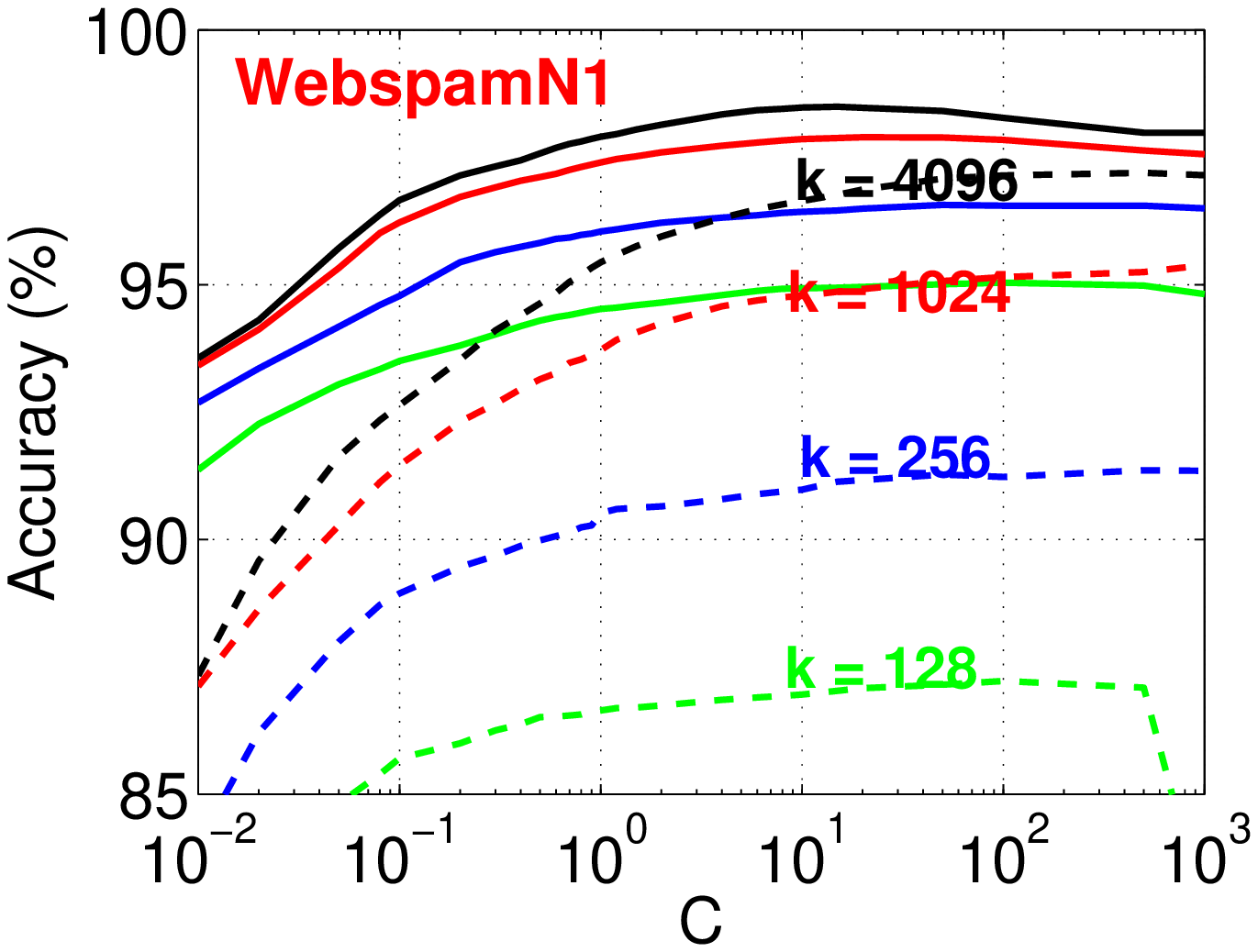}\hspace{0.3in}
\includegraphics[width=2.5in]{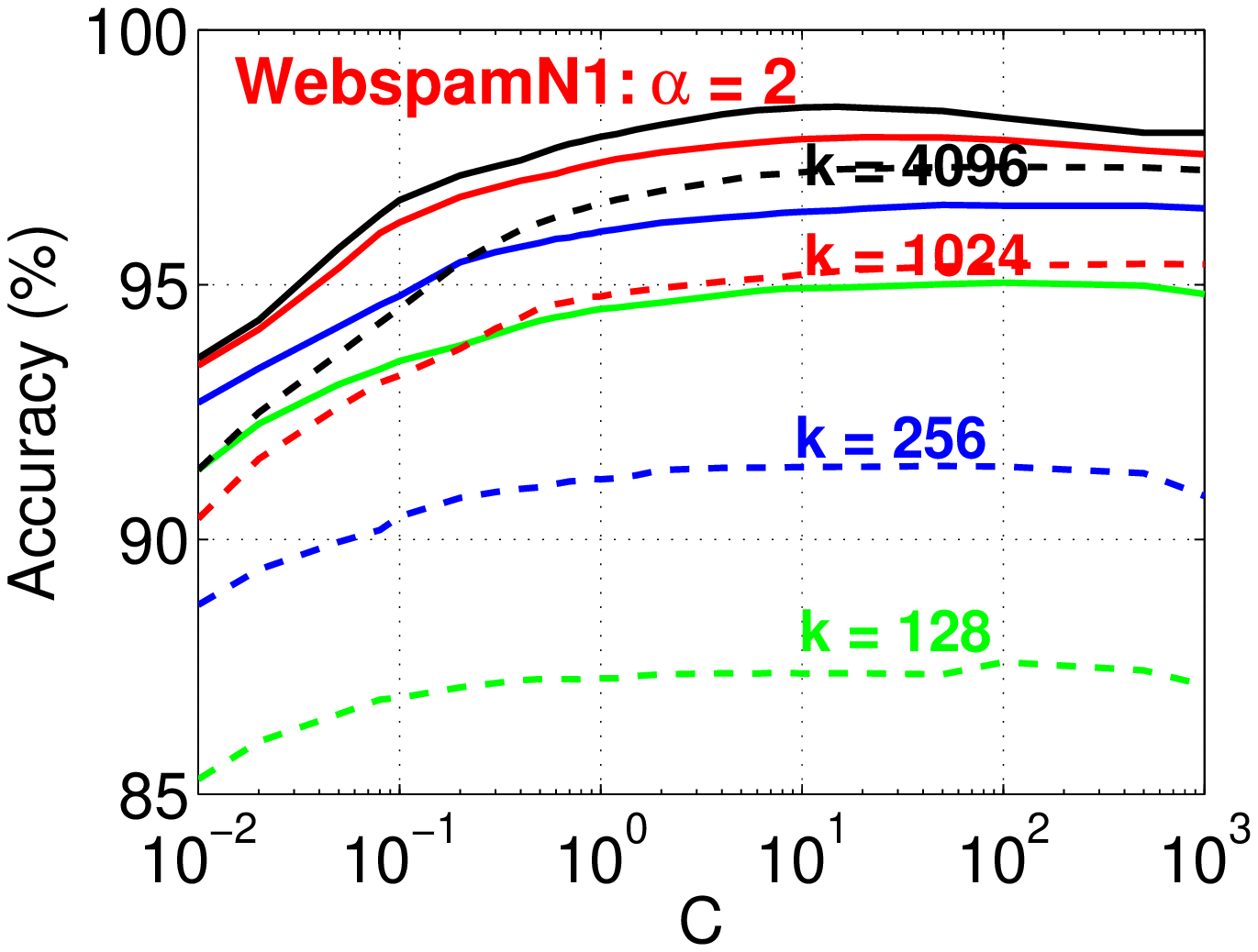}
}

\end{center}
\vspace{-.3in}
\caption{\textbf{Experiments on a larger dataset}. Left panel: Classification accuracies of the linearized min-max kernel  (solid curves) and the linearized  RBF (dashed curves) kernel.  Right panel: Classification accuracies of the linearized min-max kernel  (solid curves) and the linearized  acos (dashed curves) kernel (i.e., $\alpha=2$).  The linearization method for the min-max kernel (i.e., 0-bit CWS) substantially outperforms the linearization methods for  the other two  kernels.}\label{fig_WebspamN1}\vspace{-0.1in}
\end{figure}

Figure~\ref{fig_WebspamN1} provides the comparison study on the ``WebspamN1'' dataset, which has 175,000 examples for training and 175,000 examples for testing. It is too large for using the LIBSVM pre-computed kernel functionality in common workstations. On the other hand, we can easily linearize the nonlinear kernels and run LIBLINEAR on the transformed dataset.\\

The left panel of Figure~\ref{fig_WebspamN1} compares the results of linearization method (i.e., 0-bit CWS) for the min-max kernel with the results of the linearization method for the RBF kernel. The right panel compares  0-bit CWS with sign Gaussian random projections (i.e., $\alpha=2$). We do not present the results for $\alpha=1$ since they are quite similar. The plots again confirm that  0-bit CWS significantly outperforms the linearization methods for both  the RBF kernel  and  the acos kernel.

\section{Kernel Combinations}

It is  an interesting idea to combine kernels for better (or more robust) performance. One simple strategy is to use multiplication of  kernels. For example, the following two new  kernels
\begin{align}\label{eqn_MM-acos}
&\text{MM-acos}(u,v) = \text{MM}(u,v)\times \text{acos}(u,v)\\\label{eqn_MM-acos-chi2}
&\text{MM-acos-}\chi^2(u,v) = \text{MM}(u,v)\times \text{acos-}\chi^2(u,v)
\end{align}
combine the min-max kernel with the acos kernel or the acos-$\chi^2$ kernel. They are still positive definite because they are the multiplications of positive definite kernels.\\

\begin{table}[h!]
\caption{Classification accuracies (in $\%$) of the two new kernels: MM-acos defined in (\ref{eqn_MM-acos}) and MM-acos-$\chi^2$  defined in (\ref{eqn_MM-acos-chi2}), as presented in the last two columns.
}
\begin{center}{\vspace{-0in}
{\small\begin{tabular}{l c c  c c c c}
\hline \hline
Dataset      &min-max  &acos &acos-$\chi^2$ & MM-acos & MM-acos-$\chi^2$\\
\hline
Covertype10k    &80.4  &{\bf81.9}   &81.6   &{\bf81.9}   &{\bf81.9}\\
Covertype20k    &83.3  &{\bf85.3}   &85.0  &{\bf85.3}  &{\bf85.3} \\
IJCNN5k   &94.4  &{\bf96.9}    &96.6   &95.6   &95.4\\
IJCNN10k &95.7    &{\bf97.5}    &97.4   &96.2   &96.1\\
Isolet   &96.4    &96.5   &96.1   &{\bf96.7}   &96.6\\
Letter   &96.2    &97.0   &97.0   &{\bf97.2}   &{\bf97.2}\\
Letter4k  &91.4   & {\bf93.3}  & {\bf93.3}  &92.9   &92.8\\
M-Basic   &96.2    &95.7   &95.8  &{\bf96.6}   &96.5\\
M-Image    &80.8   &76.2   &75.2   &{\bf81.0}   &80.8\\
MNIST10k  &95.7    &95.2   &95.2   &{\bf96.1}   &96.1\\
M-Noise1  &{\bf71.4}     &65.0  & 64.0  &71.0   &70.8\\
M-Noise2  &{\bf72.4}  &66.9   &65.7  &72.2   &72.0\\
M-Noise3    &73.6    &69.0   &68.0  &{\bf73.9}   &73.5\\
M-Noise4 &{\bf76.1}  &73.1   &71.1  &75.8   &75.5\\
M-Noise5  &{\bf79.0}   &76.6   &74.9  &78.7   &78.5\\
M-Noise6 &84.2    &83.9   &82.8  &{\bf84.6}   &84.3\\
M-Rand   &84.2    &83.5   &82.3  &{\bf84.5}   &84.3\\
M-Rotate &84.8    &84.5  &84.6   &{\bf86.5}   &86.4\\
M-RotImg  &41.0     &41.5  &39.3   &{\bf42.8}   &41.8\\
Optdigits   &97.7    &97.7  &97.5   &97.8  &{\bf97.9}\\
Pendigits  &97.9   & {\bf98.3}  &98.1   &98.2  &98.0\\
Phoneme  &92.5   &92.2   &90.2   &{\bf92.6}  &92.1\\
Protein &{\bf72.4}    &69.2   &70.5   &71.2   &71.4\\
RCV1   &{\bf96.9}   &96.5   &96.7   &96.8   &96.8\\
Satimage   &90.5    &89.5   &89.4  &{\bf91.2}   &90.9\\
Segment &98.1  &97.6   &97.2   &98.1  &{\bf98.3}\\
SensIT20k   &86.9  &85.7  &{\bf87.5}  &87.1   &87.3\\
Shuttle1k   &{\bf99.7}   &{\bf99.7}   &{\bf99.7}   &{\bf99.7} &{\bf99.7}\\
Spam   & 95.0 &94.2   &{\bf95.2}   &94.9   &95.0\\
Splice  &95.2 & 89.2   &91.7   &{\bf95.9}   &95.7\\
USPS   &95.3 &95.3   &{\bf95.5}   &{\bf95.5}   &{\bf95.5}\\
Vowel  &59.1   &{\bf63.0}    &61.3   &58.9   &58.7\\
WebspamN1-20k  &97.9  &98.1   &{\bf98.5}   &98.0   &98.2\\
YoutubeVision &72.2    &69.6  &{\bf74.4}   &72.0   &72.3\\
\hline\hline
\end{tabular}}
}
\end{center}\label{tab_CombKernel}\vspace{-0.1in}
\end{table}

Table~\ref{tab_CombKernel} presents the kernel SVM experiments for these two new kernels (i.e., the last two columns). We can see that for majority of the datasets, these two  kernels outperform the min-max kernel. For a few datasets, the min-max kernel still performs the best (for example, ``M-Noise1''); and on these datasets, the acos kernel and the acos-$\chi^2$ kernel usually do not perform as well. Overall, these two new kernels appear to be  fairly robust combinations. Of course, the story will not be complete until  we have also studied their corresponding linearization methods.

A recent study~\cite{Proc:Li_UAI14} explored the idea of combing the ``resemblance'' kernel with the linear kernel,  designed only for sparse non-binary data. Since most of the datasets we experiment with are not  sparse,  we can not directly use the special kernel developed in~\cite{Proc:Li_UAI14}.  

Now  we study the linearization  methods for these two new kernels, which turn out to be easy. Take the MM-acos kernel as an example.  We can separately and independently generate samples for the min-max kernel and the acos kernel. The sample for the min-max kernel can be viewed as a binary vector with one 1. For example, if the sample for the min-max kernel is $[0,\ 0,\ 1,\ 0]$ and the sample for the acos kernel is -1.
Then we can encode the combined sample as $[0,\ 0,\ 0,\ 0,\ 1,\ 0,\ 0,\ 0]$. If the sample for the acos kernel is 1, then the combined vector becomes  $[0,\ 0,\ 0,\ 0,\ 0,\ 1,\ 0,\ 0]$. Basically, if the $j$-th location in the vector corresponding to  the original min-max sample is 1, then the combined vector will double the length and all the entries will be zero except the ($2j$-1)-th or ($2j$)-th location, depending on the sample value of acos kernel.

Clearly, the idea also applies for combining  min-max kernel with  RBF kernel. We just need to replace the ``1'' in the vector for the min-max kernel sample with the sample  of the RBF kernel.

\begin{figure}[h!]
\begin{center}

\hspace{-0in}\mbox{
\includegraphics[width=2.2in]{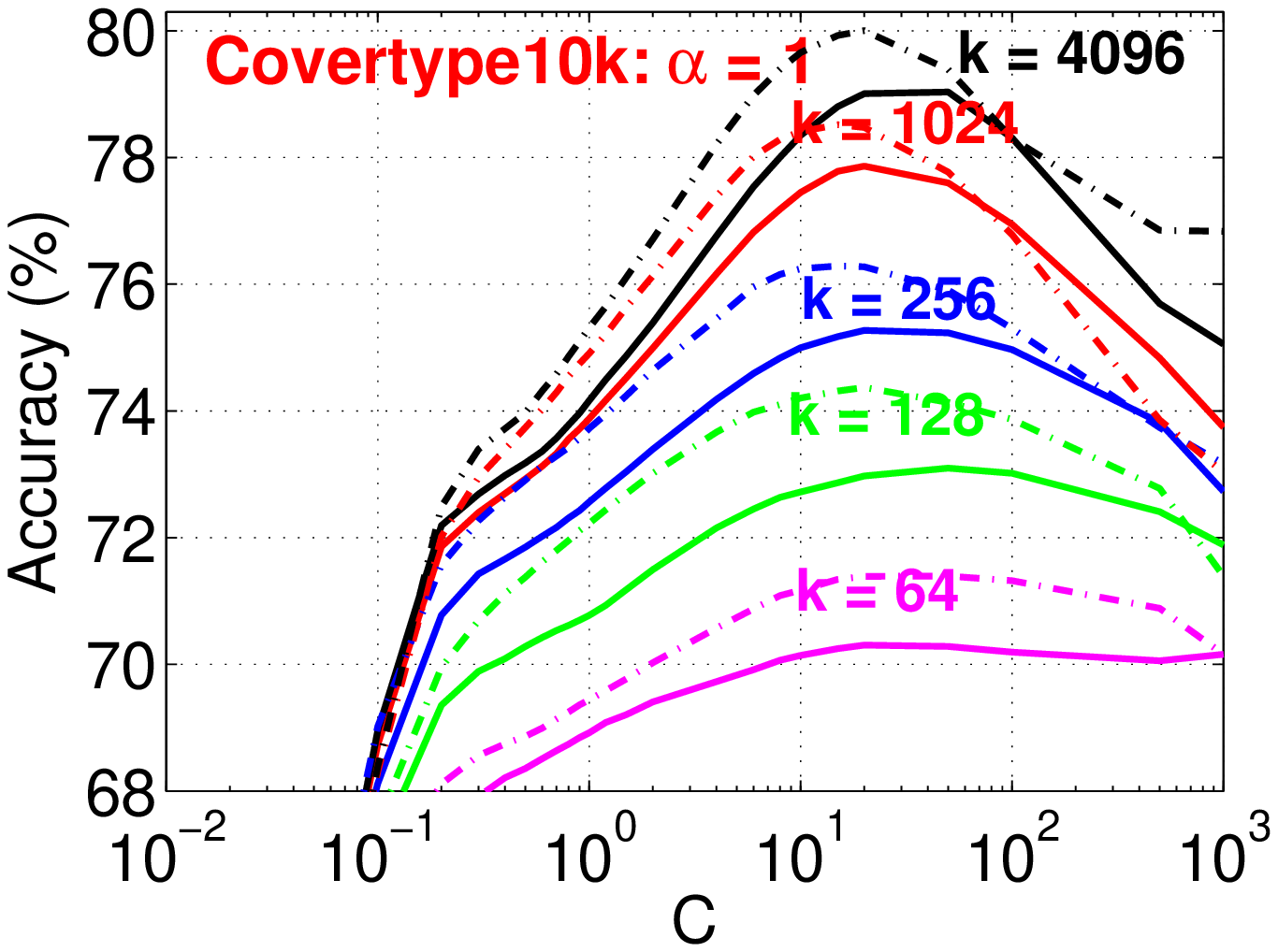}\hspace{0.3in}
\includegraphics[width=2.2in]{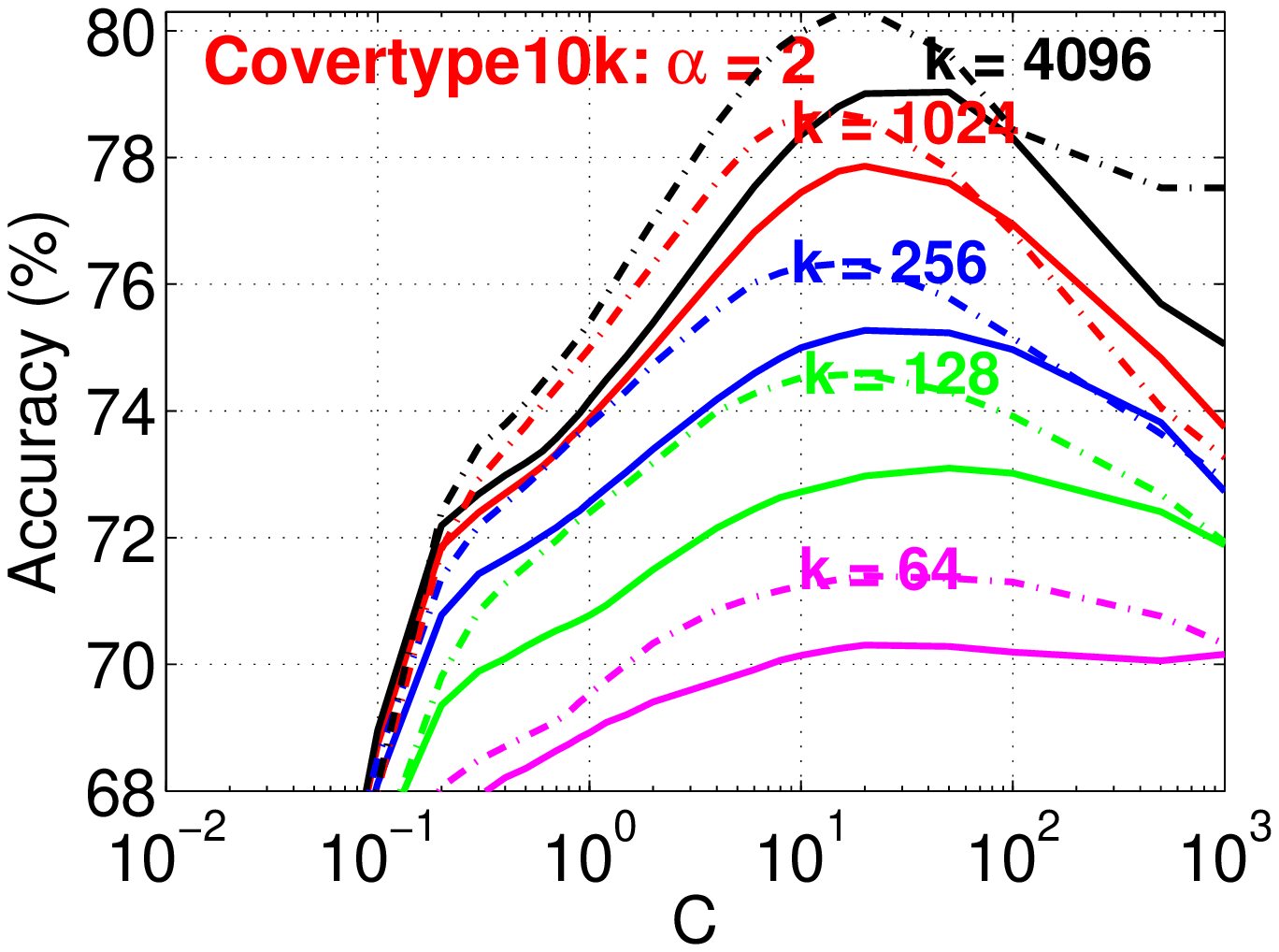}
}

\vspace{-0.038in}

\hspace{-0in}\mbox{
\includegraphics[width=2.2in]{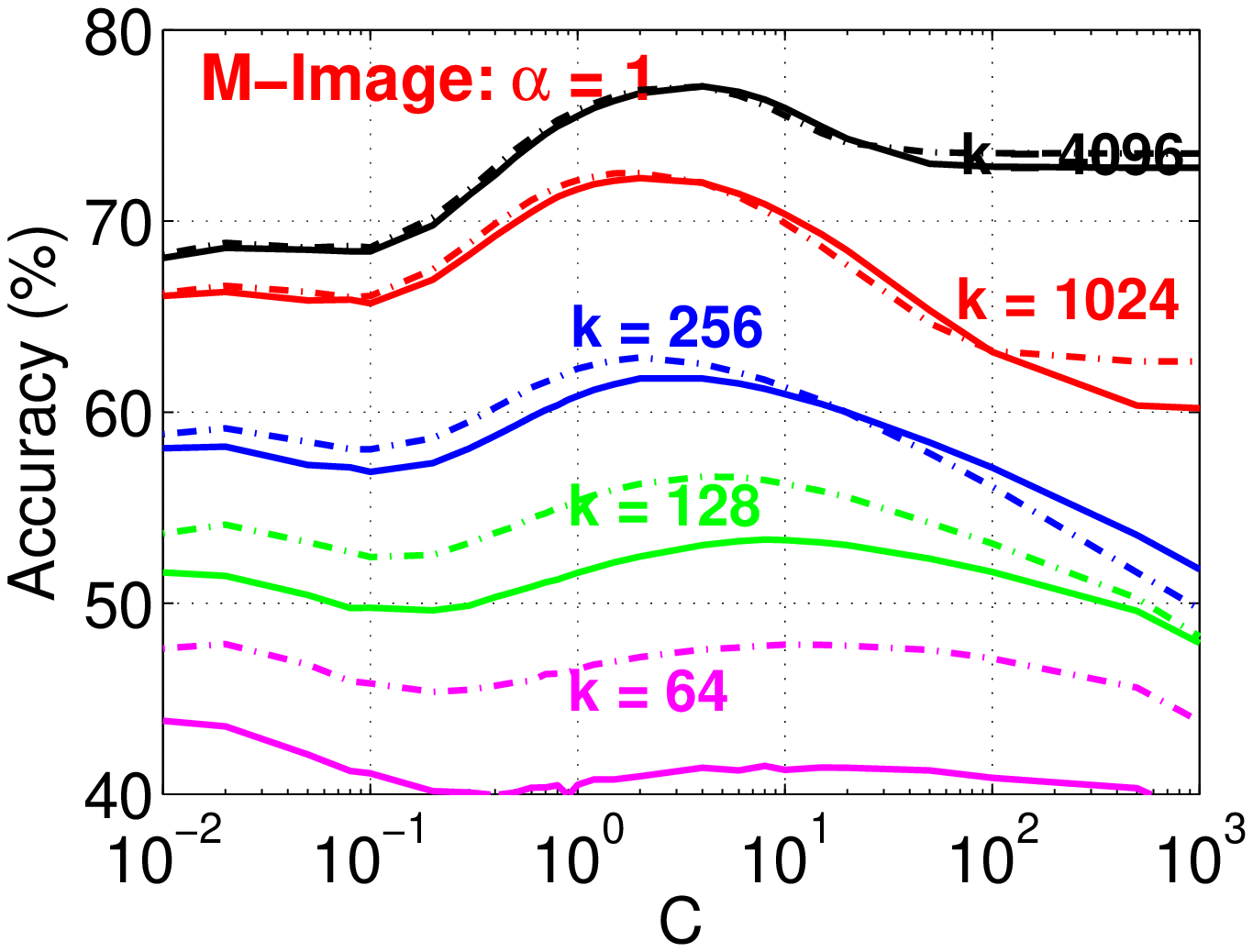}\hspace{0.3in}
\includegraphics[width=2.2in]{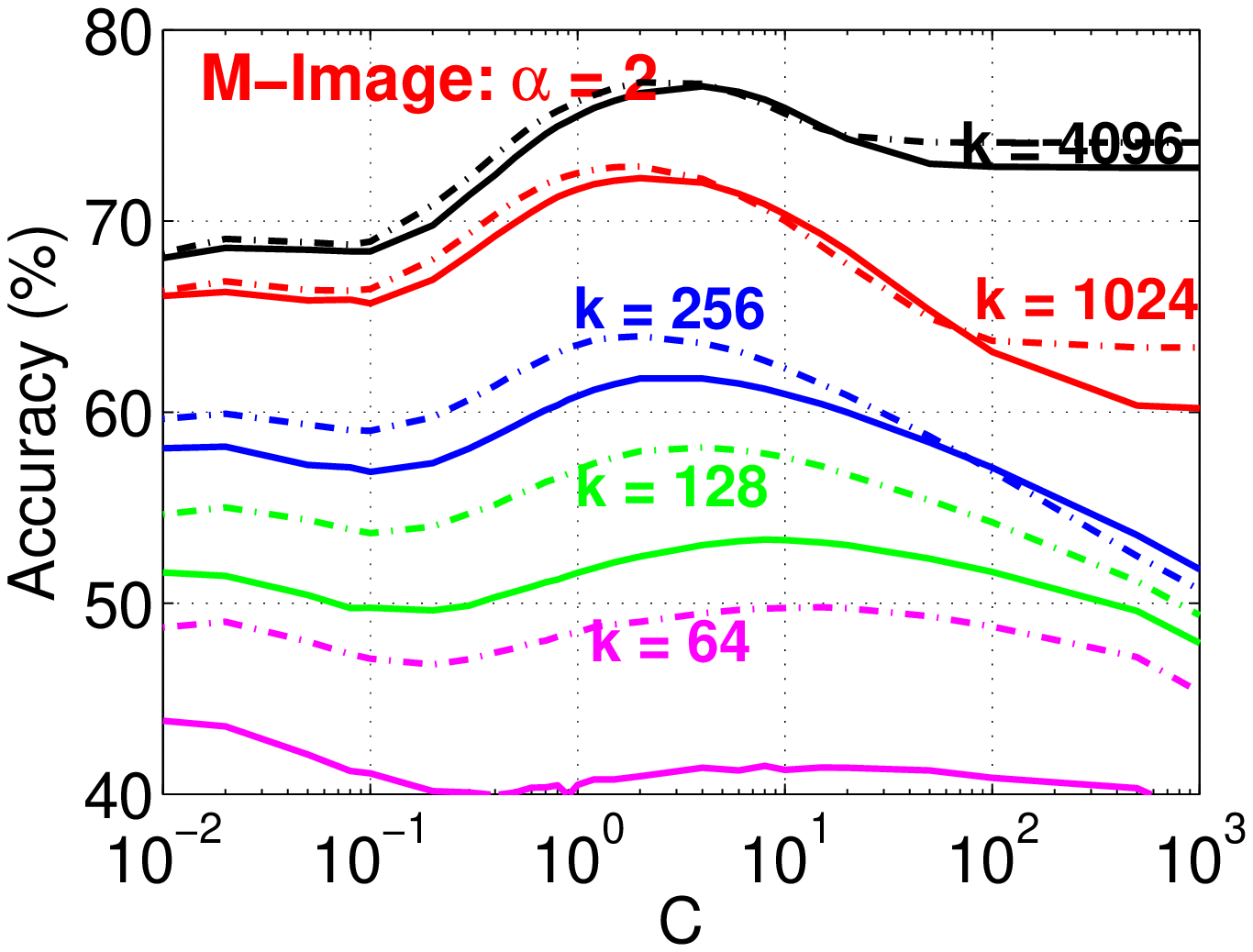}
}

\vspace{-0.038in}

\hspace{-0in}\mbox{
\includegraphics[width=2.2in]{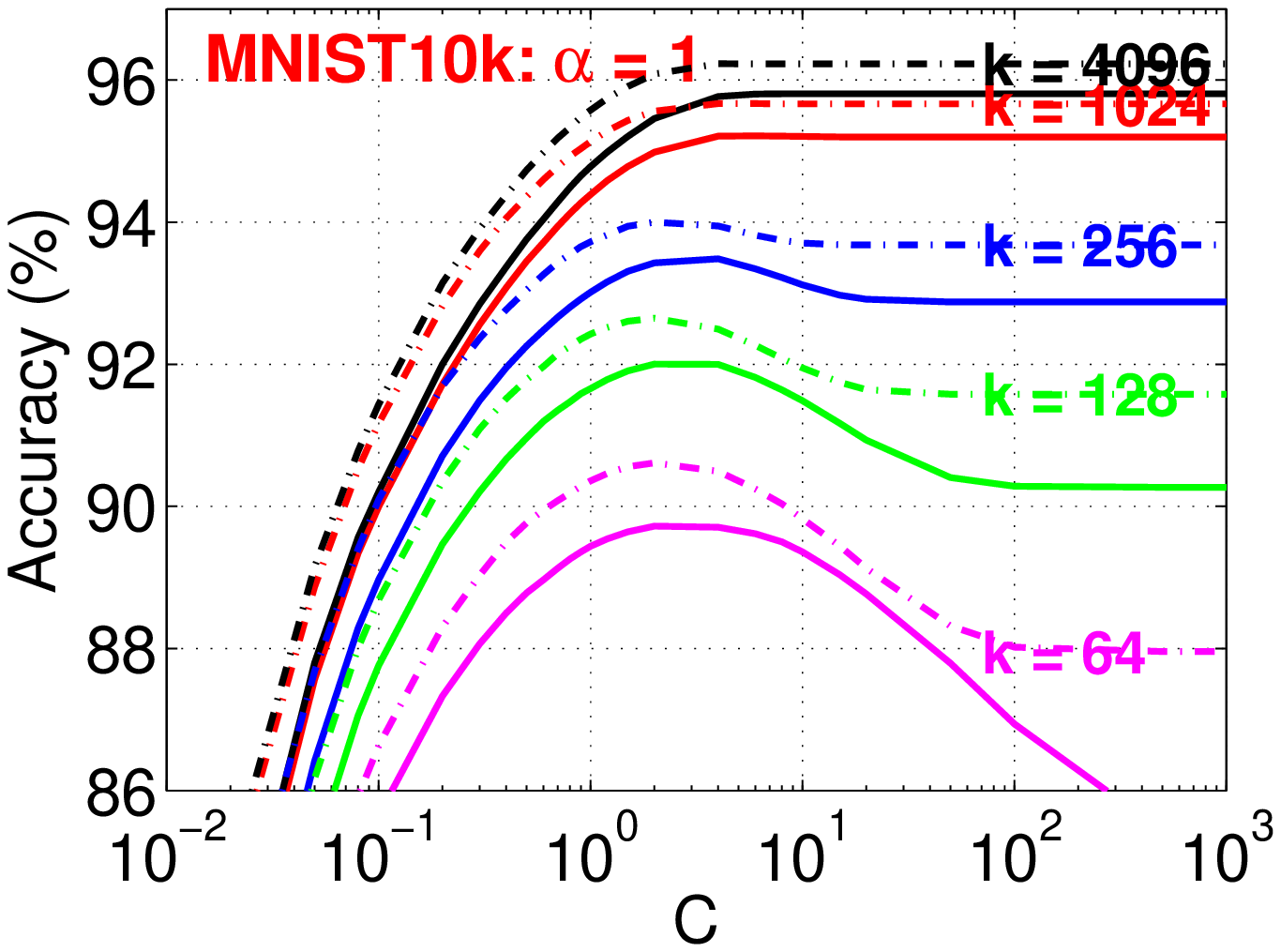}\hspace{0.3in}
\includegraphics[width=2.2in]{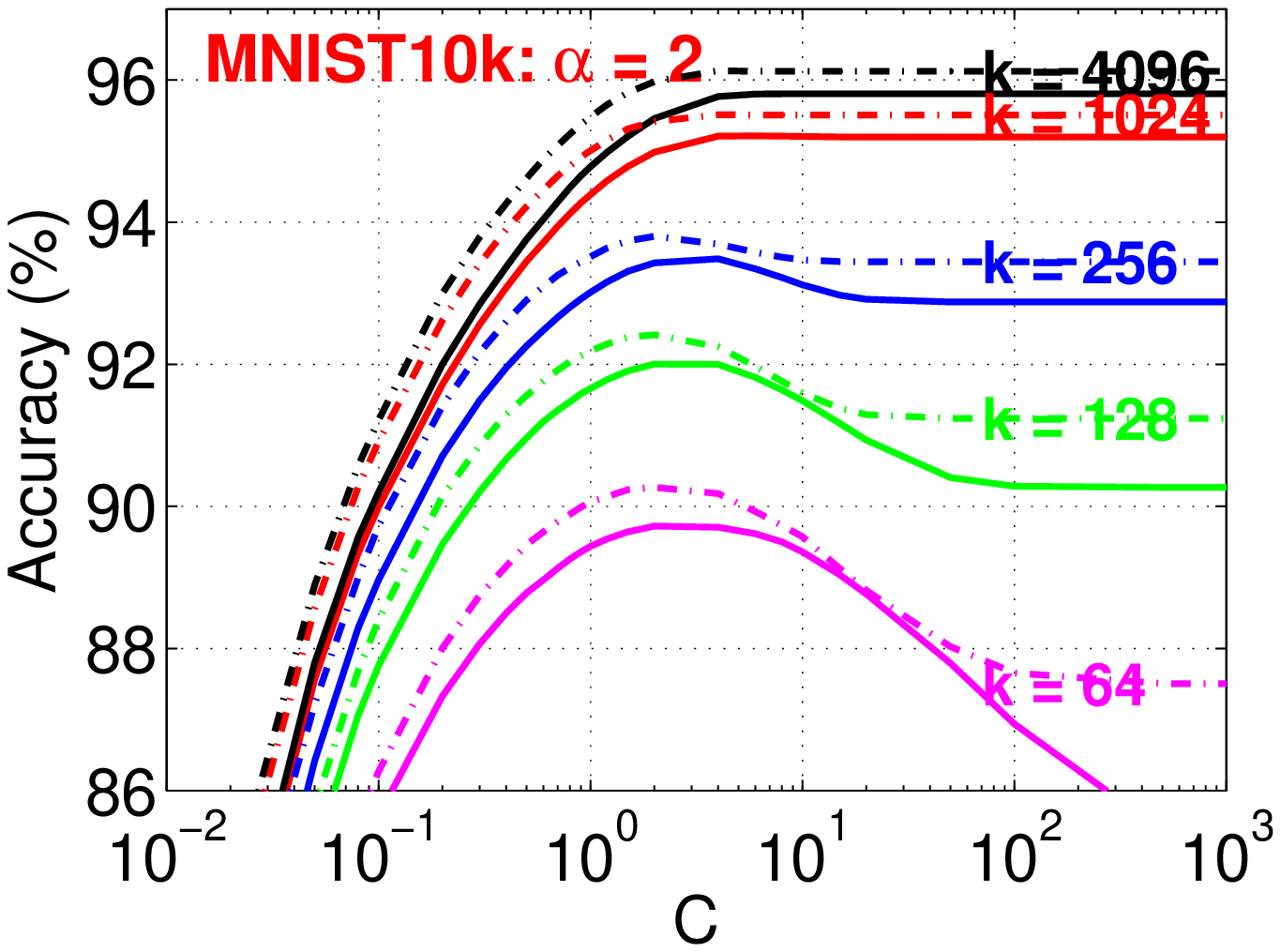}
}

\vspace{-0.038in}

\hspace{-0in}\mbox{
\includegraphics[width=2.2in]{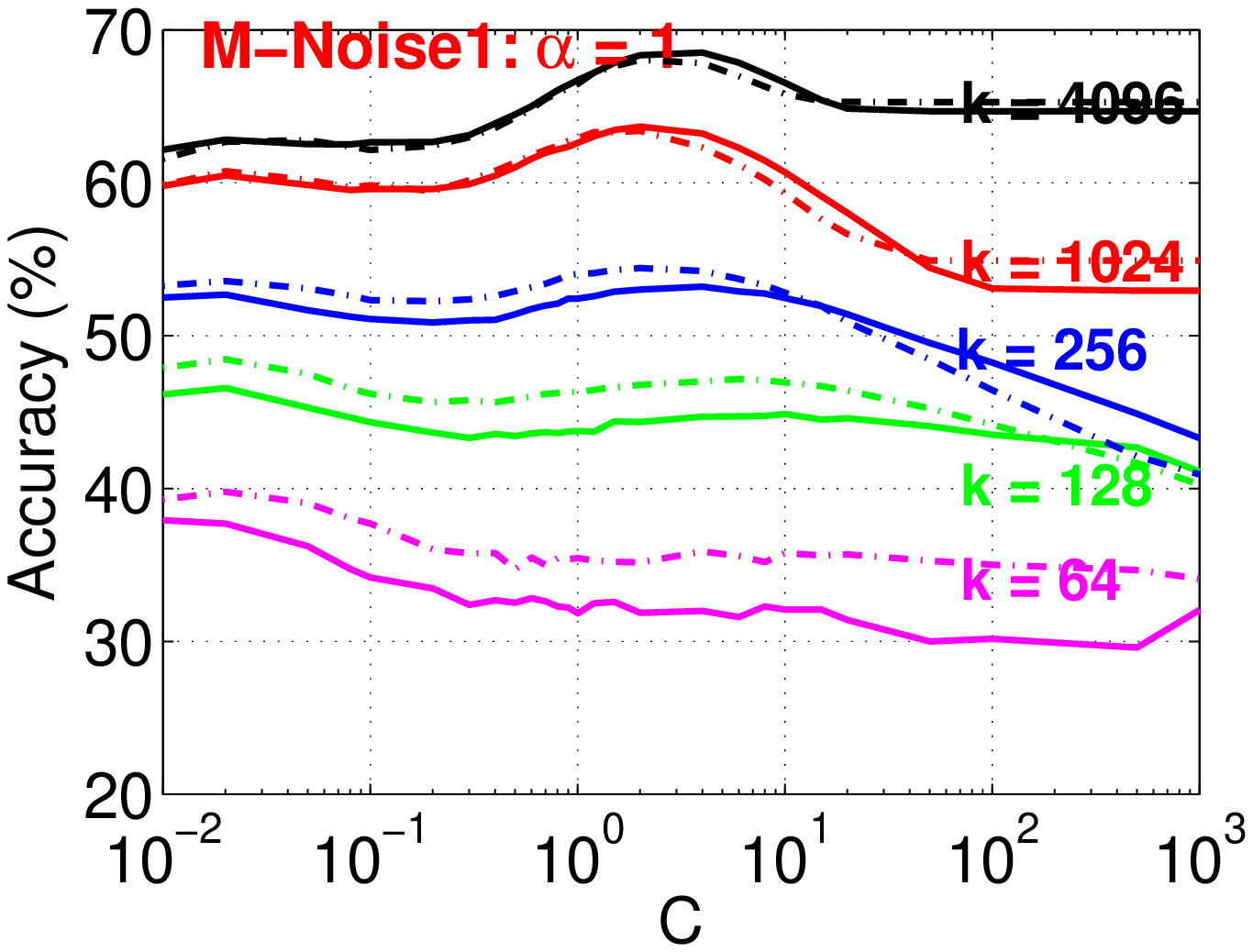}\hspace{0.3in}
\includegraphics[width=2.2in]{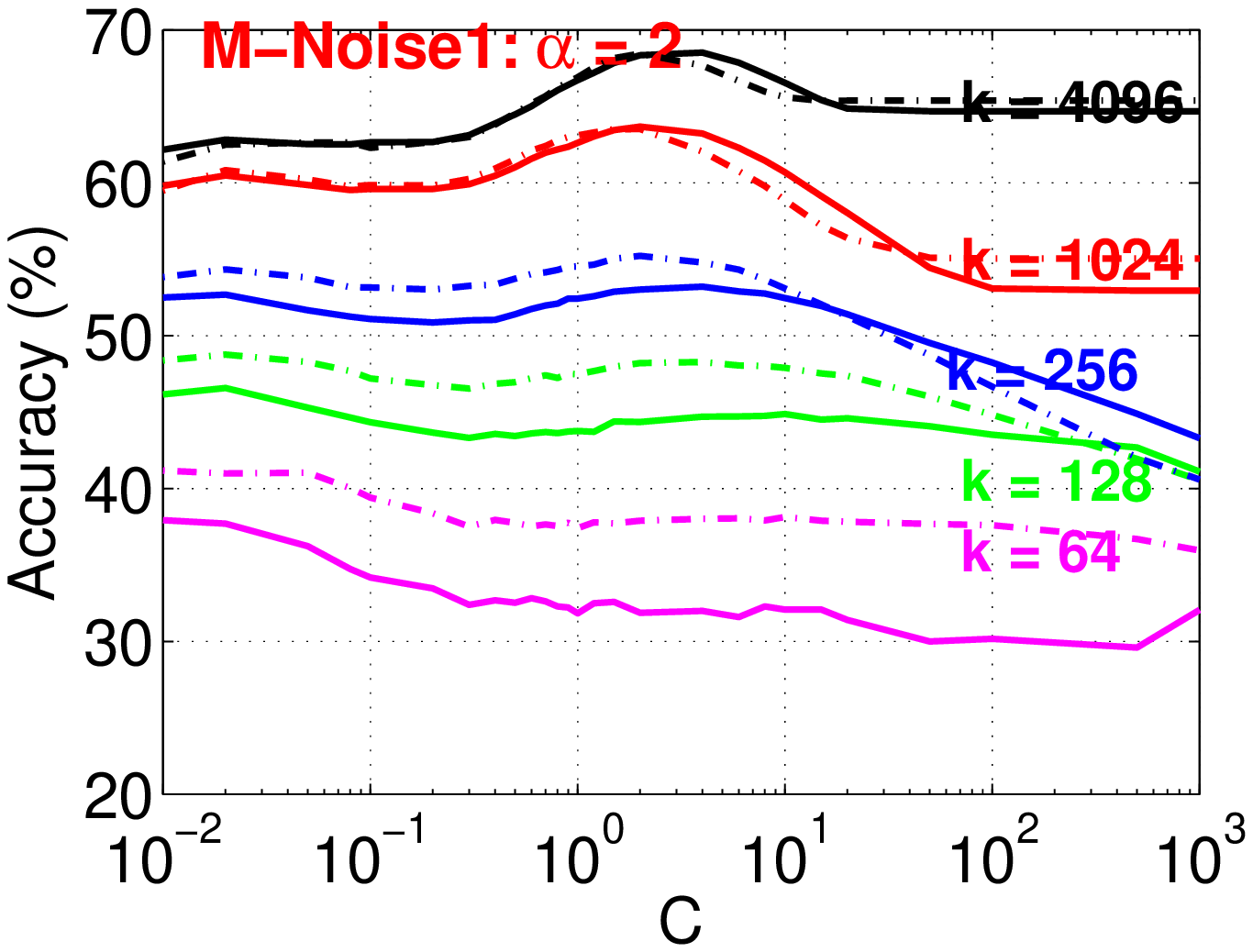}
}

\vspace{-0.038in}

\hspace{-0in}\mbox{
\includegraphics[width=2.2in]{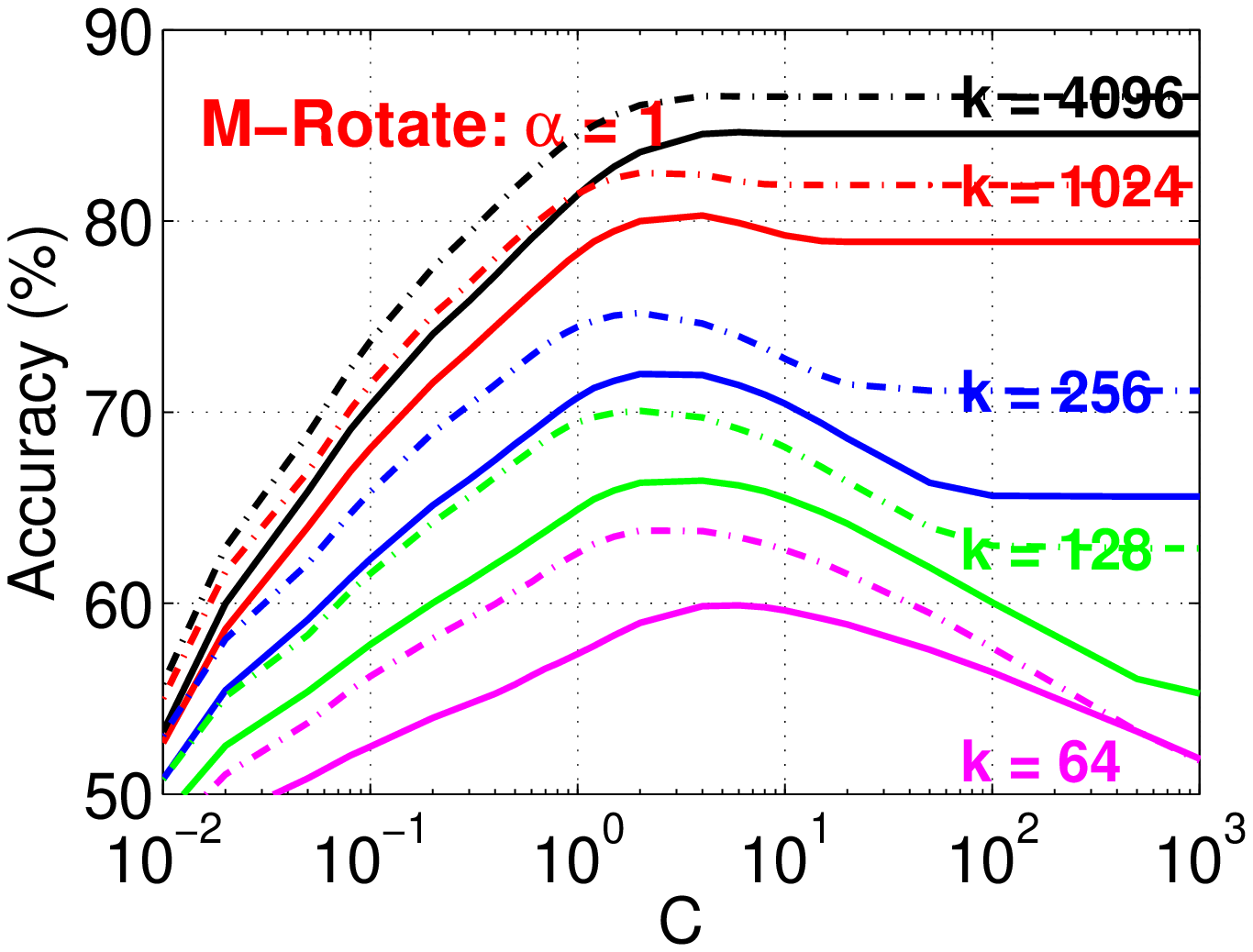}\hspace{0.3in}
\includegraphics[width=2.2in]{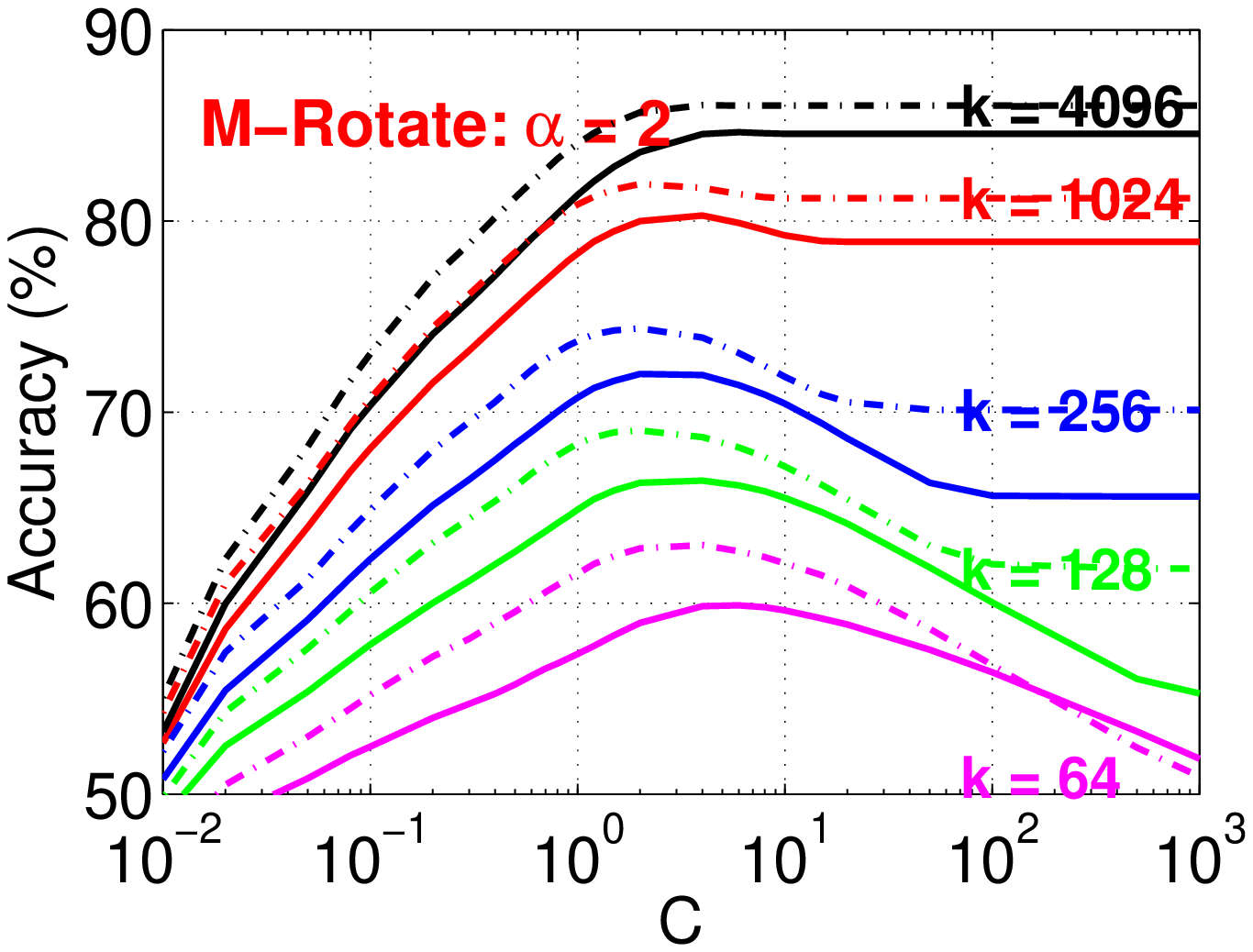}
}

\vspace{-0.3in}
\end{center}
\caption{Classification accuracies of the linearized min-max kernel  (solid curves), the MM-acos kernel (right panels ($\alpha=2$), dash-dotted curves), and the MM-acos-$\chi^2$ kernel (left panels ($\alpha=1$), dash-dotted curves), using LIBLINEAR. We report the results for $k\in\{64, 128,  256,1024, 4096\}$. We can see that the linearized MM-acos kernel and the linearized MM-acos-$\chi^2$ kernel outperform the linearized min-max kernel   when $k$ is not  large.}\label{fig_CWScombAcos1}
\vspace{-0in}
\end{figure}

\newpage\clearpage

\begin{figure}[h!]
\begin{center}

\hspace{-0in}\mbox{
\includegraphics[width=2.2in]{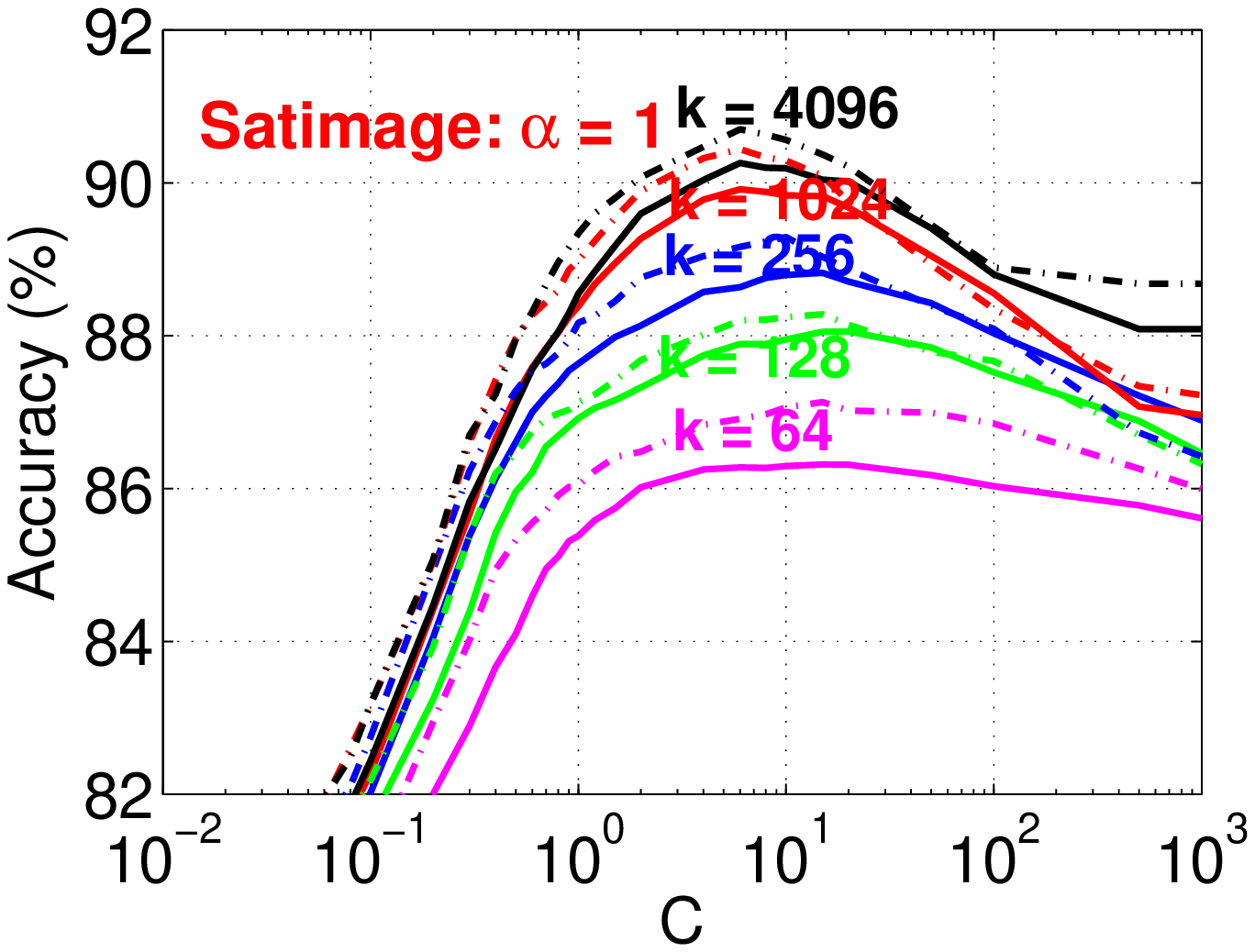}\hspace{0.3in}
\includegraphics[width=2.2in]{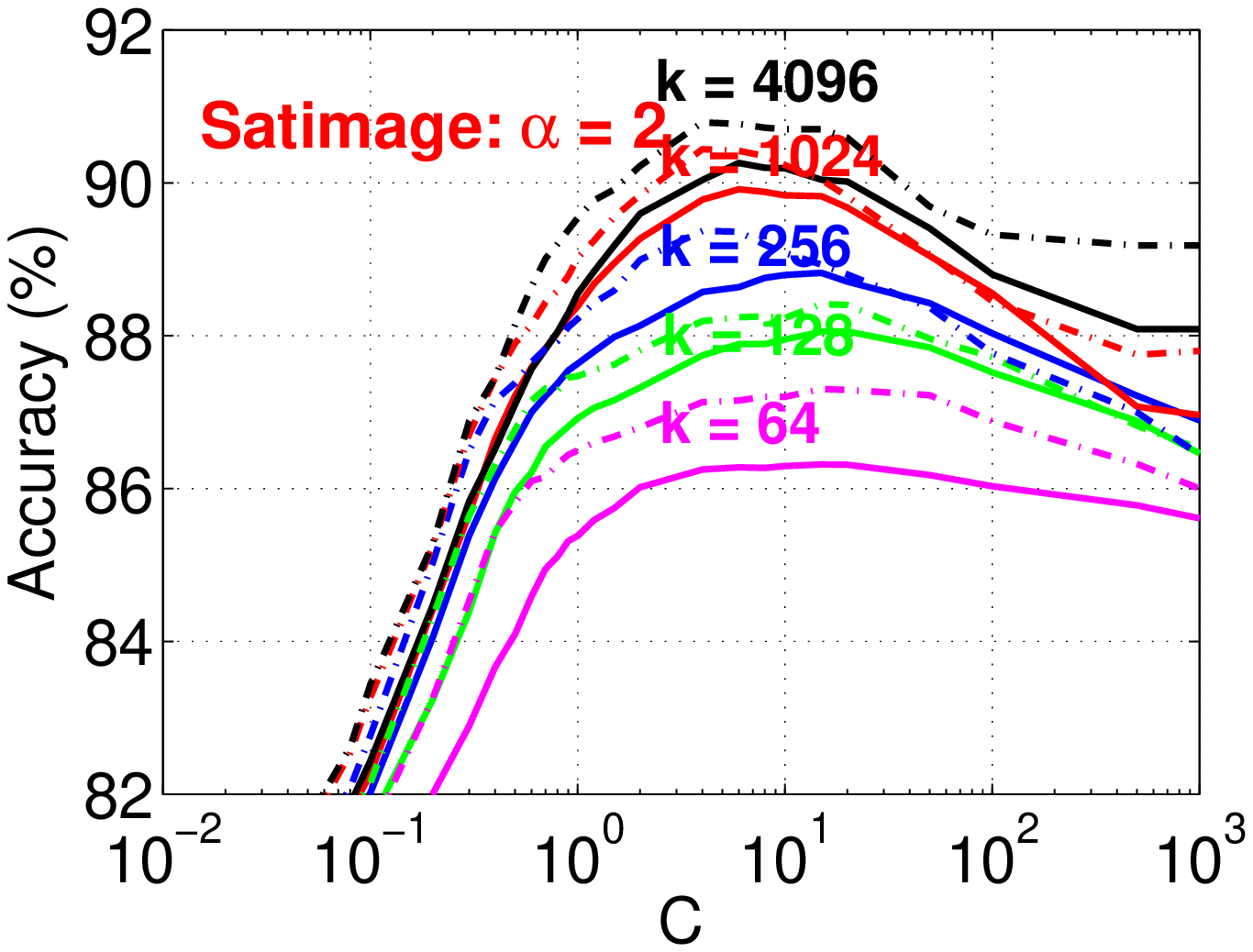}
}

\vspace{-0in}

\hspace{-0in}\mbox{
\includegraphics[width=2.2in]{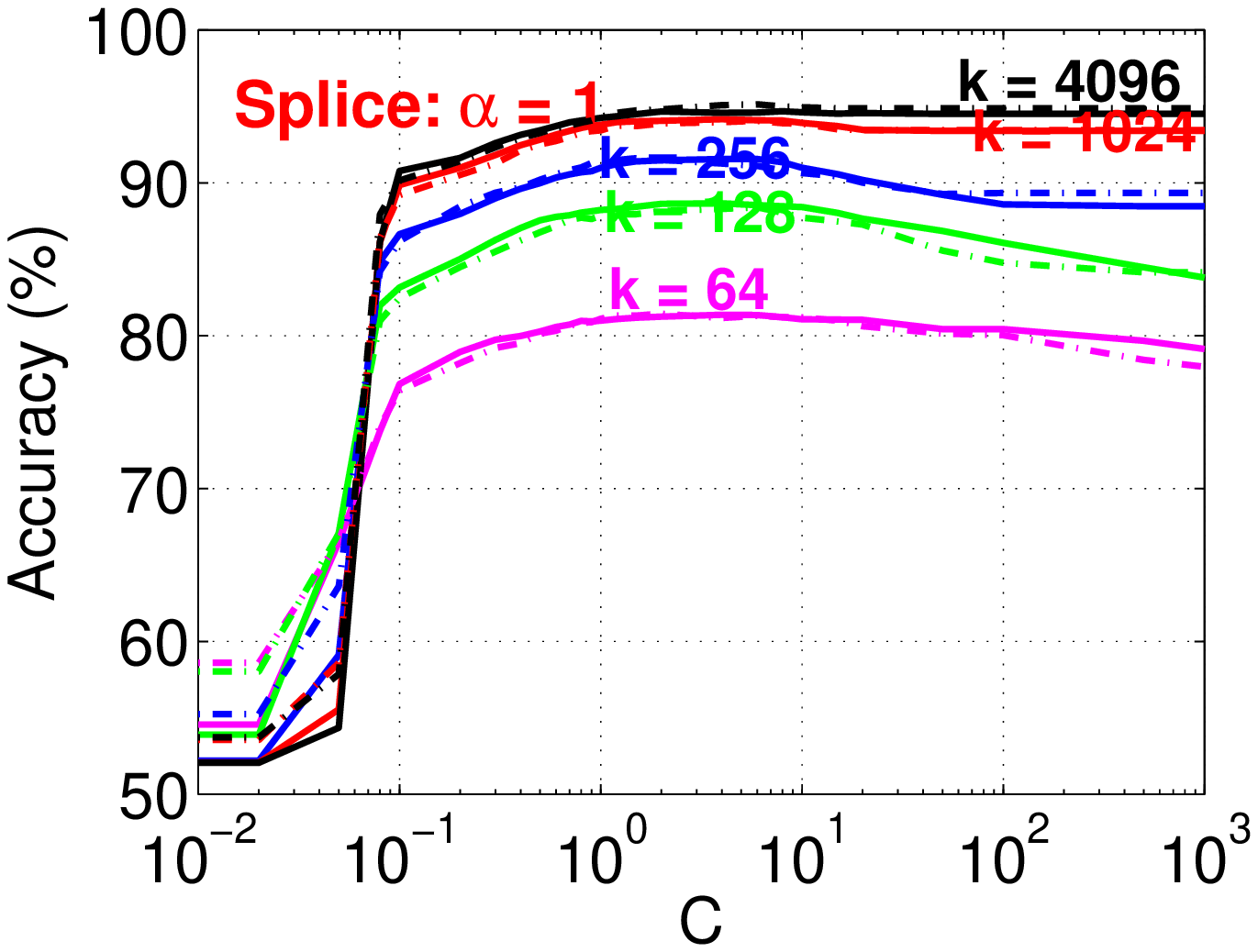}\hspace{0.3in}
\includegraphics[width=2.2in]{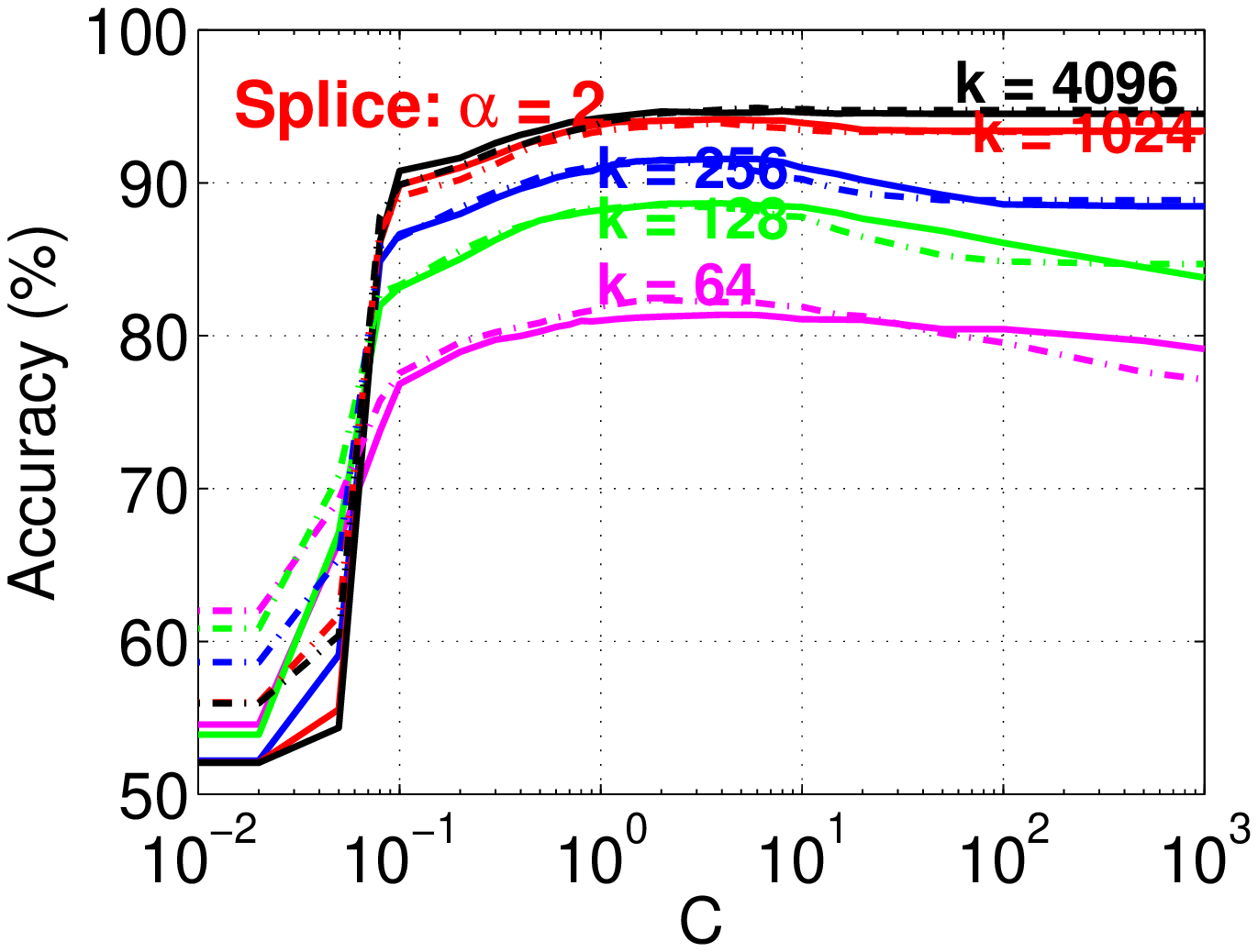}
}

\vspace{-0in}

\hspace{-0in}\mbox{
\includegraphics[width=2.2in]{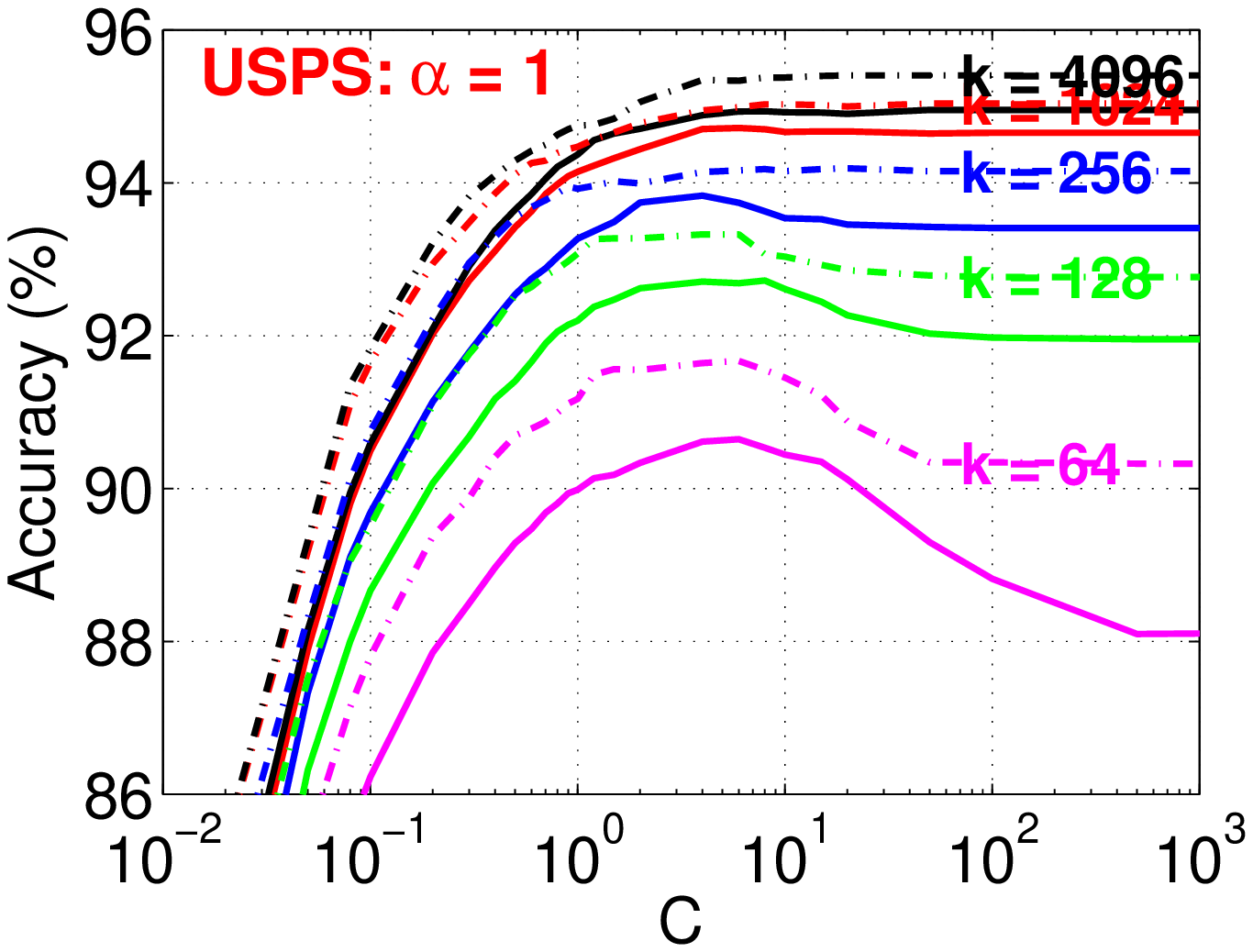}\hspace{0.3in}
\includegraphics[width=2.2in]{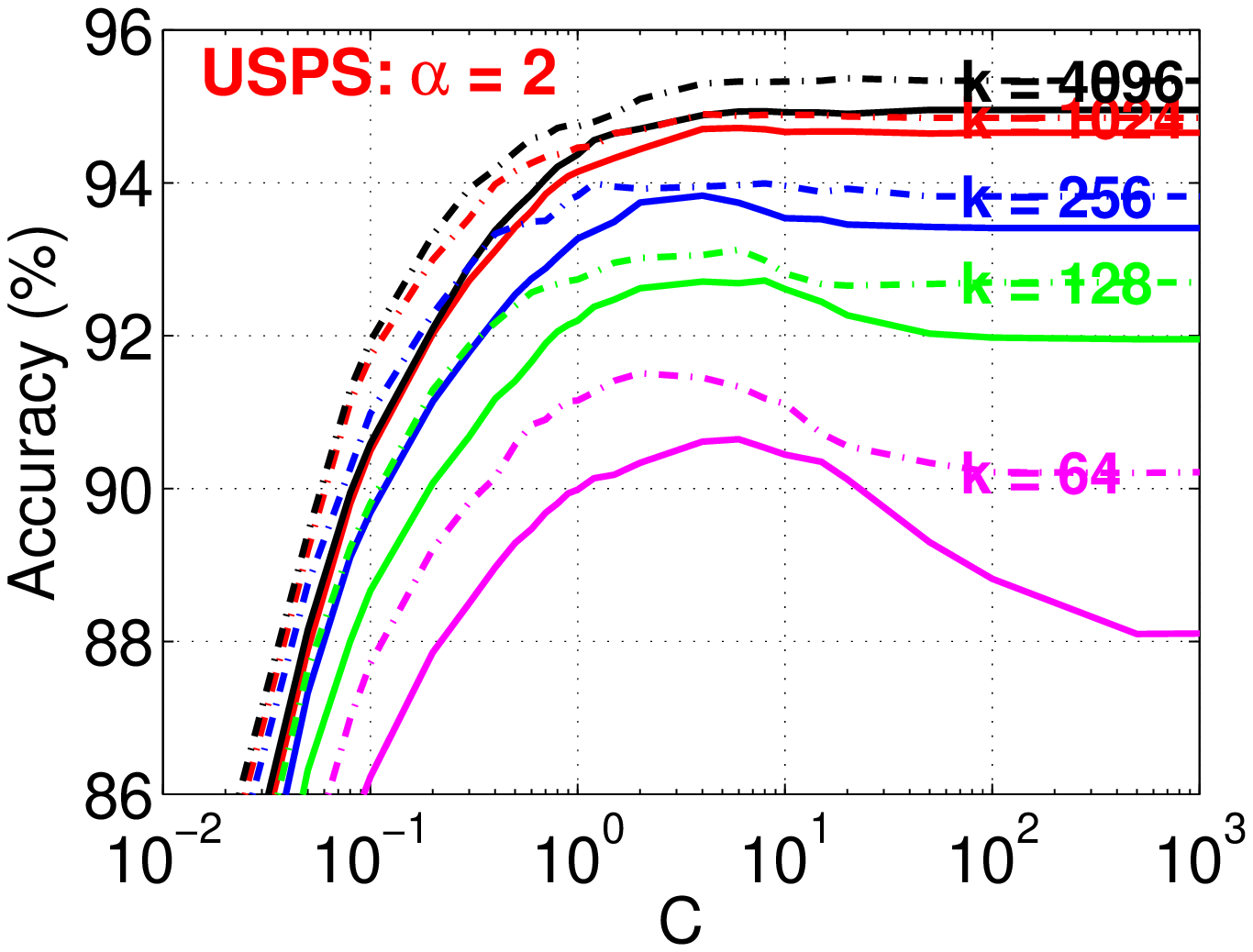}
}

\vspace{-0in}

\hspace{-0in}\mbox{
\includegraphics[width=2.2in]{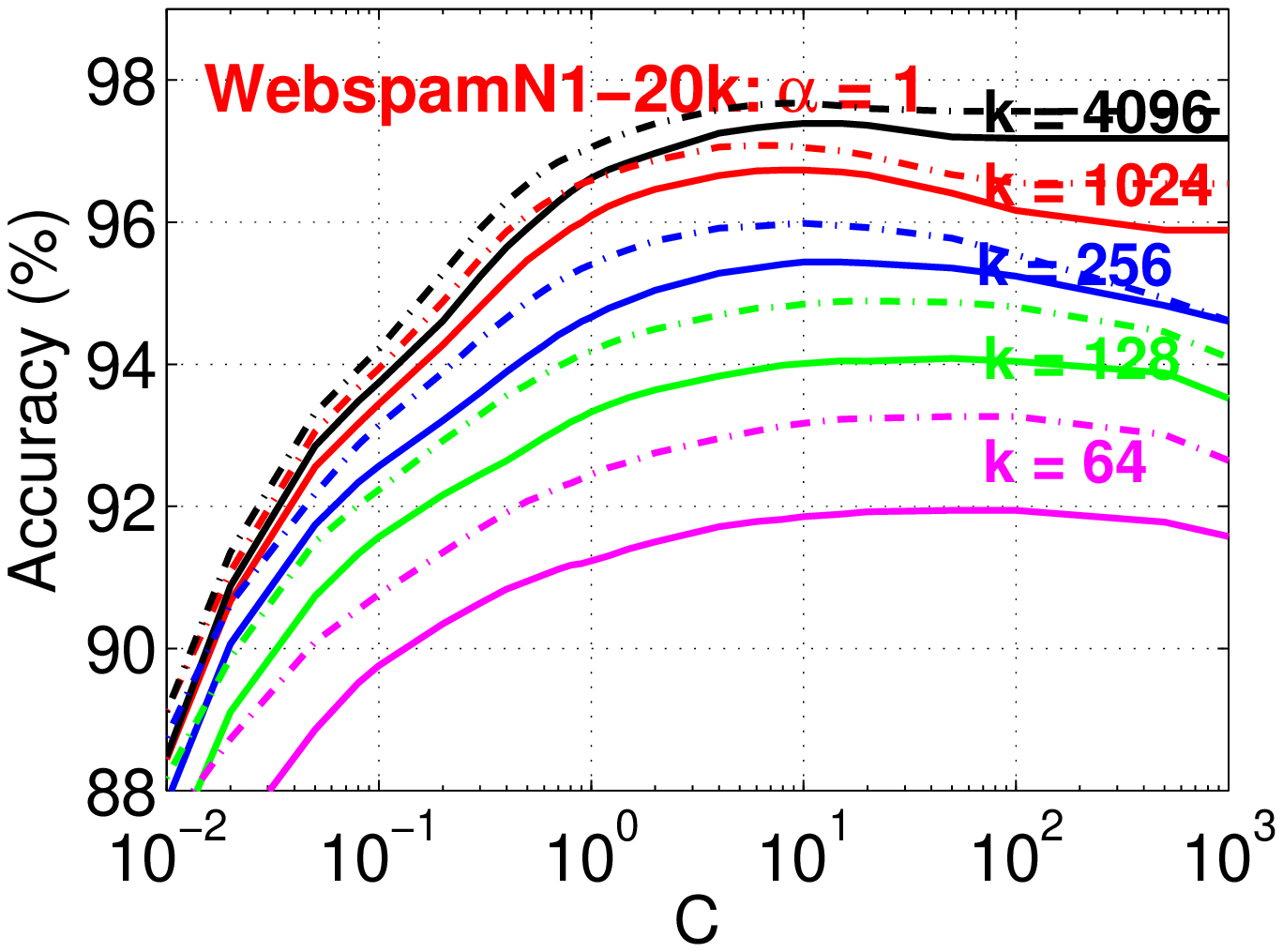}\hspace{0.3in}
\includegraphics[width=2.2in]{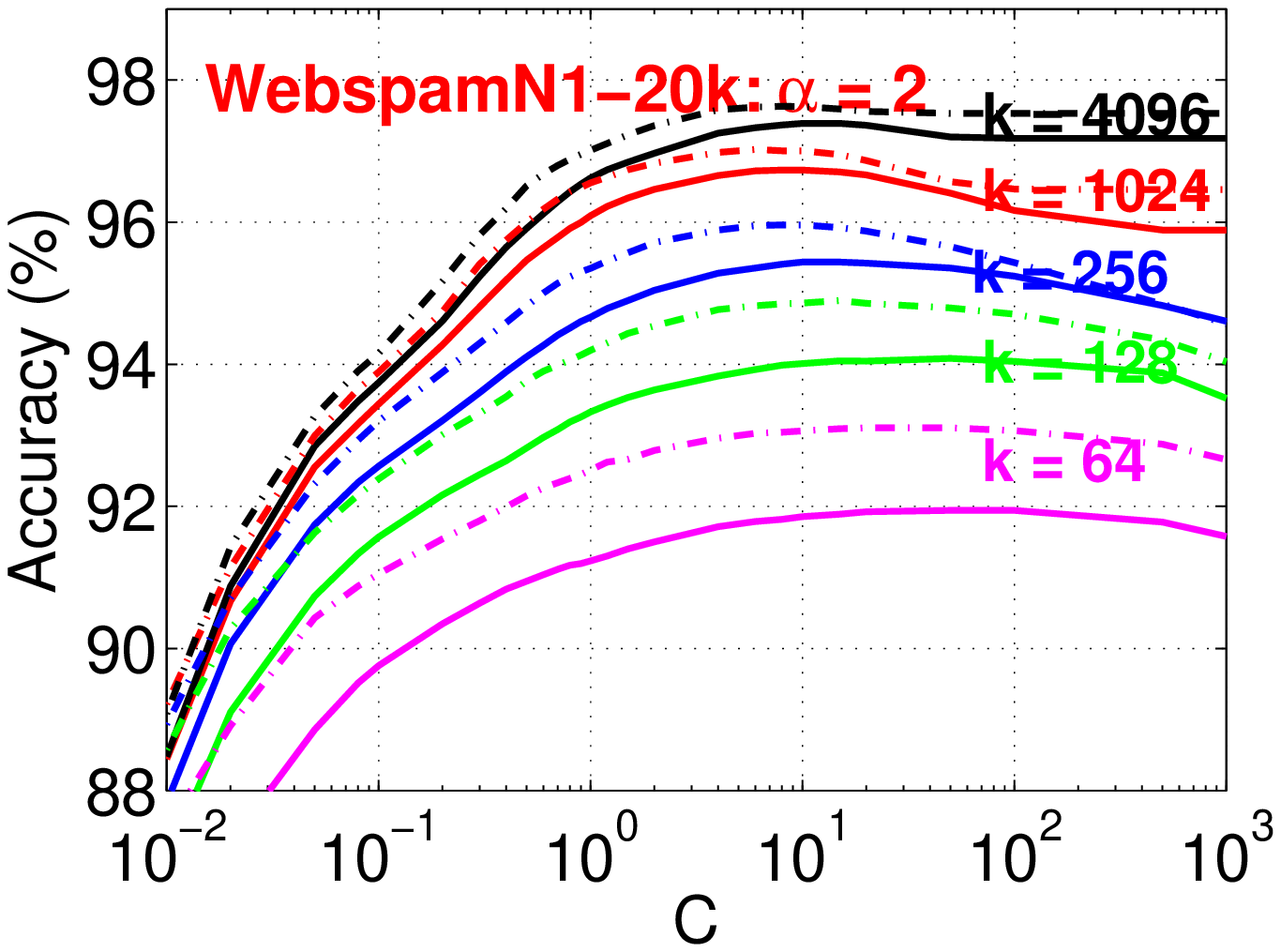}
}

\vspace{-0in}

\hspace{-0in}\mbox{
\includegraphics[width=2.2in]{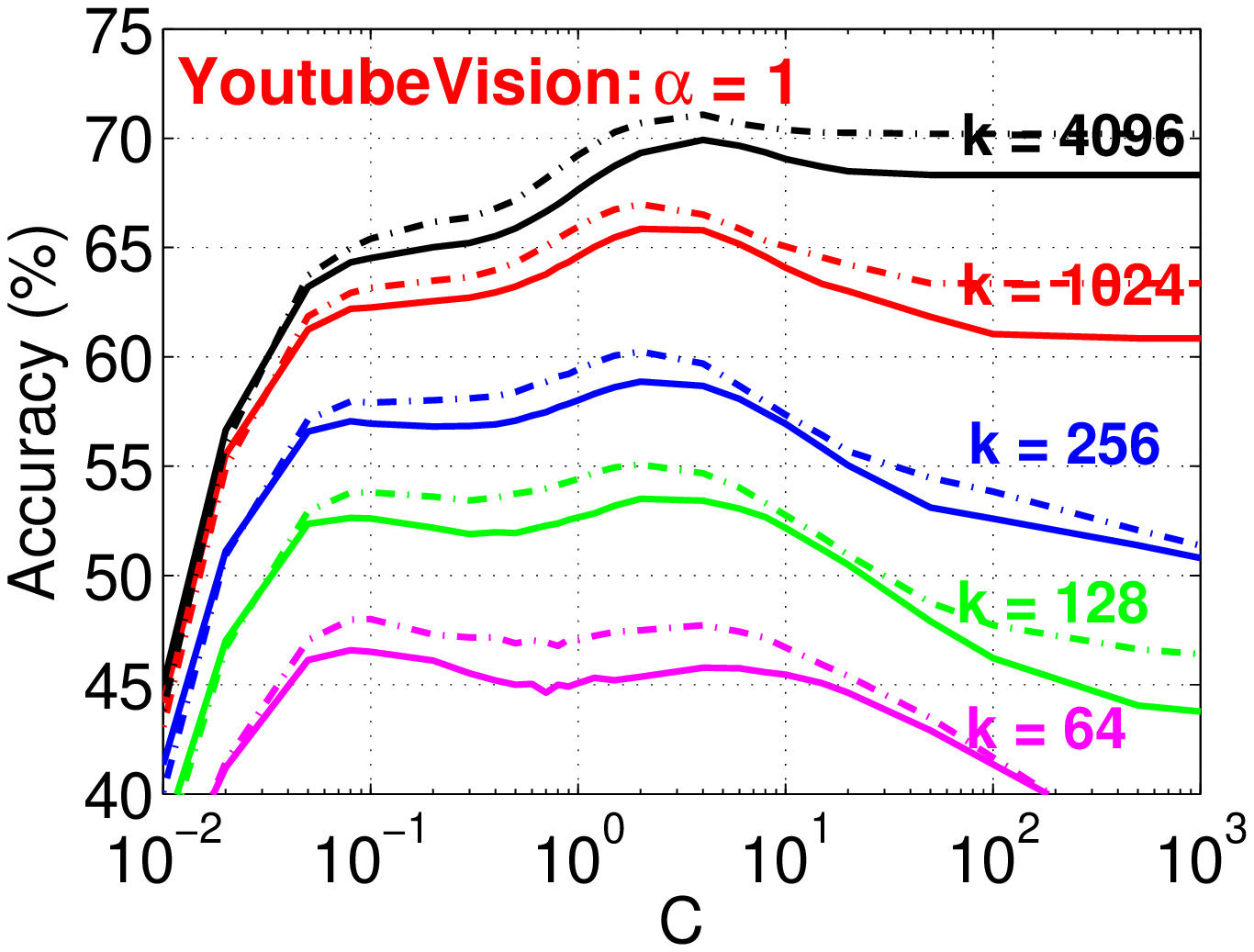}\hspace{0.3in}
\includegraphics[width=2.2in]{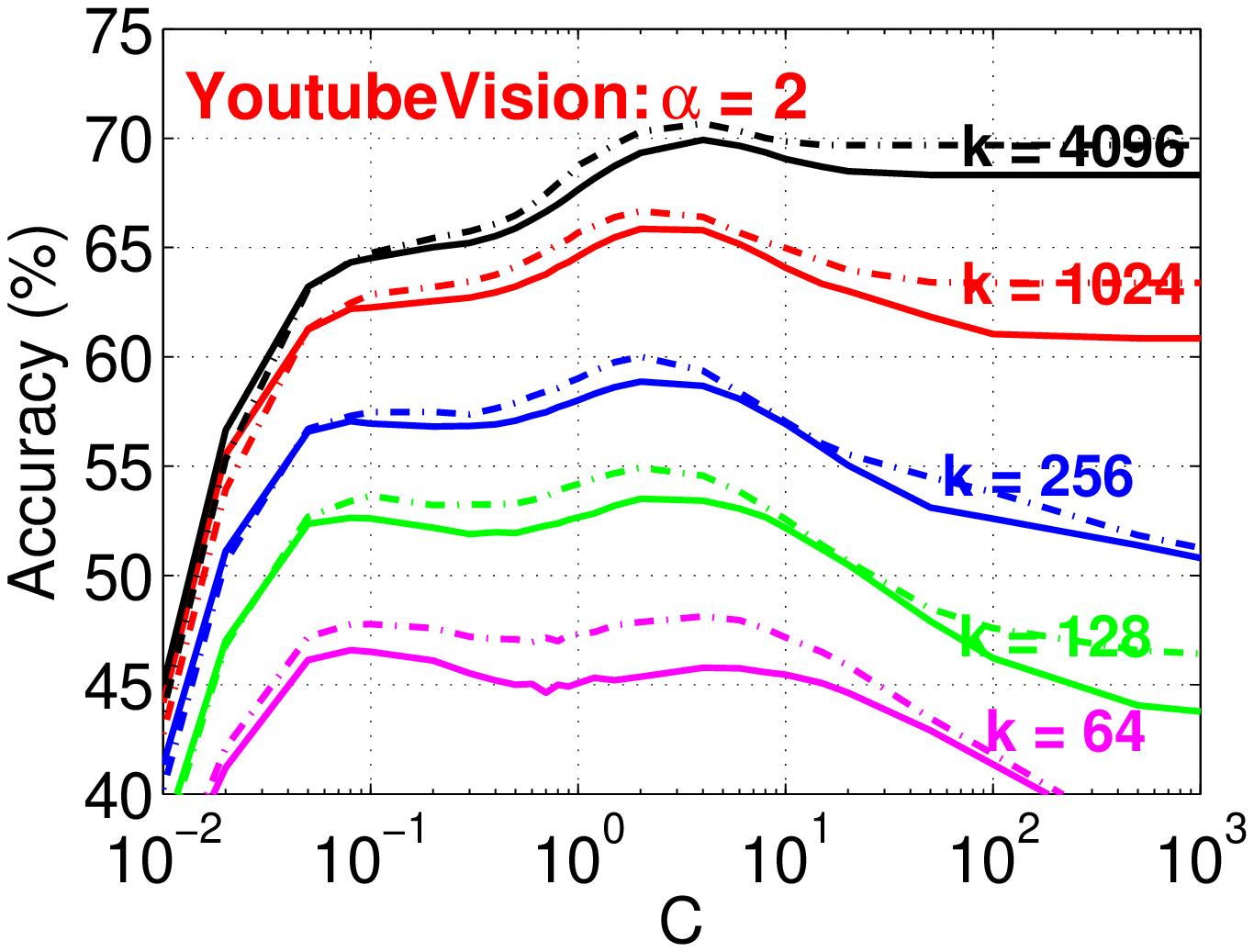}
}

\end{center}
\vspace{-0.3in}
\caption{Classification accuracies of the linearized min-max kernel  (solid curves), the MM-acos kernel (right panels ($\alpha=2$), dash-dotted curves), and the MM-acos-$\chi^2$ kernel (left panels ($\alpha=1$), dash-dotted curves), using LIBLINEAR. We report the results for $k\in\{64, 128,  256,1024, 4096\}$. }\label{fig_CWScombAcos2}\vspace{-0in}
\end{figure}

\newpage\clearpage

Figure~\ref{fig_CWScombAcos1} and Figure~\ref{fig_CWScombAcos2} report the linear SVM results using linearized data for the MM-acos kernel (right panels) and the MM-acos-$\chi^2$ kernel (left panels), to compare with the results using linearized data for the min-max kernel (solid curves). We can see that the linearization methods for the MM-acos kernel and the MM-acos-$\chi^2$ kernel  outperform the linearization method for the min-max kernel when $k$ is not large. These preliminary results are encouraging. 

\section{Conclusion}

Nonlinear kernels can be potentially very useful if there are efficient (in both storage and memory) algorithms for computing them. It has been known that the RBF kernel, the acos kernel, and the acos-$\chi^2$ kernel can be linearized via randomization algorithms. There are two major aspects when we compare nonlinear kernels: (i) the accuracy of the original kernel; (ii) how many samples are needed in order to reach a good accuracy. In this paper, we try to address these two issues by providing an extensive empirical study on a wide variety of publicly available datasets. \\

To simplify the linearization procedure for the RBF kernel, we propose the folded RBF (fRBF) kernel and demonstrates that its performance (either with the original kernel or with linearization) is very similar to that of the RBF kernel. On the other hand, our extensive nonlinear kernel SVM experiments demonstrate  that the RBF/fRBF kernels, even with the best-tuned parameters, do not always achieve the best accuracies. The min-max kernel (which is tuning-free) in general performs well (except for some very low dimensional datasets). The acos kernel and the acos-$\chi^2$ kernel also  perform reasonably well.

Linearization is a crucial step in order to use nonlinear kernels for large-scale applications. Our experimental study illustrates that the linearization method for the min-max kernel, called ``0-bit CWS'', performs  well in that it does not require a large number of samples  to reach a good accuracy. In comparison, the linearization methods for the RBF/fRBF kernels and the acos/acos-$\chi^2$ kernels typically require many more samples (e.g., $\geq 4096$).

Our study motivates  two interesting  research problems for future work: (i) how to design better (and still linearizable) kernels to improve the tuning-free kernels; (ii) how to improve the linearization algorithms for the RBF/fRBF kernels as well as the acos/acos-$\chi^2$ kernels, in order to reduce the required sample sizes.  The interesting and simple idea of combing two nonlinear kernels by multiplication appears to be effective but we still hope to find an even better strategy in the future.\\

Another challenging task is to develop (linearizable) kernel algorithms to compete with (ensembles of) trees in terms of accuracy. It is known that tree algorithms are usually slow. Even though the parallelization of trees is  easy, it will still consume excessive energy (e.g., electric power).  One can see from~\cite{Proc:ABC_ICML09,Proc:ABC_UAI10}  that trees are in general perform really well in terms of accuracy and can be remarkably more accurate than other methods in some datasets (such as ``M-Noise1'' and ``M-Image''). On top of the fundamental works~\cite{Article:FHT_AS00,Article:Friedman_AS01}, the recent papers~\cite{Proc:ABC_ICML09,Proc:ABC_UAI10} improved tree algorithms via two ideas: (i) an explicit tree-split formula using 2nd-order derivatives; (ii) a re-formulation of the classical  logistic loss function which leads to a  different set of first and second derivatives  from  textbooks. Ideally, it would be great to develop statistical machine learning algorithms which are as accurate as (ensembles of) trees and are as fast as linearizable kernels.

\clearpage\newpage

\appendix

\section{Consistent Weighted Sampling}\label{app_CWS}

\begin{algorithm}{
\textbf{Input:} Data vector $u$ = ($u_i\geq 0$, $i=1$ to $D$)

\textbf{Output:} Consistent uniform sample ($i^*$, $t^*$)

\vspace{0.08in}

For $i$ from 1 to $D$

\hspace{0.25in}$r_i\sim Gamma(2, 1)$, \ $c_i\sim Gamma(2, 1)$,  $\beta_i\sim Uniform(0, 1)$

\hspace{0.2in} $t_i\leftarrow \lfloor \frac{\log u_i }{r_i}+\beta_i\rfloor$, \ $y_i\leftarrow \exp(r_i(t_i - \beta_i))$,\  $a_i\leftarrow c_i/(y_i \exp(r_i))$

End For

$i^* \leftarrow arg\min_i \ a_i$,\hspace{0.3in}  $t^* \leftarrow t_{i^*}$
}\caption{Consistent Weighted Sampling (CWS)}
\label{alg_CWS}
\end{algorithm}

Given a data vector $u\in\mathbb{R}^D$, Alg.~\ref{alg_CWS} (following~\cite{Proc:Ioffe_ICDM10})  provides the procedure for generating one CWS sample $(i^*, t^*$). In order to generate $k$ such samples, we have to repeat the procedure  $k$ times using  independent  random numbers $r_i$, $c_i$, $\beta_i$.

\section{Proof of Lemma~\ref{lem_fRBF}}\label{app_fRBF}

 Let $t = \sqrt{\gamma}$. Using the bivariate normal density function, we obtain
\begin{align}\notag
&E\left(\cos(t x)\cos(t y)\right)\\\notag
=&\int_{-\infty}^\infty\int_{-\infty}^\infty \cos(t x)\cos(t y) \frac{1}{2\pi}\frac{1}{\sqrt{1-\rho^2}}e^{-\frac{x^2+y^2-2\rho xy}{2(1-\rho^2)}} dxdy\\\notag
=&\int_{-\infty}^\infty\int_{-\infty}^\infty \cos(t x)\cos(t y) \frac{1}{2\pi}\frac{1}{\sqrt{1-\rho^2}}
 e^{-\frac{x^2+y^2-2\rho xy+\rho^2x^2-\rho^2x^2}{2(1-\rho^2)}} dxdy\\\notag
=&\int_{-\infty}^\infty\frac{1}{2\pi}\frac{1}{\sqrt{1-\rho^2}}e^{-\frac{x^2}{2}} \cos(t x) dx\int_{-\infty}^\infty\cos(t y)e^{-\frac{(y-\rho x)^2}{2(1-\rho^2)}} dy\\\notag
=&\int_{-\infty}^\infty\frac{1}{2\pi}e^{-\frac{x^2}{2}} \cos(t x) dx
\int_{-\infty}^\infty\cos(t y\sqrt{1-\rho^2}+t \rho x)e^{-y^2/2} dy\\\notag
=&\int_{-\infty}^\infty\frac{1}{2\pi}e^{-\frac{x^2}{2}} \cos(t x)\cos(t \rho x) dx
\int_{-\infty}^\infty\cos(t y\sqrt{1-\rho^2})e^{-y^2/2} dy\\\notag
=&\int_{-\infty}^\infty\frac{1}{2\pi}e^{-\frac{x^2}{2}} \cos(t x)\cos(t \rho x) \sqrt{2\pi}e^{-t^2\frac{1-\rho^2}{2}} dx\\\notag
=&\frac{1}{\sqrt{2\pi}}e^{-t^2\frac{1-\rho^2}{2}}\int_{-\infty}^\infty e^{-\frac{x^2}{2}} \cos(t x)\cos(t\rho x) dx\\\notag
=&\frac{1}{\sqrt{2\pi}}e^{-t^2\frac{1-\rho^2}{2}}\frac{\sqrt{2\pi}}{2}  \left[e^{-t^2\frac{(1-\rho)^2}{2}} + e^{-t^2\frac{(1+\rho)^2}{2}}\right]\\\notag
=&\frac{1}{2}e^{-t^2(1-\rho)}+\frac{1}{2}e^{-t^2(1+\rho)}
\end{align}

\newpage

{
\bibliographystyle{abbrv}
\bibliography{../bib/mybibfile}

}

\end{document}